%% file: main.tex
\documentclass[twoside,12pt]{article}

\usepackage{blindtext}

\usepackage{jmlr2e}

\usepackage{graphicx}
\usepackage[caption = false]{subfig}
\usepackage{amsmath}
\usepackage{booktabs}
\usepackage{multirow}
\usepackage{multicol}
\usepackage{comment}
\usepackage{thmtools}
\usepackage[table]{xcolor}
\usepackage{soul}
\usepackage[linesnumbered,ruled,vlined]{algorithm2e}
\usepackage{rotating}
\input{letterfonts}

\usepackage[capitalize]{cleveref}
\crefname{section}{Sec.}{Secs.}
\Crefname{section}{Section}{Sections}
\Crefname{table}{Table}{Tables}
\crefname{table}{Tab.}{Tabs.}
\crefname{hyp}{hypothesis}{hypotheses}
\Crefname{hyp}{Hypothesis}{Hypotheses}

\newtheorem{mlemma}{Lemma}[section]

\newcommand{\del}[2]{\frac{\partial #1}{\partial #2}}
\newcommand{\Del}[3]{\frac{\partial^{#1} #2}{\partial {#3}^{#1}}}
\newcommand{\DDel}[4]{\frac{\partial^{#1} #2}{\partial {#3} \partial{#4}}}

\newcommand{\bs}[1]{\boldsymbol{#1}}

\usepackage{lastpage}

\ShortHeadings{Mutual Information Optimization}{S. Manna et al.}
\firstpageno{1}

\begin{document}

\title{MIO: Mutual Information Optimization using Self-Supervised Binary Contrastive Learning}

\author{\name Siladittya Manna \email smanna@isical.ac.in \\
       \addr Computer Vision and Pattern Recognition Unit\\
       Indian Statistical Institute Kolkata, India\\
       \AND
       \name Saumik Bhattacharya \email saumik@ece.iitkgp.ac.in \\
       \addr Department of Electrical and Electronic Communication Engineering\\
       Indian Institute of Technology Kharagpur, India\\
       \AND
       \name Umapada Pal \email umapada@isical.ac.in \\
       \addr Computer Vision and Pattern Recognition Unit\\
       Indian Statistical Institute Kolkata, India\\
       }

\editor{}

\maketitle

\begin{abstract}
Self-supervised contrastive learning frameworks have progressed rapidly over the last few years. In this paper, we propose a novel loss function for contrastive learning. We model our pre-training task as a binary classification problem to induce an implicit contrastive effect. We further improve the n\"aive loss function after removing the effect of the positive-positive repulsion and incorporating the upper bound of the negative pair repulsion. Unlike existing methods, the proposed loss function optimizes the mutual information in positive and negative pairs. We also present a closed-form expression for the parameter gradient flow and compare the behaviour of self-supervised contrastive frameworks using Hessian eigenspectrum to analytically study their convergence. The proposed method outperforms SOTA self-supervised contrastive frameworks on benchmark datasets such as CIFAR-10, CIFAR-100, STL-10, and Tiny-ImageNet. After 200 pretraining epochs with ResNet-18 as the backbone, the proposed model achieves an accuracy of 86.36\%, 58.18\%, 80.50\%, and 30.87\% on the CIFAR-10, CIFAR-100, STL-10, and Tiny-ImageNet datasets, respectively, and surpasses the SOTA contrastive baseline by 1.93\%, 3.57\%, 4.85\%, and 0.33\%, respectively. The proposed framework also achieves a state-of-the-art accuracy of 78.4\% (200 epochs) and 65.22\% (100 epochs) Top-1 Linear Evaluation accuracy on ImageNet100 and ImageNet1K datasets, respectively.
\end{abstract}


\section{Introduction}
\label{sec:intro}

Self-supervised learning (SSL) has emerged as one of the pillars of its super-domain unsupervised learning. Self-supervised learning primarily aims to learn representations from data without human-annotated labels. Generally, this primary aim is fulfilled by optimizing the parameter values of the model using a pre-defined task. SSL generally consists of two phases: (1) Pre-training (or Pretext) and (2) Downstream (or target). The representations learned by the encoder in the pre-training phase are used in the downstream task in the form of transferred weights. This description gives a hint of the similarity between SSL and Transfer Learning (TL). Although the two look similar, SSL and TL have distinct differences. The major dissimilarities between SSL and Transfer Learning (TL) can be summarised in a few words as given below.

The unlabelled dataset in the pre-training task of SSL is the same as the one used in the downstream task, whereas, in TL the pre-trained weights are obtained by training on a more diverse and large-scale annotated dataset, which differs from the one used in the downstream task too. Having a different dataset for the pre-training and the downstream task often becomes problematic, as contemporary deep learning methods require a large number of samples to yield results close to human proficiency in the target task. Researchers often adopt methods like selective fine-tuning of layers to prevent overfitting on small-scale datasets in the downstream task. However, the performance may suffer due to the destruction of the co-adaptation of the weights between the consecutive layers or from representation specificity \cite{howtran}. The self-supervised learning frameworks aim to overcome this issue by pre-training on the target dataset itself. As explored in \cite{howtran}, fine-tuning the target dataset may improve performance, as the representation specificity no longer hampers the optimization process. The pre-training phase learns weights that provide a better initialization point for the target task as the higher-level representations become more correlated or specific to the features on the target dataset. \\
\indent
In the past, several innovative approaches were proposed for the pre-training tasks which paved the way for efficient self-supervised representation learning. Both contrastive and non-contrastive learning approaches have succeeded in achieving state-of-the-art results on benchmark datasets. However, the ability to yield state-of-the-art performance with less annotated data requires pre-training with large models \cite{simclrv2}. In addition to mapping feature vectors of samples from the same class or samples with similar features close to each other, self-supervised contrastive algorithms also map unlabeled dissimilar samples farther away from each other in the latent space. This characteristic of contrastive learning acts to prevent the collapse of representation in the latent space, as is often the issue in this type of learning. \\
\indent
In this work, we propose a novel loss function based on contrastive learning. 
We adopt a bottom-up approach in constructing the proposed loss function. We initially adopted a pairwise binary contrastive learning approach (MIOv1) from which we eliminated the positive-positive repulsion to obtain its modified form, MIOv2.
To further improve the performance we take an upper bound of the repulsion term in MIOv2 to increase the repulsion between the samples constituting the negative pairs. Consequently, we obtain our proposed loss MIOv3.

We analytically show that the difference in the mutual information between the negative pairs and the positive pairs forms the lower bound of the proposed loss. In addition to that, we also attempt to analytically understand the conditions under which the SSL methods can achieve convergence. To do this, we first calculate the hessian of the loss with respect to the parameters and prove that the function approximated by the neural network has a Lipschitz continuous gradient under finite parameter assumption. We then apply this knowledge to figure out the convergence criterion using an approximation to the Polyak-Lojasiewicz inequality.

The primary contributions of this work can be summarized as follows:
 \begin{itemize}
     \item We propose a novel loss function for contrastive self-supervised learning by modeling the pre-training task as a binary classification problem.
     \item We compare the performance of the proposed algorithm to the state-of-the-art (SOTA) self-supervised learning algorithms under the constraints of limited training periods on the task of image classification. The proposed method outperforms the SOTA methods in most cases.
     \item We show analytically that the proposed loss optimizes the mutual information in both positive and negative pairs.
     \item We present an analysis of the Hessian of the function approximated by the model parameters. We further prove that the function has a Lipschitz continuous gradient, subject to the satisfiability of a specific constraint. To the best of our knowledge, such Hessian-based analyses have never been explored for SSL tasks.
     \item We also show  here empirically that SSL methods do not converge to minimizers. We show that contrastive self-supervised learning frameworks converge to a strict saddle point for limited training.
 \end{itemize}

 The rest of the paper is organized as follows. In \Cref{sec:litsurvey}, we give a brief overview of related work done in the recent past. \Cref{sec:methodology} describes the proposed methodology. At first, it describes the base loss function and shows the relation between mutual information and the proposed framework. This section also describes the step-by-step process of how we arrived at the proposed loss function. The section ends with a convergence analysis of self-supervised learning frameworks.  
 In \Cref{sec:expts}, we discuss the details of the experimental configurations that are used to establish the proof of concept. This section also analyzes the performance of the proposed loss function and compares it with the other existing self-supervised algorithms. 
 In \Cref{sec:ablation_studies}, we further extend our analysis 
 to show the effect of decreasing the number of parameters in an SSL model.
 Finally, \Cref{sec:conc} concludes the paper.


\section{Literature Survey}
\label{sec:litsurvey}

\textbf{Self-Supervised Learning}
During the initial days of self-supervised learning, a lot of techniques are designed based on handcrafted pre-training tasks, which are also known as pretext tasks. These handcrafted tasks include geometric transformation prediction \cite{videogeotr,rotnet, videorotnet}, context prediction \cite{contextpred,contextenc}, jigsaw puzzle solving \cite{noroozi, videojig, iterreorg, damagedjig}, temporal order related tasks for videos \cite{temporder, sssptemporder, cliporder, skipclip, shuffleandlearn}, pace prediction in videos \cite{pacepred}, image colorization \cite{imgcolor}, etc. These pretext tasks are aimed at learning representations that are invariant to transformations, context, etc. Although these tasks successfully rolled the wheels of self-supervised learning, the performances of the models pre-trained with these tasks are not at par with their supervised counterparts on the target tasks.

\indent Recently, several algorithms like SimCLR \cite{simclr}, MoCov1 \cite{moco}, MoCov2 \cite{mocov2}, BYOL \cite{byol}, SimSiam \cite{simsiam}, Barlow Twins \cite{barlow}, DCL/DCLW \cite{dcl} and VICReg \cite{vicreg} have emerged throughout the last few years as SSL techniques that do not require explicit pretext tasks. Some of these algorithms like SimCLR \cite{simclr}, MoCov1 \cite{moco}, MoCov2 \cite{mocov2}, and DCL/DCLW \cite{dcl} are based on the contrastive learning principle, while others like SimSiam \cite{simsiam}, BYOL \cite{byol}, Barlow Twins \cite{barlow}, VICReg \cite{vicreg} use non-contrastive loss functions to learn representations from the data. 

Self-supervised contrastive learning (SSCL) treats each data point as a separate class. Thus, a pair made of any two samples constitutes a negative pair, and a positive pair of samples is obtained by pairing two augmented versions of the same sample \cite{cpc, moco, simclr}. Recently, most of the SSCL-based techniques have been designed by optimizing the InfoNCE \cite{cpc} loss function. 
InfoNCE loss in contrastive learning is the same as the categorical cross-entropy loss, but the cosine similarity values between the samples in a pair are treated as logit values. Thus, InfoNCE loss can be considered the negative logarithm of the probability of predicting a positive pair.
The main principle behind this learning strategy is to learn an approximate function that maps the feature vector of similar data points closer and dissimilar data points far away. The quality of representation learned by the self-supervised model is generally evaluated from the model's performance on a $k$NN classification task using $k = 200$. 
Recently the researchers have proposed a framework in \cite{moco, mocov2}, that uses two networks (online and target) in the pre-training phase. The target network is momentum updated using the online network parameters to simulate a slow learning network. It also uses a memory bank to considerably increase the batch size, which proves useful in self-supervised contrastive learning by preventing representational collapse. SimCLR \cite{simclr} uses large batch sizes along with sample pairing to increase the number of negative pairs in a single batch without using any memory bank. Both MoCo and SimCLR frameworks use an encoder and a non-linear multi-layered perceptron (MLP) called a projector during the pre-training phase. DCL/DCLW \cite{dcl} is a recent improvement over contrastive learning frameworks. The authors showed that the performance of the SSL models can be improved by decoupling the positive and negative coupling introduced by the positive pair-related term in the denominator of the InfoNCE loss function.

In non-contrastive algorithms like SimSiam \cite{simsiam}, BYOL \cite{byol}, Barlow Twins \cite{barlow}, or VICReg \cite{vicreg}, the authors use only the positive pairs for self-supervised representation learning. BYOL \cite{byol} optimizes the mean squared error between the feature vectors of the two augmented versions of a sample constituting the positive pair to ensure the invariance of representations. SimSiam \cite{simsiam} optimizes the negative of the cosine similarity between two samples in a positive pair. The loss function of SimSiam and BYOL is essentially the same. However, SimSiam does not use a momentum encoder like BYOL to enforce variation in the positive pair $(x_1,x_2)$. Instead, it uses a method called stop-gradient which prevents the back-propagation of the gradient for the projected feature vector $p_1$ (output taken from projector MLP) of a sample in the positive pair. The flow of gradient occurs only for the predicted feature vector $z_2$ (output taken from the predictor MLP) of the other sample in the positive pair. In other words, the projected feature vector $p_1$ is treated as a non-differentiable constant vector detached from the computational graph. Barlow Twins \cite{barlow} minimizes the cross-correlation between any two feature dimensions under the assumption that each feature dimension is normally distributed. The VICReg \cite{vicreg} framework aims at minimizing the variance of each feature dimension to stay above a pre-defined threshold value, along with decorrelating any two feature dimensions by diagonalizing the cross-covariance matrix to prevent information collapse. The VICReg framework also uses an additional invariance term that minimizes the distance between features of the samples in a positive pair. Recently, methods like ZeroCL \cite{Zhang2022zerocl}, DINO \cite{caron2021dino}, WMSE \cite{ermolov2021wmse}, and ARB \cite{zhang2022arb} have emerged. ZeroCL and WMSE both use a spectral decomposition stage to apply a whitening transformation to the features. This step increases the computational complexity of the pre-training algorithm. Although ZeroCL \cite{Zhang2022zerocl} and WMSE \cite{ermolov2021wmse} claim to be negative-free frameworks, both methods use negative samples to compute batch statistics. ARB \cite{zhang2022arb} builds on Barlow twins \cite{barlow} and uses an optimization objective based on the nearest orthonormal basis. However, it uses a spectral decomposition step to deal with non-full rank matrices, which is also computationally expensive. On the other hand, DINO uses a self-distillation framework to learn representations without using a contrastive-based framework.



In this work, we present a new perspective on contrastive learning where we propose a novel SSL framework. While most previous works take an information theoretic or empirical approach to understand the working principle behind their respective frameworks, we take a novel approach by analyzing the Hessian spectrum and the Lipschitz continuity to understand the extent of convergence of our proposed and other contemporary SSL methods. In the subsequent sections, we will see a step-by-step breakdown of the analysis of the aforementioned contributions of this work.

\section{Methodology}
\label{sec:methodology}

In this section, we propose a novel loss function for contrastive learning. First, we will discuss the motivation and the base loss function from which we derive our proposed loss function in \Cref{subsec:proploss}. Then, we will discuss the modifications and the reason behind those to explain how we arrived at the proposed loss function in the subsequent subsections. 

For the analysis, we consider the self-supervised model consisting of an encoder and a non-linear projector. Let the input, encoder, encoder output (projector input), projector, and the final feature vector (output from the projector) be denoted by $x$, $f$,  $h$, $g$, and $z$, respectively. The input images $x \in \mathbb{R}^H \times \mathbb{R}^W \times \mathbb{R}^C$ when pass through the encoder $f$, a latent vector $h \in \mathbb{R}^F$ is obtained. This latent vector $h$ gives the final feature vector $z \in \mathbb{R}^D$ when passed through the projector $g$. The proposed loss function 
takes the feature vectors and outputs a scalar. Furthermore, let us denote the parameters of the encoder $f$ by $\theta$ and that of the projector $g$ by $\psi$.

To understand the flow of information we can devise the following equations
\begin{equation}
    z = g_{\psi}(h) = g_{\psi}(f_{\theta}(x))
\end{equation}
\subsection{Motivation behind MIOv1 loss function}
\label{subsec:proploss}

To understand the motivation behind the proposed loss function, let us reiterate the working principle behind contrastive learning. The primary objective of the contrastive learning algorithm is to learn an approximate mapping function that maps the features of the augmented versions of a sample close to each other. For samples belonging to different classes, the feature vectors are mapped as far as possible from each other. The primary motivation of our work is based on the fact that there are only two types of pairs in contrastive learning: positive and negative. Hence, the contrastive learning principle can be seen as optimizing the distance between any two samples in the feature space. In this work, we morph the contrastive learning scenario into a binary classification problem where a pair of samples is classified either as positive and pulled closer or as negative and pushed apart. To formulate the required objective function, we follow Wu et al. \cite{zhirong2018instdisc} in constructing a non-parametric sigmoid classifier. Hence, intuitively the base loss function named MIOv1 can be defined as given below:
\begin{equation}
     \centering
     \label{eqn:eqn13}
     \begin{split}
         \mathcal{L}_{v1} 
         = & -\mathop{\mathbb{E}}_{(x_i,x_j) \sim p_{+}}\left[\ln \left ( \frac{1}{1+e^{-\frac{C_{i,j}}{\tau}}} \right ) \right] -\mathop{\mathbb{E}}_{(x_k,x_l) \sim p_{-}}\left[\ln \left(1 - \frac{1}{1+e^{-\frac{C_{k,l}}{\tau}}} \right ) \right]\\
     \end{split}
 \end{equation}
where $C_{i,j}$ is the cosine similarity between two feature vectors $z_i$ and $z_j$ obtained by passing $x_i$ and $x_j$ through the encoder and the projector. $p_{+}$ and $p_{-}$ are the distribution of positive pairs and negative pairs on $\mathbb{R}^n \times \mathbb{R}^n$, respectively and $\tau$ is the temperature parameter. 

Considering $\mathcal{X}_+$ and $\mathcal{X}_-$ as the sets of positive and negative pairs sampled from the distributions of positive and negative pairs, $p_+$ and $p_-$, respectively, we can rewrite $\mathcal{L}_{v1}$ as,
\begin{equation}
     \centering
     \label{eqn:eqn13b}
     \begin{split}
         \mathcal{L}_{v1} 
        = & - \frac{1}{T_P} \sum_{(x_i,x_j)\in \mathcal{X}_{+}} \ln \left ( \frac{1}{1+e^{-\frac{C_{i,j}}{\tau}}} \right ) - \frac{1}{T_N} \sum_{(x_k,x_l)\in \mathcal{X}_{-}} \ln \left ( 1 - \frac{1}{1+e^{-\frac{C_{k,l}}{\tau}}} \right ) \\
     \end{split}
 \end{equation}

We follow the same sampling procedure as in SimCLR \cite{simclr}. Taking a batch size of $N$, we augment each sample in the batch to obtain two augmented samples from each sample, forming $N$ pairs and $2N$ samples in total. We can form $4N^2$ pairs in total, out of which $2N$ are positive pairs and $2N$ are self-pairs, and these $4N$ pairs will not contribute to the contrastive repulsion. Thus, the total number of negative pairs that can be formed is $4N^2 - 4N$. The MIOv1 loss can be expressed as follows:
\begin{equation}
     \centering
     \label{eqn:eqn14}
     \begin{split}
         \mathcal{L}_{v1} 
         = &-\frac{1}{T_P}\sum_{n = 1}^N \left[\ln \left(\frac{1}{1+e^{-\frac{C_{n,n'}}{\tau}}}\right) + \ln \left(\frac{1}{1+e^{-\frac{C_{n',n}}{\tau}}}\right)\right] - \frac{1}{T_N}\sum_{n = 1}^{2N} \sum_{\substack{m=1\\m\neq n,n'}}^{2N}{\ln \left(1 - \frac{1}{1+e^{-\frac{C_{n,m}}{\tau}}}\right)}\\
         =& -\frac{1}{N}\sum_{n = 1}^N \ln \left(\frac{1}{1+e^{-\frac{C_{n,n'}}{\tau}}}\right)  - \frac{1}{T_N}\sum_{n = 1}^{2N} \sum_{\substack{m=1\\m\neq n,n'}}^{2N}{\ln \left(1 - \frac{1}{1+e^{-\frac{C_{n,m}}{\tau}}}\right)}\\
     \end{split}
 \end{equation}
 
 
 \noindent where $n' = n+N$, $T_P = 2N$ and $T_N = 4N^2 - 4N$. We can deduce a small relation between $T_P$ and $T_N$ which can be stated as $T_N = T_P^2 - 2T_P$. 
 An illustrative example of how we obtain the sets of positive and negative pairs of samples is provided in Section 1 of the Supplementary.

\subsection{Effect of Removing a Positive-Positive Repulsion}
\label{subsec:miov2}

\begin{figure*}[!ht]
    \centering
    \subfloat[]{\label{fig:unifvstemp}\includegraphics[width = 0.4\linewidth]{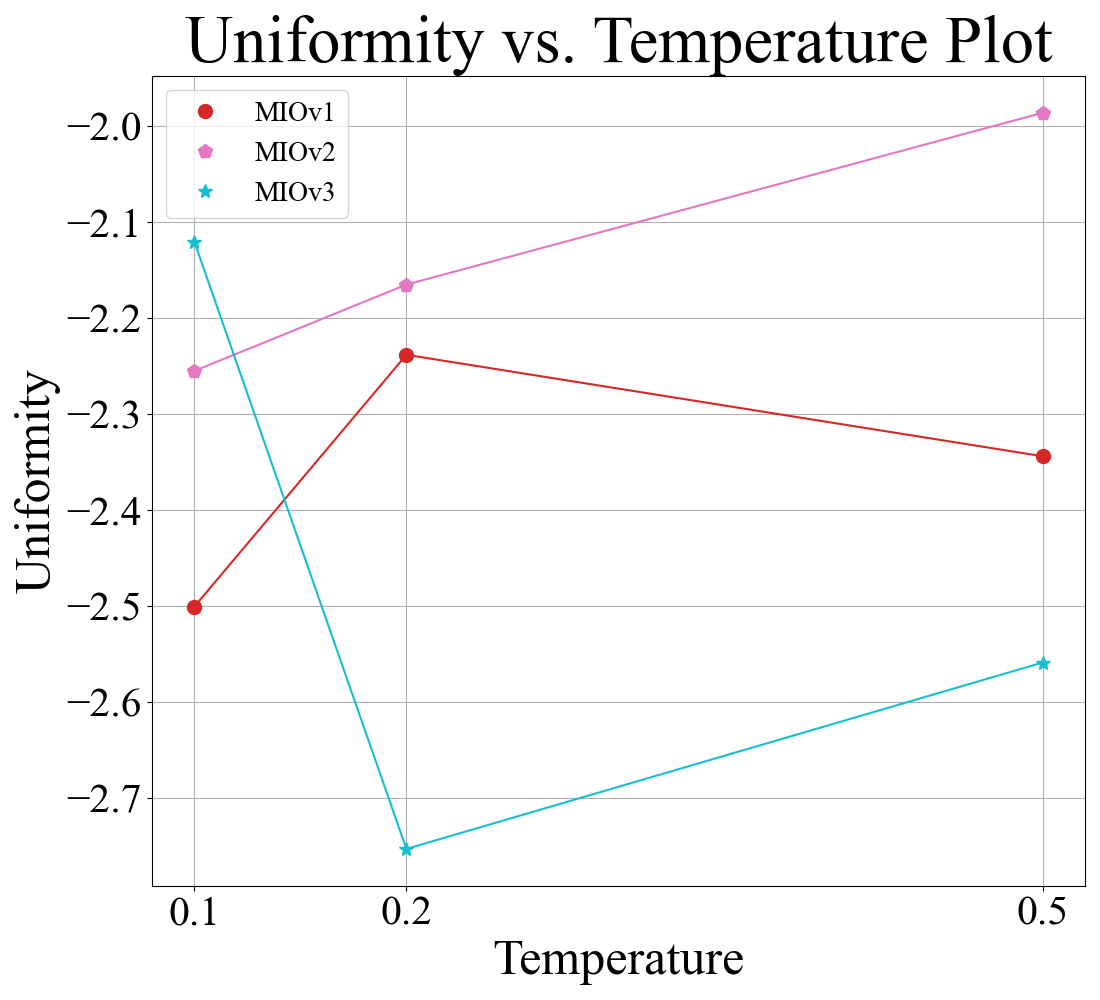}} 
    \subfloat[]{\label{fig:algnvstemp}\includegraphics[width = 0.4\linewidth]{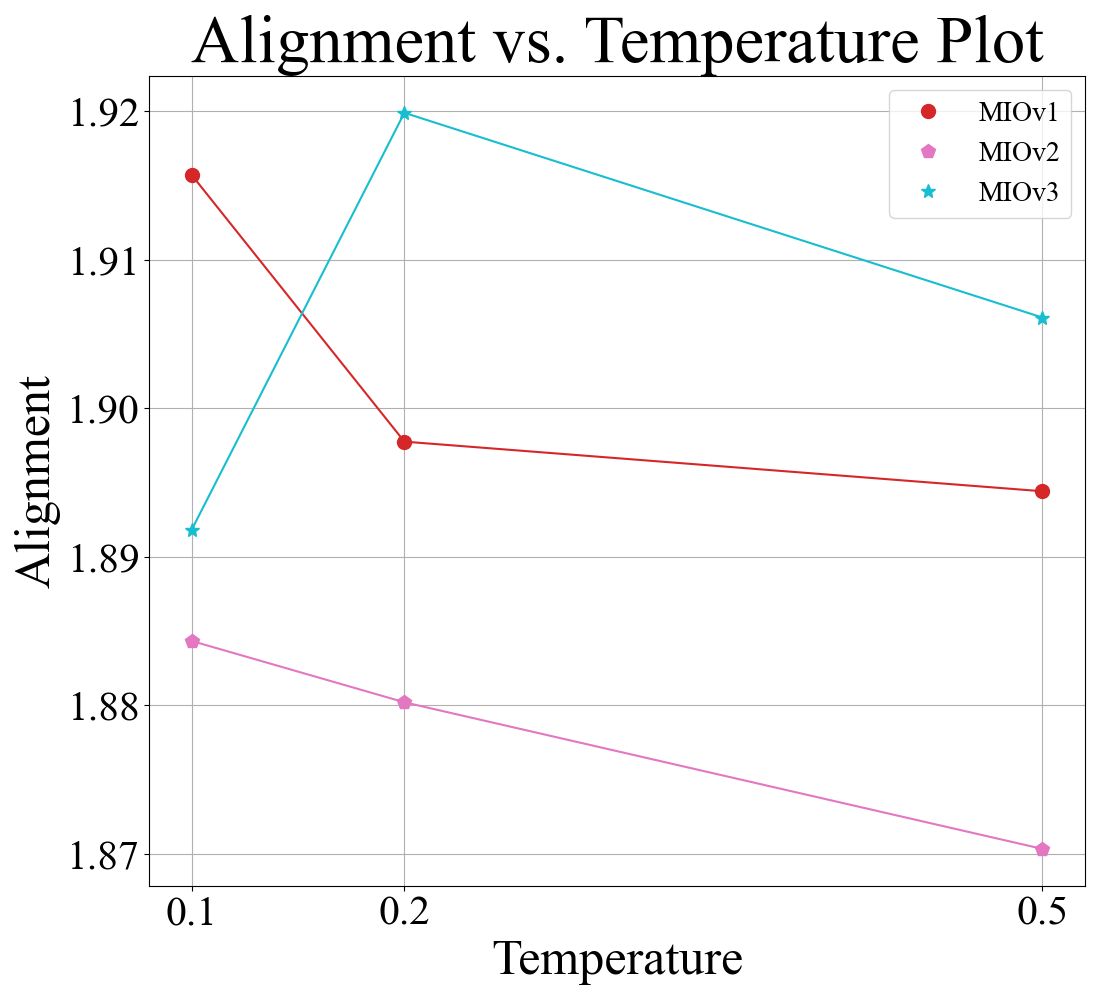}} \\
    \subfloat[]{\label{fig:icunifvstemp}\includegraphics[width = 0.4\linewidth]{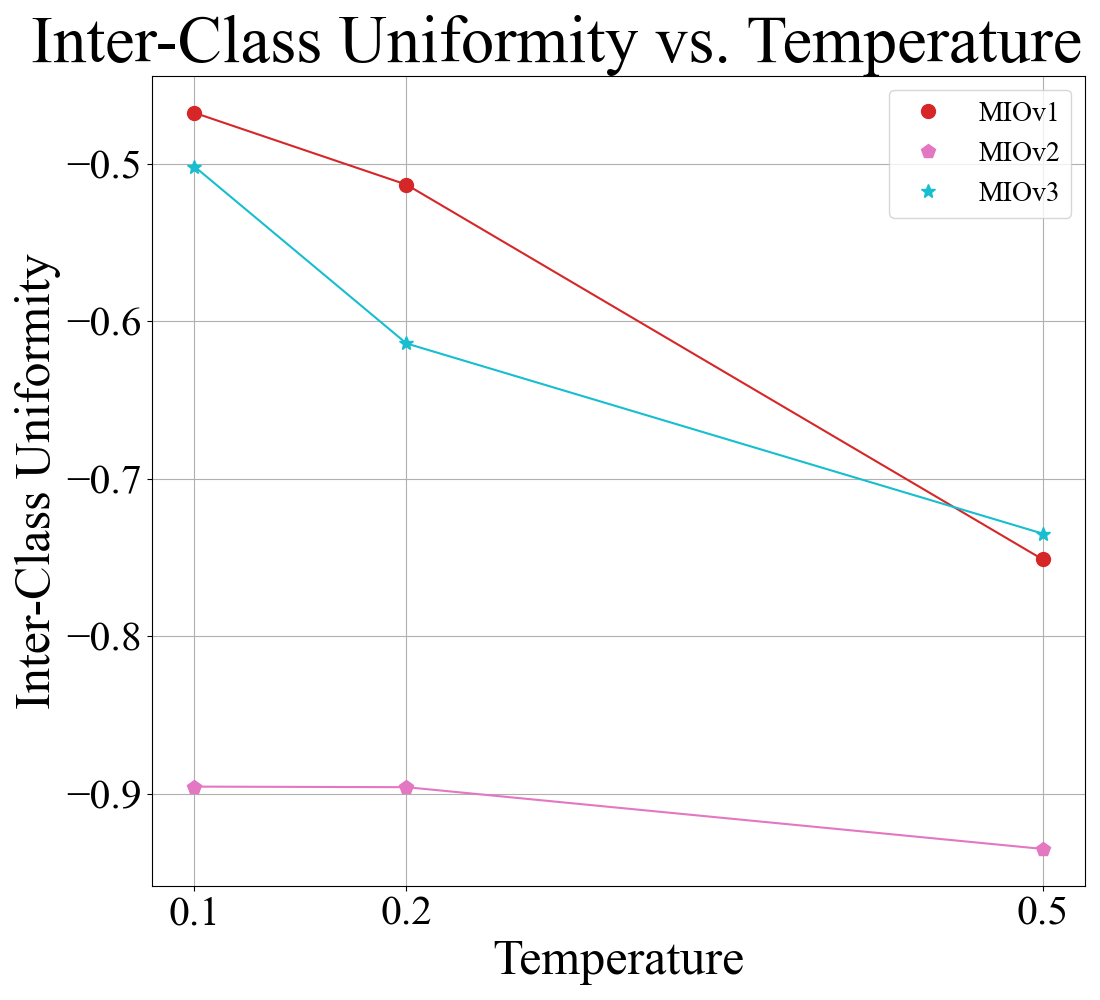}} 
    \subfloat[]{\label{fig:accvstemp}\includegraphics[width = 0.4\linewidth]{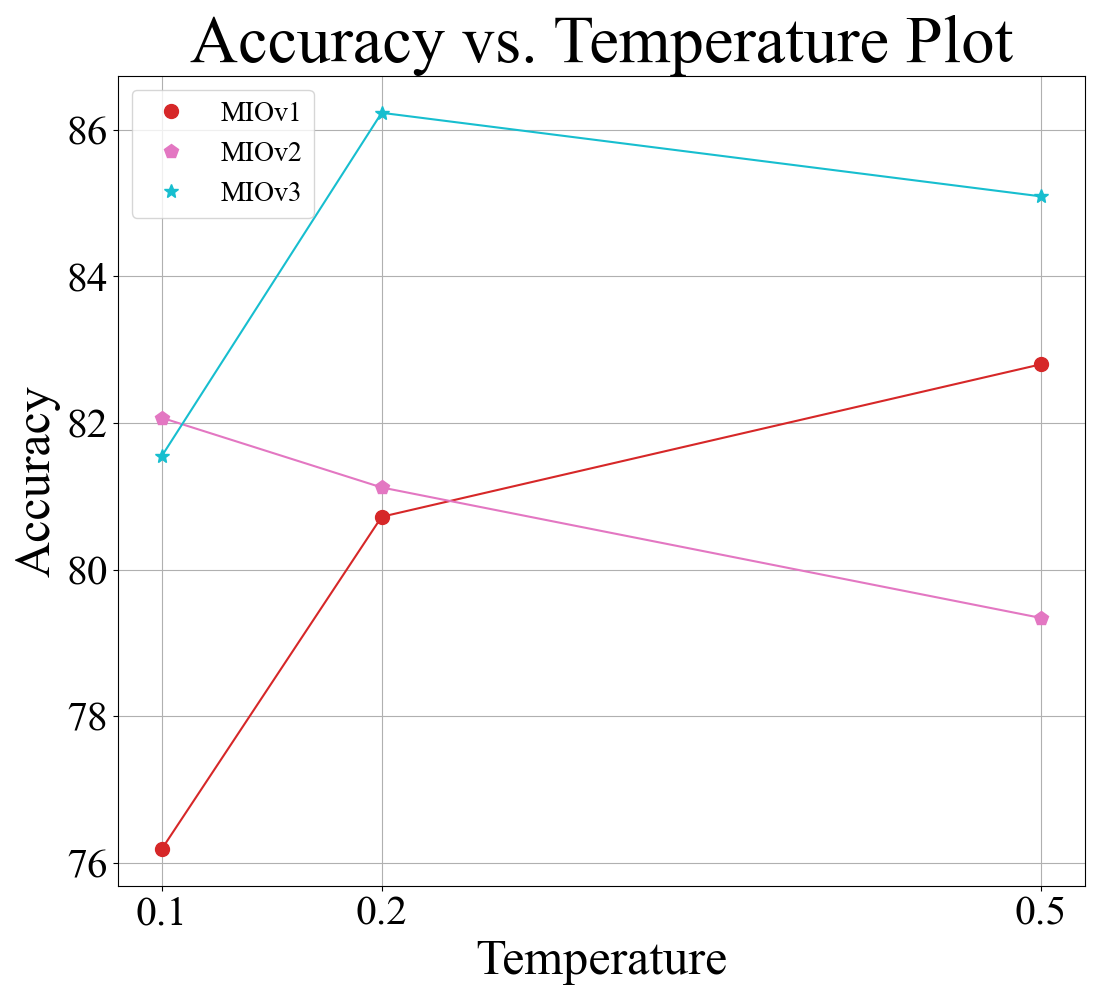}} 
    \caption{(a) Uniformity vs. Temperature, (b) Alignment vs. Temperature plot (c) Inter-class Uniformity vs Temperature, and (d) Accuracy vs Temperature plot at temperatures $\tau \in \{ 0.1, 0.2, 0.5\}$ for MIOv1, MIOv2 and MIOv3 on the CIFAR10 dataset.}
    \label{fig:v1vsv2vsv3stats}
\end{figure*}

We can expand \Cref{eqn:eqn14}, to get,
\begin{equation*}
     \centering
     \begin{split}
         \mathcal{L}_{v1}
        = & -\frac{1}{N}\sum_{n = 1}^N \ln \left(\frac{1}{1+e^{-\frac{C_{n,n'}}{\tau}}}\right) - \frac{1}{T_N}\sum_{n = 1}^{2N} \sum_{\substack{m=1\\m\neq n,n'}}^{2N}{\ln \left(1 - \frac{1}{1+e^{-\frac{C_{n,m}}{\tau}}}\right)}\\
\end{split}
 \end{equation*}
 \begin{equation}
     \centering
     \label{eqn:miov2e1}
     \begin{split}
        = & -\frac{1}{N}\sum_{n = 1}^N{\frac{C_{n,n'}}{\tau}} +\frac{1}{N}\sum_{n = 1}^N{\ln \left(1+e^{\frac{C_{n,n'}}{\tau}}\right)} + \frac{1}{T_N}\sum_{n = 1}^{2N} \sum_{\substack{m=1\\m\neq n,n'}}^{2N}{\ln \left({1+e^{\frac{C_{n,m}}{\tau}}}\right)}\\
     \end{split}
 \end{equation}

 where $n = n+N$ and $T_N$ bear the same meaning as in \Cref{eqn:eqn14}. In the \Cref{eqn:miov2e1}, we see that minimizing the loss $\mathcal{L}_{v1}$, minimizes the second term. This means that the terms $C_{n,n'}$ and $C_{n,m}$ are also minimized. However, $C_{n,n'}$ being the cosine similarity of the samples in a positive pair should be maximized to $+1$. We see that a repulsive force will take effect on the samples in the positive pair due to the minimization of the second term in the last line of Equation \ref{eqn:miov2e1}. Elimination of this repulsive force should improve the performance and result in faster convergence in the optimization process. To this end, we arrive at our second loss as mentioned in Equation \ref{eqn:miov2e2}, which we term as MIOv2.

\begin{equation}
     \centering
     \label{eqn:miov2e2}
     \begin{split}
         \mathcal{L}_{v2}
        = & -\frac{1}{N}\sum_{n = 1}^N{\frac{C_{n,n'}}{\tau}} + \frac{1}{T_N}\sum_{n = 1}^{2N} \sum_{\substack{m=1\\m\neq n,n'}}^{2N}{\ln \left({1+e^{\frac{C_{n,m}}{\tau}}}\right)}\\
     \end{split}
 \end{equation}
 
 The above equation can also be written as, 
 \begin{equation}
     \centering
     \label{eqn:miov2e2b}
     \begin{split}
         \mathcal{L}_{v2}
        = & - \mathbb{E}_{(x_i,x_j)\sim p_{+}} \left[\frac{C_{i,j}}{\tau}\right] + \mathbb{E}_{(x_k,x_l)\sim p_{-}} \left[\ln \left(1+e^{\frac{C_{k,l}}{\tau}}\right)\right]\\
     \end{split}
 \end{equation}

 where $p_+$ and $p_-$ denote the same quantities as in \Cref{eqn:eqn14}. 


To analyze the relation between the loss functions, we will first expand the expression for $\mathcal{L}_{v1}$, as follows,

\begin{equation}
    \centering
    \begin{split}
        \mathcal{L}_{v1} 
         = & -\mathop{\mathbb{E}}_{(x_i,x_j) \sim p_{+}}\left[ln \left ( \frac{1}{1+e^{-\frac{C_{i,j}}{\tau}}} \right ) \right] -\mathop{\mathbb{E}}_{(x_k,x_l) \sim p_{-}}\left[ln \left(1 - \frac{1}{1+e^{-\frac{C_{k,l}}{\tau}}} \right ) \right]\\
         = & - \mathbb{E}_{(x_i,x_j)\sim p_{+}} \left[\frac{C_{i,j}}{\tau}\right] + \mathbb{E}_{(x_i,x_j)\sim p_{+}} \left[ln \left(1+e^{\frac{C_{i,j}}{\tau}}\right)\right] + \mathbb{E}_{(x_k,x_l)\sim p_{-}} \left[ln \left(1+e^{\frac{C_{k,l}}{\tau}}\right)\right]\\
          = & \mathcal{L}_{v2} + \mathbb{E}_{(x_i,x_j)\sim p_{+}} \left[ln \left(1+e^{\frac{C_{i,j}}{\tau}}\right)\right]\\
    \end{split}
    \label{eqn:penpersLv1exp}
\end{equation}

 Now, to analyze the phenomenon behind the difference in performance between MIOv1 and MIOv2 at different temperatures, we will look at how the loss functions behave.

 Let us consider two cases, (1) $C_{ij} > 0$, and (2) $C_{ij} \leq 0$. Without loss of generality, we can assume that $\tau > 0$. Minimizing the term $\mathbb{E}_{(x_i,x_j)\sim p_{+}} \left[ln \left(1+e^{\frac{C_{i,j}}{\tau}}\right)\right]$ of Eqn. \ref{eqn:penpersLv1exp} increases the repulsion between the samples constituting the positive pair. We will use the notation $\mathcal{R}_{pp}$ to denote this term from here onwards. Now, for Case (1), as temperature $\tau$ increases, the magnitude of $\ln(1+e^{\frac{C_{ij}}{\tau}})$ decreases. Hence, $\mathcal{R}_{pp}$ decreases, and the repulsive force acting on the samples in the positive pairs is reduced. From \Cref{fig:v1vsv2vsv3stats}, we can observe that for both MIOv1 and MIOv2, as the temperature increases initially (from $\tau = 0.1$ to $\tau = 0.2$) the increase in alignment and decrease in inter-class uniformity indicates that the samples in each cluster move close to each other, and hence the rise in uniformity. This is primarily due to the effect of increasing temperature on $\mathcal{R}_{pp}$ in MIOv1. 
 However, when $\tau$ decreases, the magnitude of $\ln(1+e^{\frac{C_{ij}}{\tau}})$ increases, consequently it increases $\mathcal{R}_{pp}$ as well as the repulsion between the samples in the positive pairs. 
 This effect is detrimental to the performance, as the samples in the positive pairs are mapped far apart. 
 For Case (2), the variation of $\mathcal{R}_{pp}$ with temperature will be inverted, that is, with decreasing temperature, the value of $\mathcal{R}_{pp}$ will decrease, and vice versa. 

Without $\mathcal{R}_{pp}$ in MIOv2, the repulsion between the samples in the positive pair vanishes. Hence, intuitively MIOv2 should optimize better than MIOv1. At low temperatures, the magnitude of $\mathcal{R}_{pp}$ in MIOv1 increases, preventing samples in positive pairs from being mapped close to each other. 
However, at high temperatures, the magnitude of $\mathcal{R}_{pp}$ decreases. Consequently, the difference between MIOv1 and MIOv2 is reduced. In some cases, as empirically observed, MIOv1 outperforms MIOv2 at higher temperatures. This is primarily due to the absence of $\mathcal{R}_{pp}$.  
 
 Thus, without $\mathcal{R}_{pp}$ (MIOv2), the parameters are better optimized at lower temperatures than with $\mathcal{R}_{pp}$ (MIOv1), whereas the reverse is true at higher temperatures, as evident from Table \ref{tab:tempabltab}.

\subsection{Does optimizing an upper bound of negative pair repulsion result in better representation learning?}
\label{subsec:miov3}

In MIOv2, we eliminated the positive-positive repulsion. One notable issue with lower temperatures is the instability that it can bring along as the magnitude of the gradients also increases. To maintain stability we need to tread at higher temperatures. However, we also need to maintain uniformity at higher temperatures.
However, to further improve performance without positive-negative pair coupling, we start by looking at the second term of MIOv2, that is, $\mathbb{E}_{(x_k,x_l)\sim p_{-}} \left[\ln \left(1+e^{\frac{C_{k,l}}{\tau}}\right)\right]$. We denote this term by $\mathcal{R}_{nn}$. To improve uniformity we need to increase the repulsion between samples in negative pairs further. We achieve this by incorporating the upper bound of the term mentioned above in MIOv2, resulting in our final proposed loss function, MIOv3.

To arrive at our final loss function, we follow some mathematically justifiable steps. Using Mean Value Theorem \cite{mvtbook}, there exists $\xi \in (0, x)$, such that,
\begin{equation}
    \centering
    \begin{split}
        ln(1 + x) = ln(1+x) - ln(1) & = x \cdot \left[\frac{\partial ln(1+x)}{\partial x} \right]_{x = \xi}\\ &= x \cdot \frac{1}{1+\xi} \leq x \\
    \end{split}
\end{equation}

Using the above relation in $ln\left(1+e^{\frac{C_{k,l}}{\tau}}\right)$ from \Cref{eqn:miov2e2b}, we get,
\begin{equation}
    \centering
    \begin{split}
        ln\left(1+e^{\frac{C_{k,l}}{\tau}}\right) \leq e^{\frac{C_{k,l}}{\tau}}\\
    \end{split}
\end{equation}

Replacing $ln\left(1+e^{\frac{C_{k,l}}{\tau}}\right)$ by $e^{\frac{C_{k,l}}{\tau}}$ in \Cref{eqn:miov2e2b}, we get, 
\begin{equation}
\label{eqn:lv3_1}
    \centering
    \begin{split}
        \mathcal{L}_{v3}
        = & - \mathbb{E}_{(x_i,x_j)\sim p_{+}} \left[\frac{C_{i,j}}{\tau}\right] + \mathbb{E}_{(x_k,x_l)\sim p_{-}} \left[e^{\frac{C_{k,l}}{\tau}}\right]\\
        = & - \sum_{(x_i,x_j)\in \mathcal{X}_{+}} \left[\frac{C_{i,j}}{\tau}\right] + \sum_{(x_k,x_l)\in \mathcal{X}_{-}} \left[e^{\frac{C_{k,l}}{\tau}}\right]\\
    \end{split}
\end{equation}


We can rewrite the above equation as,
\begin{equation}
\label{eqn:lv3_2}
    \centering
    \begin{split}
        \mathcal{L}_{v3} = & - \frac{1}{N} \sum_{\substack{n=1}}^{N} \frac{C_{n,n'}}{\tau} + \frac{1}{T_N} \sum_{\substack{n=1}}^{2N} \sum_{\substack{m=1\\m \neq n,n'}}^{2N} e^{\frac{C_{n,m}}{\tau}} \\
    \end{split}
\end{equation}

where $n' = n+N$, $T_N = 2N(2N-2)$ for a batch of size $N$. $\mathcal{X}_+$ and $\mathcal{X}_-$ are sets of positive and negative pairs of samples obtained from the distribution of positive and negative pairs, $p_+$ and $p_-$, respectively. We call this version of the loss as MIOv3, that is, $\mathcal{L}_{v3}(g_{\psi}(f_{\theta}(X)))$, which is our final proposed loss function. Pre-training a ResNet50 (ResNet18) model on ImageNet100 (CIFAR10) using the same hyper-parameters configuration as MIOv1 and MIOv2.
We observe that the performance improves considerably and even surpasses the contemporary contrastive learning frameworks on ImageNet100 (CIFAR10).


In Sec. \ref{subsec:miov2}, we already discussed the two cases, for which analyzed the behavior of MIOv1 and MIOv2. with increasing temperature $\tau$, the magnitude of the term $\mathcal{R}_{nn}$ decreases or increases depending on the cosine similarity of the samples in the concerned pair.

Let us denote the upper bound of $\mathcal{R}_{nn}$ by $\mathcal{O}_{\mathcal{R}nn}$. Now, using $\mathcal{O}_{\mathcal{R}nn}$ in place of $\mathcal{R}_{nn}$ increases the repulsion between the samples in the negative pairs. 
This effect helps in maintaining the uniformity of the samples, thereby preventing the collapse observed in MIOv2 at high temperatures. However, with the decrease in temperature, for the cases of $C_{ij} > 0$, $\mathcal{O}_{\mathcal{R}nn}$ grows faster than $\mathcal{R}_{nn}$. This results in an exponential increase in the repulsion between the false negative pairs. Consequently, alignment \cite{unifalign} of samples is hindered. For $C_{ij} < 0$, we observe that $\mathcal{O}_{\mathcal{R}nn} \rightarrow \mathcal{R}_{nn}$. Hence, the effect is not so evident in this case.

\color{black}
\subsection{Relation of Proposed loss and Mutual Information}
\label{subsec:relmi}

In this subsection, we are going to derive the relationship between the MIOv3 loss function and mutual information \cite{shannon, eit, flmmi} between the samples in a pair. The final expression of the lower bound of the MIOv3 loss function will allow us to visualize the optimization process intuitively. 

Let us define the class-conditional probabilities be as follows,

\begin{equation*}
\begin{split}
    p((z_i,z_j)|k=1) & = p_m((z_i,z_j);\theta) = P_+^{i,j}\\
    p((z_i,z_j)|k=0) & = p_n((z_i,z_j)) = P_-^{i,j}
\end{split}
\end{equation*}

Here, $P_{+}^{i,j}$ is the probability of obtaining the pair $(z_i, z_j)$ given the sample is drawn from the positive pair distribution, i.e. $k=1$, and $P_{-}^{i,j}$ is the probability of obtaining the pair $(z_i, z_j)$ given the sample is drawn from the positive pair distribution, i.e. $k=0$.






Now, the probability of the pair $(z_i, z_j)$ being a positive pair in a binary classification setting can be expressed as:
\begin{equation}
    \label{eqn:mi2}
    \centering
    \begin{split}
        P(k = 1 | (z_i, z_j)) & =  \frac{P(k=1)P_{+}^{i,j}}{P(k=1)P_{+}^{i,j} + P(k=0)P_{-}^{i,j}} = \frac{P_{+}^{i,j}}{P_{+}^{i,j} + P_{-}^{i,j}}\\
    \end{split}
\end{equation}

\noindent
where $P(k=1)$ and $P(k=0)$ are the class prior probabilities and $P(k=1) = P(k=0)$. The complete analysis behind the reason for considering $P(k=1) = P(k=0)$ is given in details in Supplementary Sec. 5.2.

\begin{figure}[ht!]
    \centering
    \includegraphics[width = 0.8\linewidth]{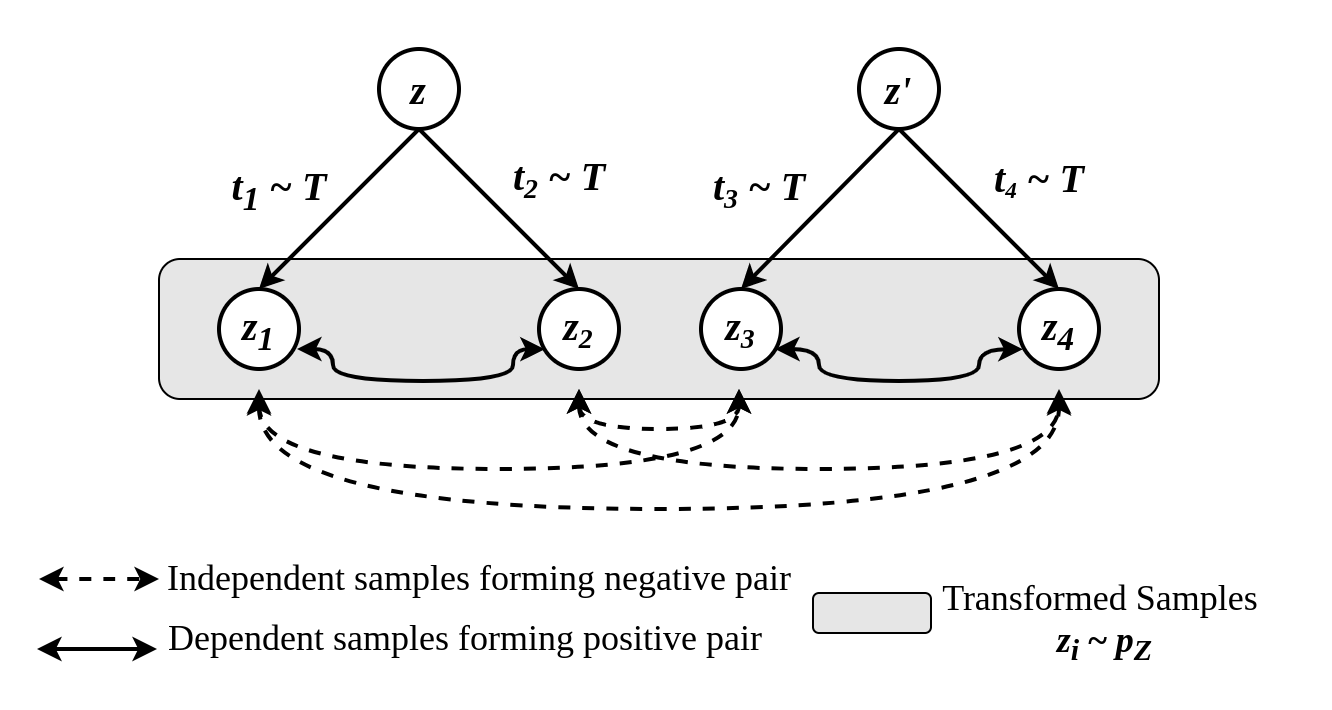}
    \caption{Graphical Model \cite{pgmbook} showing the dependency between two samples in a positive pair and the independency between two samples forming a negative pair. Here, $z$ and $z'$ are two different samples in a dataset. $t_1, t_2, t_3, t_4$ are randomly chosen transformations from the distribution $T$. $z_1$ and $z_2$ are obtained by applying $t_1$ and $t_2$ on $z$. $z_3$ and $z_4$ are obtained by applying $t_3$ and $t_4$ on $z'$.}
    \label{fig:sampldep}
\end{figure}

Considering $P_{Z}(z_i)$ as the probability of obtaining $z_i$ from the distribution $p_Z$ over all possible transformed samples of $z$ and $P_{Z,Z}(z_i,z_j)$ as the probability of obtaining $(z_i, z_j)$ from the joint distribution $p_{Z,Z}$, we deduce the following relations. When considering $(z_i, z_j)$ as a positive pair, the parent sample $z$ from which we obtain a positive pair is not observed. Hence, we cannot consider $z_i$ and $z_j$ as independent \cite{pgmbook}. In \Cref{fig:sampldep}, for example, the positive transformed pair $(z_1, z_2)$ is obtained from the same sample $z$. Thus, $P_{+}^{i,j}$ is equal to the probability $P_{Z,Z}(z_i, z_j)$. Again, when considering $(z_i, z_j)$ as a negative pair, there will be no dependency between the two samples, for example, $(z_1, z_3)$ or $(z_2, z_4)$ in \Cref{fig:sampldep}. Thus, $z_i$ and $z_j$ can be considered independent and $P_{-}^{i,j}$ can be considered as the product of $P_{Z}(z_i)$ and $P_{Z}(z_j)$.

\noindent
Therefore, using the same idea, \Cref{eqn:mi2} can be expanded as follows:
\begin{equation}
    \label{eqn:mi3}
    \centering
    \begin{split}
        P(k = 1 | (z_i, z_j)) & = \frac{p_{Z,Z}(z_i, z_j)}{p_{Z}(z_i)p_{Z}(z_j) + p_{Z,Z}(z_i, z_j)} = \frac{\frac{p_{Z,Z}(z_i, z_j)}{p_{Z}(z_i)p_{Z}(z_j)}}{1 + \frac{p_{Z,Z}(z_i, z_j)}{p_{Z}(z_i)p_{Z}(z_j)}}\\
    \end{split}
\end{equation}

\noindent
Let us define the scoring function 
\begin{equation}
\label{eqn:mi1}
    s(z_i, z_j) = e^{C_{i,j}}
\end{equation}
\noindent
where $C_{i,j}$ is the cosine similarity between $z_i$ and $z_j$.

\noindent
We can also express $P(k=1|(z_i, z_j))$ in terms of $s(z_i, z_j)$ as follows,
\begin{equation}
    \label{eqn:mi4}
    \centering
    \begin{split}
        P(k = 1 | (z_i, z_j)) & = \frac{1}{1+e^{-C_{i,j}}} = \frac{e^{C_{i,j}}}{1+e^{C_{i,j}}}=\frac{s(z_i, z_j)}{1 + s(z_i, z_j)}\\
    \end{split}
\end{equation}

\noindent
Thus, comparing \Cref{eqn:mi3} and (\ref{eqn:mi4}), we get,
\begin{equation}
\label{eqn:mi5}
    s(z_i, z_j) = e^{C_{i,j}} = \frac{p_{Z,Z}(z_i, z_j)}{p_{Z}(z_i)p_{Z}(z_j)}
\end{equation}

\noindent
Putting \Cref{eqn:mi5} in \Cref{eqn:lv3_2}, we get, 
 \begin{equation}
    \label{eqn:miov3_mi}
    \centering
    \begin{split}
        \mathcal{L}_{v3} = & - \mathop{\mathbb{E}}_{\substack{(x_i,x_j) \sim p_{+}}} \left[ log \left(\frac{p_{Z,Z}(z_i, z_j)}{p_{Z}(z_i)p_{Z}(z_j)} \right ) \right] + \mathop{\mathbb{E}}_{\substack{(x_k,x_l) \sim p_{-}}} \left[ \frac{p_{Z,Z}(z_k, z_l)}{p_{Z}(z_k)p_{Z}(z_l)} \right]\\
        = & - \mathop{\mathcal{I}}_{\substack{(z_i, z_j) \sim p_{+}}}(z_i, z_j) + \mathop{\mathbb{E}}_{\substack{(x_k,x_l) \sim p_{-}}} \left[ \frac{p_{Z,Z}(z_k, z_l)}{p_{Z}(z_k)p_{Z}(z_l)} \right]\\
        \geq & - \mathop{\mathcal{I}}_{\substack{(z_i, z_j) \sim p_{+}}}(z_i, z_j) + \mathop{\mathbb{E}}_{\substack{(x_k,x_l) \sim p_{-}}} \left[ log \left(1 + \frac{p_{Z,Z}(z_k, z_l)}{p_{Z}(z_k)p_{Z}(z_l)} \right ) \right]\\
        \geq & - \mathop{\mathcal{I}}_{\substack{(z_i, z_j) \sim p_{+}}}(z_i, z_j) + \mathop{\mathbb{E}}_{\substack{(x_k,x_l) \sim p_{-}}} \left[ log \left( \frac{p_{Z,Z}(z_k, z_l)}{p_{Z}(z_k)p_{Z}(z_l)} \right ) \right]\\
        \geq & - \mathop{\mathcal{I}}_{\substack{(z_i, z_j) \sim p_{+}}}(z_i, z_j) + \mathop{\mathcal{I}}_{\substack{(z_k, z_l) \sim p_{-}}}(z_k, z_l)\\
    \end{split}
\end{equation}

From \Cref{eqn:miov3_mi}, we can infer that the proposed loss function $\mathcal{L}_{v3}$ works by maximizing the mutual information between the samples in a positive pair $(z_i, z_j)$. It also minimizes the mutual information between the samples in a negative pair $(z_k, z_l)$. 

\subsection{Do contrastive SSL methods converge?}
\label{subsec:sslconvproof}

The loss landscape of the different models in the frameworks depends on the loss function used. The function $\mathcal{L}_{v3} \circ g_{\psi} \circ f_{\theta}$ is a non-convex function of the parameter space $\mathbb{P}$. The input pair space $\chi$ is mapped to the latent space $\mathbb{R}^D$ by a function $g_{\psi} \circ f_{\theta}$ or $(g \circ f)_{\mathcal{P}}$, where $\mathcal{P} = \{\theta, \psi\}$ denotes a point in the parameter space $\mathbb{P}$. The paired embedding obtained from the function $g_{\psi} \circ f_{\theta}$ or $(g \circ f)_{\mathcal{P}}$, in the self-supervised pretraining phase, constitutes a point in the embedding space $\mathcal{E}: \mathbb{R}^D \times \mathbb{R}^D$ and is mapped to the loss landscape $\mathbb{L}$, i.e., $\mathcal{L}_{v3} \circ (g \circ f)_{\mathcal{P}} : \chi \rightarrow \mathbb{L}$.

To analytically check if SSL pre-training truly converges, we need to proceed in three short steps. First, we need to calculate the Hessian of $\mathcal{L}_{v3}$ with respect to the parameters.  Without loss of generality, we show the Hessian of $\mathcal{L}_{v3}$ with respect to $\psi$ in Supplementary Sec. 2.2. Next, we need to check if $\mathcal{L}_{v3}$ has a $L$-Lipschitz continuous gradient with respect to the parameters. In Supplementary Sec. 2.3, we show that the norm of the Hessian matrix $\mathcal{H}$ is bounded by $L$ indirectly by showing that the composite function approximated by $\mathcal{L}_{v3} \circ g_{\psi} \circ f_{\theta}: \chi \rightarrow \mathbb{R}$ has a Lipschitz continuous gradient, under the constraint that $\sum_d h^{(d)}_{\theta n} < \infty$, and $\sum_{w \in \mathcal{P}} w < \infty$. 
Thus, we prove that $\mathcal{L}_{v3} \circ g_{\psi} \circ f_{\theta}$ belongs to a class of twice-differentiable continuous real-valued functions. 
Finally, in this section, we use the Polyak-Lojasiewicz (PL) inequality, defined in the local neighbourhood of the initialization point, to show that the SSL methods converge to local minima only under long pre-training. 

As the learning rate decreases, the conditions become more favorable for descent into a convex valley in the loss landscape. However, decreasing the learning rate deters the optimizer from proceeding with the same ease on flat plateaus or at inflection points to escape local minima. To ensure convergence along the steepest eigendirection, it is necessary to have a learning rate $\eta \leq \frac{1}{L} = \frac{1}{\lambda_{max}}$ \cite{optlsml}, where $L$ is the Lipschitz constant.


Following \cite{hamed2016linconv}, rewriting the Polyak-Lojasiewicz Inequality (See Section 4 in Supplementary for detailed discussion) in terms of loss function $\mathcal{L}$, for $\mu > 0$ the linear convergence rate is given by 
\begin{equation}
    \centering
    \label{eqn:plin1}
    \begin{split}
        \mathcal{L}(w_t) - \mathcal{L}^* \leq ( 1- \frac{\mu}{L})^t (\mathcal{L}(w_0) - \mathcal{L}^*)\\
    \end{split}
\end{equation}
\noindent
where $w_t, w_0$ are the parameter state at the $t^{th}$ and $0^{th}$ step. For gradient descent algorithm or minimization problems, $L > 0$ always, at the minimum, and $\mu < L$. 

In \citet{lee2016gdconvmin}, it is stated that a twice differentiable continuous function which is initialized randomly converges to a local minimum \textit{almost surely}. Thus, given a \textit{step size small enough}, we can derive the convergence rate in the local neighbourhood of the initialization point for the function approximated by the deep neural network. Therefore, as mentioned in \citet{hamed2016linconv}, we can analyze the convergence phase in terms of locally satisfying the PL inequality.

If we consider convergence along each eigendirection, we can calculate the expected convergence rate across the whole loss landscape. Considering a single eigendirection corresponding to maximum eigenvalue $\lambda_i$, the above equation reduces to,
\begin{equation}
    \centering
    \label{eqn:plin2}
    \begin{split}
        \mathcal{L}(w^i_t) - \mathcal{L}^*(w^i) \leq ( 1- \frac{\mu_i}{\lambda_i})^t (\mathcal{L}(w^i_0) - \mathcal{L}^*(w^i))\\
    \end{split}
\end{equation}
subject to the satisfiability of 
\begin{equation}
\label{eqn:plin3}
    \centering
    \begin{split}
        \frac{1}{2} \lVert \nabla \mathcal{L}^i(w) \rVert ^2 \geq  \mu_i (\mathcal{L}^i(w) - \mathcal{L}^*) \; \text{for} \; \mu_i > 0
    \end{split}
\end{equation}

We take an expectation over all the eigendirections to calculate a proxy for the linear convergence rate. 
\begin{equation}
    \centering
    \label{eqn:plin4}
    \begin{split}
        \mathbb{E}_i \left[ \mathcal{L}(w^i_t) - \mathcal{L}^*(w^i) \right] &\leq \mathbb{E}_i \left[ ( 1- \frac{\mu_i}{\lambda_i})^t (\mathcal{L}(w^i_0) - \mathcal{L}^*(w^i)) \right]\\
        &\leq \mathbb{E}_i \left[ ( 1- \frac{\mu_i}{\lambda_{max}})^t (\mathcal{L}(w^i_0) - \mathcal{L}^*(w^i)) \right]\\
        &\leq \mathbb{E}_i \left[ ( 1- \frac{\mu_i}{L})^t (\mathcal{L}(w^i_0) - \mathcal{L}^*(w^i)) \right]\\
        &\leq \mathbb{E}_i \left[ ( 1- \frac{\mu_{min}}{L})^t (\mathcal{L}(w^i_0) - \mathcal{L}^*(w^i)) \right]\\
        &\leq \mathbb{E}_i \left[c^t (\mathcal{L}(w^i_0) - \mathcal{L}^*(w^i)) \right]\leq \delta' < \infty\\
    \end{split}
\end{equation}
where $c^t = (1 - \frac{\mu_{min}}{L})^t \rightarrow 0 $ if $t \rightarrow \infty$, as $\mu_{min} \rightarrow 0^+$, and $\mu_i > 0 \; \forall \; i$.

From \Cref{eqn:plin4}, we can see that the convergence rate becomes infinitesimal for a large value of $t$, that is, for a long training process. However, we will look at Eqn. 34 of the Supplementary, where we derive an expression of the expected gradient norm, as 
\begin{equation}
    \centering
    \begin{split}
        \sum_{t=1}^T \left(\eta_t-\frac{\eta_t^2 L}{2} \right) \mathbb{E}_t \left[ \lVert \nabla \mathcal{L}^i(w) \rVert_2^2\right]
        &\leq \mathcal{L}(w_t)-\mathcal{L}^* +\frac{\sigma^2 L}{2} \sum_{k=1}^T \eta_t^2 ,\\
    \end{split}
\end{equation}
where we have assumed that the variance of the stochastic gradient is bounded above by $\sigma^2$.

Putting the expression for the proxy of the convergence rate in place of $\mathcal{L}(w_i) - \mathcal{L}^*$, we get, 
\begin{equation}
    \label{eqn:plin4b}
    \centering
    \begin{split}
    &\sum_{t=1}^T \left(\eta_t-\frac{\eta_t^2 L}{2} \right) \mathbb{E}_i \left[ \lVert \nabla \mathcal{L}^i(w) \rVert_2^2\right]\leq \mathbb{E}_i \left[ \mathcal{L}(w^i_t) - \mathcal{L}^*(w^i) \right] +\frac{\sigma^2 L}{2} \sum_{t=1}^T \eta_t^2 < \infty\\
    \end{split}
\end{equation}

where, $\eta_t$ varies with time as $\eta_t = \eta_{min} + \frac{1}{2}(\eta_{max} - \eta_{min})\left(1 + \cos\left( \frac{t}{T}\pi\right) \right)$ with $\eta_{min} = 0$, and $T$ is the total number of steps. From the expression of $\eta_t$, we have,
\begin{equation}
\label{eqn:pl106}
\begin{split}
    \sum_{t=1}^{\infty} \eta_t \to \infty \text{ and } \sum_{t=1}^{\infty} \eta_t^2 < \infty\\
\end{split}
\end{equation}

\begin{figure*}[!ht]
    \centering
    \subfloat[][Eigenvalue spectrum of SimCLR on CIFAR10 dataset, pre-trained for 200 epochs]{\includegraphics[width=0.33\linewidth]{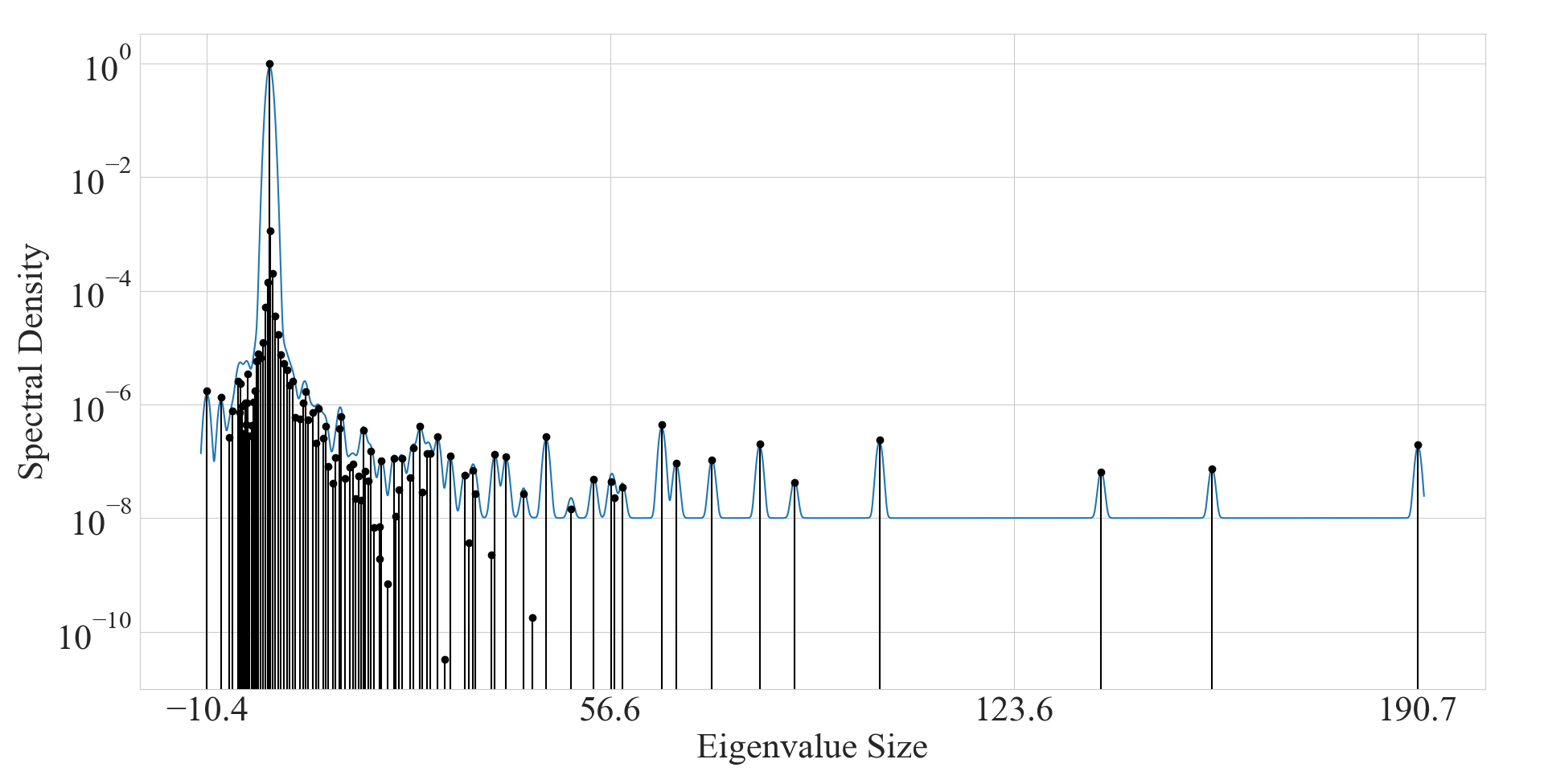}\label{fig:sc10_r18_200eps}}
    \hfill
     \subfloat[][Eigenvalue spectrum of DCL on CIFAR10 dataset, pre-trained for 200 epochs]{\includegraphics[width=0.33\linewidth]{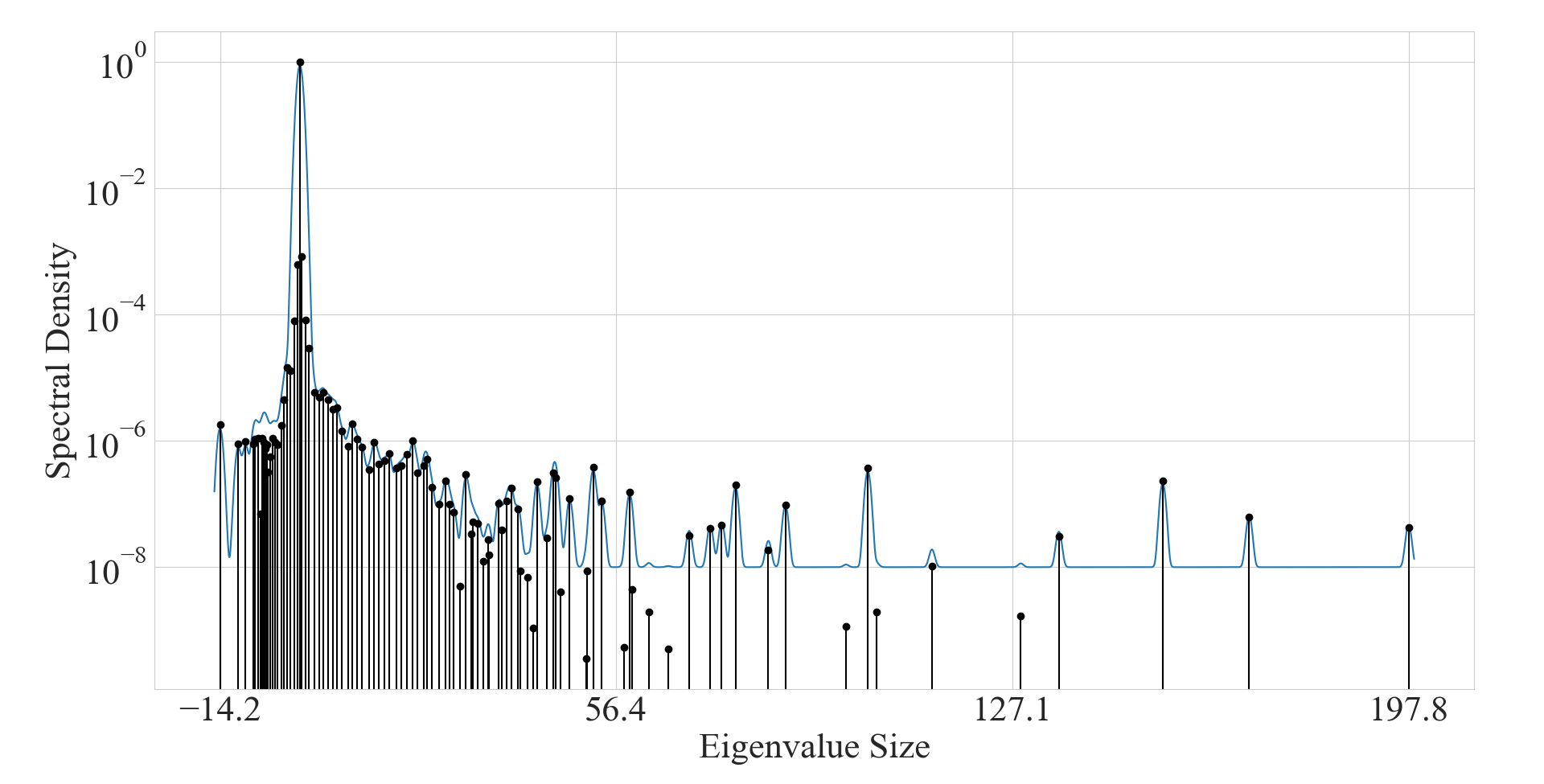}\label{fig:dc10_r18_200eps}}
     \hfill
     \subfloat[][Eigenvalue spectrum of MIOv3 on CIFAR10 dataset, pre-trained for 200 epochs]{\includegraphics[width=0.33\linewidth]{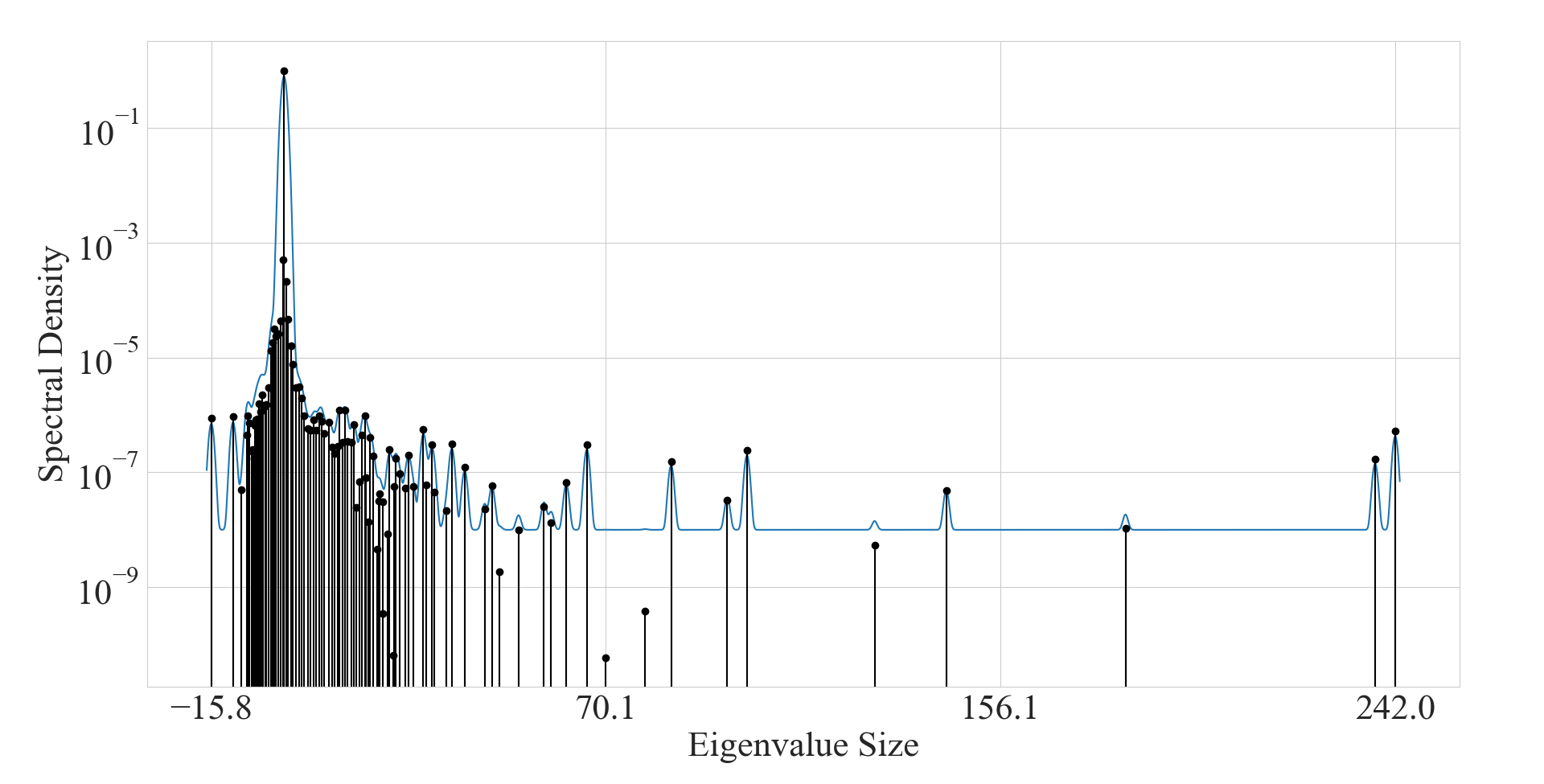}\label{fig:mc10_r18_200eps}}
     \qquad
     \subfloat[][Eigenvalue spectrum of SimCLR on CIFAR100 dataset, pre-trained for 200 epochs]{\includegraphics[width=0.33\linewidth]{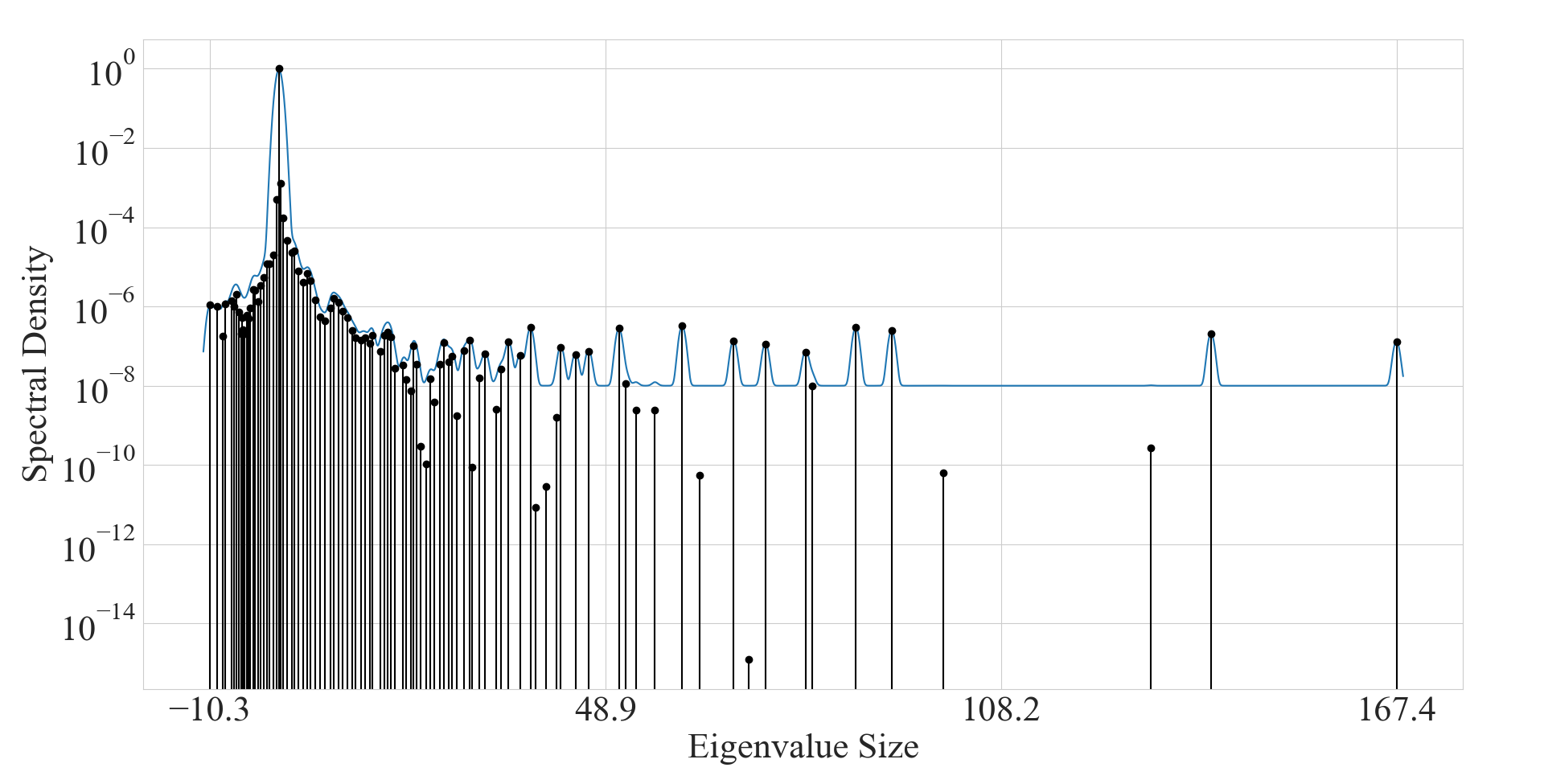}\label{fig:sc100_r18_200eps}}
     \hfill
     \subfloat[][Eigenvalue spectrum of DCL on CIFAR100 dataset, pre-trained for 200 epochs]{\includegraphics[width=0.33\linewidth]{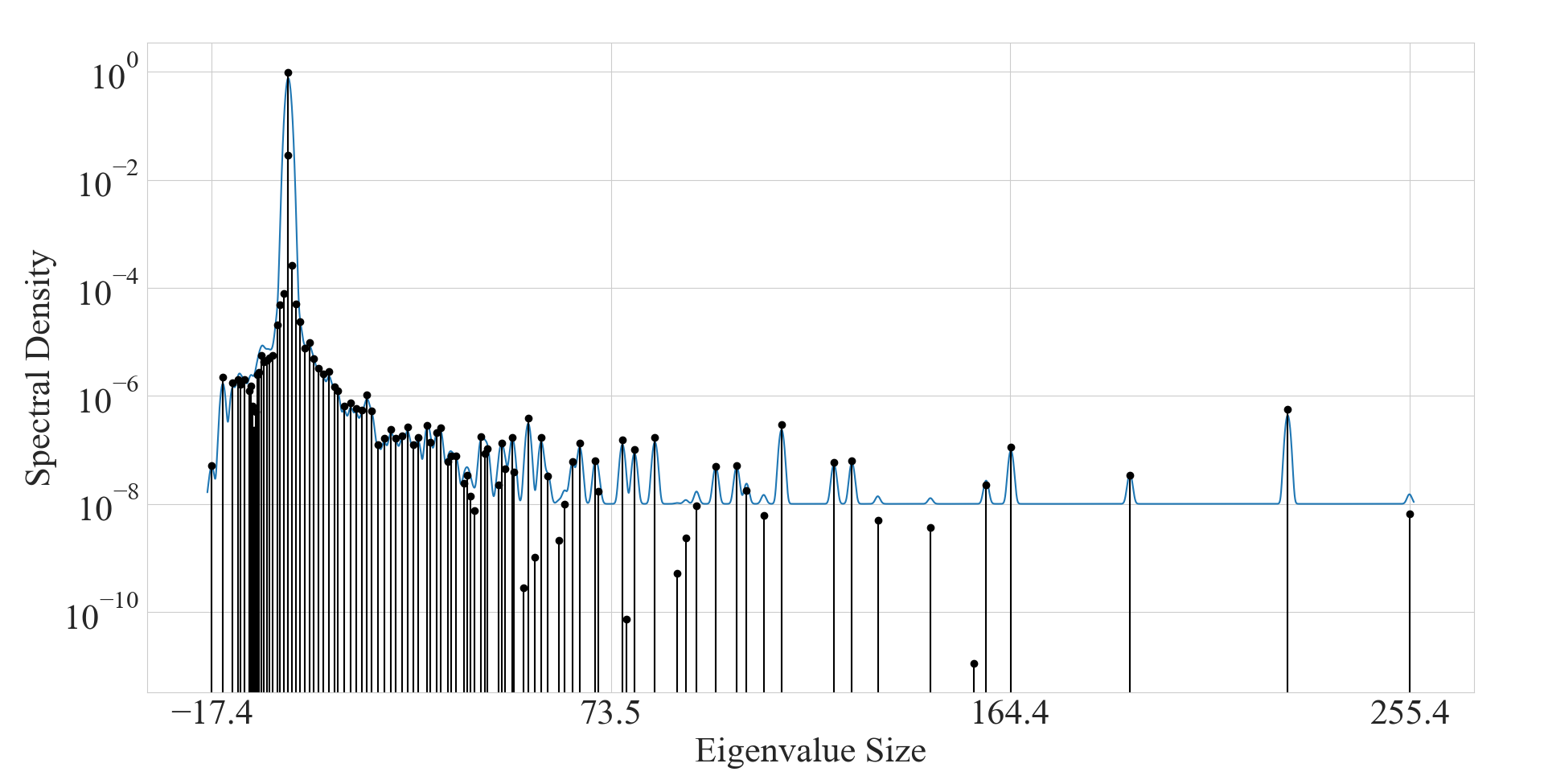}\label{fig:dc100_r18_200eps}}
     \hfill
     \subfloat[][Eigenvalue spectrum of MIOv3 on CIFAR100 dataset, pre-trained for 200 epochs]{\includegraphics[width=0.33\linewidth]{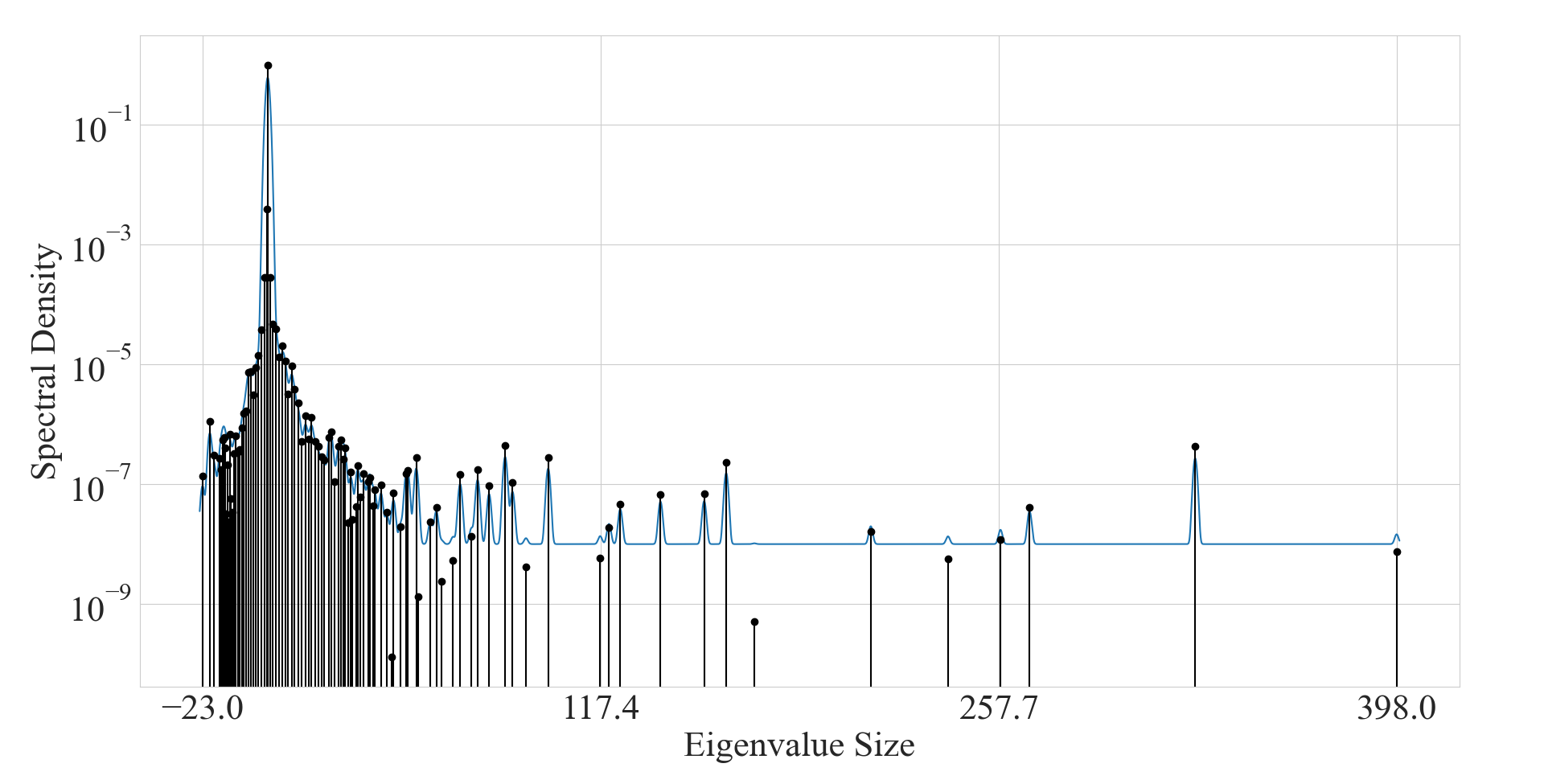}\label{fig:mc100_r18_200eps}}
    \caption{Plot of eigenvalues of parameters of ResNet18, obtained after 200 epochs of pre-training on CIFAR10 and CIFAR100 datasets with different SSL frameworks, namely, SimCLR, DCL and MIOv3.}
    \label{fig:alleigsnew}
\end{figure*}

From Eqns. (\ref{eqn:plin4}) and (\ref{eqn:plin4b}), we calculate a proxy for the linear convergence rate and deduce that only longer training results in convergence provided Eqn. (\ref{eqn:plin1}) is satisfied.

Furthermore, the fact that limited pre-training does not lead to convergence in self-supervised learning can be shown through a visualization of the eigenvalues of the parameters. In Fig. \ref{fig:alleigsnew}, we plot the eigenvalues of the parameters for different self-supervised frameworks after pre-training for 200 epochs on the CIFAR10 and CIFAR100 datasets, and we see that a small proportion of eigenvalues are negative. Although, the mathematical results of our work are also supported by the findings presented in Pascanu et al. (2014) \cite{PascanuDGB14} and Lee et al. (2016) \cite{lee2016gdconvmin}, the plots in Fig. \ref{fig:alleigsnew} show otherwise.  
 Consequently, this indicates that the parameter state at the end of pretraining does not converge to local minima along some eigendirections. However according to the findings in Lee et al. (2016) \cite{lee2016gdconvmin}, the function should have converged to a local minimizer along every eigendirection \textit{almost surely}. This phenomenon points towards the role of step size causing optimization along some eigendirections to get stuck at a saddle point or failing to escape out of the same, as discussed previously. We provide the eigenspectrums of the intermediate epochs in Sec. 3.2.4 of Supplementary.\\
 For computing the eigenvalues of the Hessian of the model parameters, we use the Deep Curvature suite \cite{granziol2020deep}, which uses the Lanczos iteration method to compute the 100 eigenvalues which approximate the entire eigenspectrum.

\section{Experimental Details}
\label{sec:expts}

In this section, first, we are going to discuss the datasets that we used for our experiments, and then the experimental configuration of the models we used. We also present the accuracies of the proposed framework on the mentioned datasets and compare them with the state-of-the-art algorithms.

\begin{table}[h]
    \centering
    \caption{Training and Test images distribution in different datasets}
    \begin{tabular}{|c|c|c|c|c|}
    \hline
        \multirow{2}{*}{Dataset} & No. of & \multicolumn{2}{c|}{Images} & Image\\ \cline{3-4}
        & classes & Training & Test & Dimensions \\ \hline \hline
         CIFAR-10 & 10 & 50000 & 10000 & 32 $\times$ 32\\ \hline
         CIFAR-100 & 100 & 50000 & 10000 & 32 $\times$ 32\\ \hline
         STL-10 & 10 & 5000 & 8000 & 96 $\times$ 96\\ \hline
         Tiny Image Net & 200 & 100000 & 10000 & 64 $\times$ 64\\ \hline
         ImageNet100 & 100 & 130000 & 5000 & 256 $\times$ 256\\ \hline
         ImageNet1K & 1000 & 1.2M & 50000 & 256 $\times$ 256\\ \hline
    \end{tabular}
    \label{tab:datatable}
\end{table}

\definecolor{lightgray}{gray}{0.9}

\begin{table*}[!ht]
    \centering
    \caption{Top-1 200-NN classification accuracy on CIFAR-10, CIFAR-100, STL-10 and Tiny ImageNet-200 datasets of SimCLR, MoCoV2, DCL, DCLW, Barlow Twins, BYOL, and MIOv3 frameworks. The configuration and implementation details are mentioned in \Cref{subsec:imple_det}.}
    \renewcommand{\arraystretch}{1.2}
    \begin{tabular}{|p{2cm}|c|c|p{1.3cm}|p{1.3cm}|p{1.2cm}|c|>{\columncolor{gray!25}}p{1.5cm}|}
    \hline
    \multirow{3}{*}{Dataset} & \multicolumn{7}{c|}{Methods}\\ \cline{2-8}
         &\multicolumn{4}{c|}{Contrastive}&\multicolumn{2}{c|}{Non-Contrastive}&Binary Contrastive\\ \cline{2-8}
        & SimCLR & MoCoV2 & SimCLR +DCL & SimCLR +DCLW & Barlow Twins & BYOL  & MIOv3\\ \hline \hline
        CIFAR-10 & 81.23 & 83.73 & 84.43 &  84.29 & 84.03 &\textbf{86.84} & 
        \underline{86.36}\\ \cline{1-1}\cline{2-8}
        CIFAR-100 & 52.99 & 54.35 & 54.24 & \underline{54.61} & 53.04& 54.02 & \textbf{58.18}\\ \cline{1-1}\cline{2-8}
        STL-10 & 75.65 & 75.64 & 74.46 & 75.49 & 73.24 & 75.87 & \textbf{80.50}\\ \cline{1-1}\cline{2-8}
        Tiny ImageNet-200 & 24.64 & 29.41 & 29.23 & \underline{30.54} &27.42 & 21.21 & \textbf{30.87}\\ \hline 
    \end{tabular}
    \label{tab:exp_res}
\end{table*}

\begin{table}[!ht]
    \centering
    \caption{Top-1 Linear evaluation accuracy on ImageNet100 and ImageNet1K datasets of SimCLR, MoCoV2, DCL, DCLW, and MIOv3 frameworks. The configuration and implementation details for each experiment are mentioned in \Cref{subsec:imple_det}.}
    \renewcommand{\arraystretch}{1.1}
    \begin{tabular}{|c|c|c|}
    \hline
        \multirow{2}{*}{Frameworks} & \multicolumn{2}{c|}{Top -1 Linear Eval. Acc.} \\ \cline{2-3}
          & ImageNet100 & ImageNet1K\\ \hline \hline
         SimCLR & 75.54 \cite{Huang2023macl} & 63.2 \cite{lightlyai}\\
         MoCoV2 & 76.80 \cite{Huang2023macl} & - \\
         SimCLR+DCL & \ul{77.38} \cite{Huang2023macl} & \ul{65.1}  \cite{lightlyai}\\
         SimCLR+DCLW & 76.58 \textit{(repro.)} & 64.2  \cite{lightlyai}\\ \hline
         MIOv3 & \textbf{78.40} & \textbf{65.22} \\ \hline
    \end{tabular}
    \label{tab:exp_res_in100}
\end{table}

\begin{table}
    \centering
    \caption{Comparison of the proposed method with non-contrastive frameworks on the ImageNet100 dataset, pre-trained for a longer duration of 400 epochs. Here 'Linear Eval. Acc.' means Linear Evaluation Accuracy.}
    \renewcommand{\arraystretch}{1.1}
    \begin{tabular}{|c|c|c|c|}
    \hline
        \multirow{2}{*}{Frameworks} & \multirow{2}{*}{Proj. Dim $\#$} & \multicolumn{2}{c|}{Linear Eval. Acc.} \\ \cline{3-4}
         & & Top - 1 & Top - 5\\ \hline \hline
         Barlow Twins \cite{barlow}& 2048 & 78.62 & 94.72 \\ 
         VICReg \cite{vicreg} & 2048 & 79.22 & 95.06 \\
         ZeroICL \cite{Zhang2022zerocl} & 256 &78.02 & \ul{95.61}\\
         ZeroFCL \cite{Zhang2022zerocl} & 2048 &79.32 & 94.94 \\
         ZeroCL \cite{Zhang2022zerocl} &2048&79.26 & 94.98 \\
         WMSE \cite{ermolov2021wmse}& 256 & 69.06 & 91.22\\
         ARB \cite{zhang2022arb}& 2048 & 79.48 & 95.51 \\
         DINO \cite{caron2021dino}& 256 & 74.84 & 92.92 \\ 
         BYOL \cite{byol}& 4096 & 80.09& 94.99\\
         LogDet \cite{zhang2024logdet}& 2048 & \ul{80.38} & 95.45\\\hline
         MIOv3 & 2048 & \textbf{81.66} & \textbf{95.84} \\ \hline 
    \end{tabular}
    \label{tab:compnoncon}
\end{table}

\subsection{Datasets}
\label{subsec:data}

We use four popular datasets to conduct the experiments, namely, CIFAR-10, STL-10, CIFAR-100, Tiny ImageNet, ImageNet100 and ImageNet1K. The dimensions of images in CIFAR-10, STL-10, CIFAR-100, Tiny ImageNet, ImageNet100 and ImageNet1K are $32 \times 32$, $96 \times 96$, $32 \times 32$, $64 \times 64$, $256 \times 256$ and $256 \times 256$, respectively. The details of the distribution of the training and test sets are given in \Cref{tab:datatable}.

\subsection{Implementation Details}
\label{subsec:imple_det}

In this section, we mention the configuration of the best-performing models for the listed frameworks. The frameworks were implemented using the lightly-ai \cite{lightlyai} library. For the experiments on ImageNet1K and ImageNet100 datasets, we used a ResNet50 \cite{resnet} backbone for all our experiments. We optimized the network parameters using a LARS optimizer with the square root learning rate scaling scheme as described in the SimCLR \cite{simclr} paper. For all our experiments we used a batch size of 256, The pre-training and the downstream tasks were run on a single 24GB NVIDIA A5000 GPU using the lightly-ai \cite{lightlyai} library. To ensure faster training and prevent out-of-memory issues, we adopted automatic mixed precision (AMP) training. Time taken for pre-training on the ImageNet100 and ImageNet1K datasets is about 36 hours and 170 hours, respectively.

For the small-scale benchmarks, all the models were trained using ResNet-18 with a batch size of 128. 
The respective loss functions of the self-supervised models were optimized using an SGD optimizer with a learning rate of 0.06 for CIFAR10 and CIFAR100, and a learning rate of 0.03 for STL-10 and Tiny-ImageNet. 
The models were pre-trained for short training periods of 200 epochs only.

We decayed the learning rate following a cosine annealing schedule. The value of weight decay used is $5 \times 10^{-4}$. The ResNet architecture is modified as mentioned in \cite{simclr} only for CIFAR10 and CIFAR100 datasets as the image dimensions are $32 \times 32$.

For MIOv1, MIOv2, and MIOv3, we used a temperature value of $0.2$. Whereas for SimCLR \cite{simclr}, DCL \cite{dcl}, and DCLW \cite{dcl}, we used a temperature of $0.1$ as recommended in the paper \cite{dcl}. For MoCov2 \cite{mocov2}, we used a temperature value of $0.07$, as recommended in its paper. The same value of temperature hyper-parameter value does not yield the best performance for all the frameworks on a particular dataset. Hence, we use a temperature value that yields the best performance for the respective frameworks. 

\section{Results and Analysis}
\label{sec:res}

In this section, we present the comparative results of the proposed framework on small-scale datasets (CIFAR-10, CIFAR-100, STL-10, Tiny-ImageNet) in Sec. \ref{subsec:smallscalecomp}, and large-scale datasets (ImageNet-100 and ImageNet-1k) in Sec. \ref{subsubsec:conimgcomp} for contrastive algorithms and Sec. \ref{subsubsec:nonconimgcomp} for non-contrastive algorithms.

\subsection{Results on Small-Scale Datasets}
\label{subsec:smallscalecomp}
In this subsection, we present the results of frameworks with MIOv1, MIOv2, and the proposed MIOv3 loss function along with the contrastive frameworks SimCLR, MoCOv2, SimCLR+DCL, SimCLR+DCLW and the non-contrastive frameworks BYOL, Barlow Twins. All the frameworks were trained and evaluated using a $k$NN classifier  with $k=200$ on four small-scale datasets as mentioned in Sec. \ref{subsec:data}. The Top-1 200-NN accuracy values are given in Table \ref{tab:exp_res}.

\subsection{{Results on Large-Scale Datasets}}
\label{subsec:largescalecomp}

For the ImageNet100 and ImageNet1K datasets, we report the Top-1 Linear evaluation accuracies.

\subsubsection{Comparison with Contrastive Algorithms}
\label{subsubsec:conimgcomp}

We compare the performance of our proposed method with the contemporary state-of-the-art contrastive frameworks on the ImageNet100 and ImageNet1K datasets in Table \ref{tab:exp_res_in100}. We pre-train our model for a duration of 200 and 100 epochs, respectively. We observe that our proposed framework comfortably outperforms the state-of-the-art contrastive SSL frameworks on the Linear Evaluation task on the ImageNet100 and ImageNet1K datasets.

\subsubsection{Comparison with Non-Contrastive Algorithms}
\label{subsubsec:nonconimgcomp}

 We compare the performance of our proposed method with the contemporary state-of-the-art non-contrastive frameworks on the ImageNet100 dataset in Table \ref{tab:compnoncon}. We pre-train our model for a longer duration of 400 epochs following ZeroCL \cite{Zhang2022zerocl} and ARB \cite{zhang2022arb}. We observe that our proposed framework comfortably outperforms the state-of-the-art non-contrastive frameworks on the Linear Evaluation task on the ImageNet100 dataset.

\section{Ablation Studies}
\label{sec:ablation_studies}

\subsection{Effect of Temperature}
\label{subsec:tempabl}

In this section, we study the effect of temperature on the proposed loss and also analyse the behaviour of MIOv1, MIOv2 and MIOv3 with varying temperature hyper-parameter values. We also notice the performance of MIOv1, MIOv2, and MIOv3 on different small-scale datasets (CIFAR-10, CIFAR-100, STL-10) and ImageNet-100 for different temperature values. As discussed in Sec. \ref{subsec:miov2} and \ref{subsec:miov3}, we observe that the performance of MIOv2 is better than MIOv1 and MIOv3 at lower temperatures. But at higher temperatures, MIOv1 performs better than MIOv2. Furthermore, the drop in performance of MIOv3 at low temperature values is primarily due to increased repulsion between samples in false negative pairs. It is noteworthy that MIOv3 achieves the best performance at temperature 0.2 and also outperforms contemporary state-of-the-art SSL frameworks.

\begin{table}[!ht]
    \centering
    \caption{Variation of performance of MIOv1, MIOv2, MIOv3 for different temperature values, supporting the effect of $\mathcal{R}_{pp}$ and $\mathcal{R}_{nn}$ as described in Sec. \ref{subsec:miov2} and \ref{subsec:miov3}.}
    \begin{tabular}{|c|c|c|c|c|c|}
    \hline
        \multirow{2}{*}{Dataset} & \multirow{2}{*}{MIOvx} & \multicolumn{4}{c|}{Temperature} \\ \cline{3-6}
       &  & 0.07 & 0.1 & 0.2 & 0.5 \\ \hline \hline
        \multirow{3}{*}{CIFAR-10} &v1 & 72.55 & 76.19 & 80.72 & 82.80 \\
        & v2 & 82.8 & \textbf{82.87} & 81.12 & 79.34 \\
        & v3 & 38.4 & 81.55 & \textbf{86.36} & \textbf{85.09} \\ \hline \hline
        \multirow{3}{*}{CIFAR-100} &v1 & 36.3 & 42.9 & 50.7 & 49.7 \\
        &v2 & 47.1 & 46.2 & 43.5 & 35.8 \\
        &v3 & 33.6 & \textbf{50.1} & \textbf{58.2} & \textbf{53.5} \\ \hline \hline
        \multirow{3}{*}{STL-10} &v1 & 62.8 & 67.3 & 72.42 & 71.78 \\
        &v2 & 71.33 & \textbf{71.3} & 70.2 & 65.0 \\
        &v3 & 32.1 & 71.0 & \textbf{75.83} & \textbf{73.8} \\ \hline \hline
        \multirow{3}{*}{IN-100} &v1 & - & 65.04 & 71.98 & 74.02 \\
        &v2 & - & 75.24 & 73.80 & 70.98 \\
        &v3 & - & \textbf{76.16} & \textbf{78.4} & \textbf{77.20} \\ \hline \hline
    \end{tabular}
    \label{tab:tempabltab}
\end{table}

\subsection{Ablation on Training Duration}
\label{subsec:ableps}

In this subsection, we study the effect of training duration on the performance of the contemporary and proposed contrastive SSL frameworks. We observe from the empirical results presented in Table \ref{tab:ablepochs}, that the proposed framework outperforms SimCLR and SimCLR+DCL when pre-trained for 200 and 1000 epochs.

\begin{table}[!ht]
    \centering
    \caption{Comparison of SimCLR, DCL and MIOv3 on CIFAR10 and CIFAR100 datasets pre-trained for 200 and 1000 epochs.}
    \begin{tabular}{|c|c|c|c|c|}
    \hline
        Epochs & Dataset & SimCLR & DCL & MIOv3 \\ \hline \hline
        200 & \multirow{2}{*}{CIFAR10} & 81.23 & 84.43 & 86.36\\
        1000 & & 89.5 & 88.16 & 89.89 \\
        \hline \hline
        200 & \multirow{2}{*}{CIFAR100} & 52.99 & 54.24 & 58.18\\
        1000 & & 60.5 & 61.03 & 62.99 \\ \hline
    \end{tabular}
    \label{tab:ablepochs}
\end{table}
\subsection{Ablation on Batch Size}
\label{subsec:ablbs}

In this subsection, we study the effect of varying batch sizes on the performance of the proposed contrastive SSL frameworks. We observe from the empirical results presented in Table \ref{tab:ablbs}, that the proposed framework performs best with batch size 128 on CIFAR10 and 256 on CIFAR100. However, the result with batch size 128 is presented in Table \ref{tab:exp_res} for a fair comparison.

\begin{table}[!ht]
    \centering
    \caption{Ablation of 200-NN Top-1 accuracy on CIFAR-10 and CIFAR-100 datasets for batch sizes of 64, 128, 256, and 512.}
    \begin{tabular}{|c|c|c|c|c|c|}
    \hline
    \multirow{2}{*}{Dataset} &\multirow{2}{*}{Method} & \multicolumn{4}{c|}{Batch size}\\ \cline{3-6}
    && 64 & 128 & 256 & 512 \\ \hline
    \multirow{3}{*}{CIFAR10} & MIOv3     & \textbf{85.9} & \textbf{86.28} & \textbf{86.19}  & \textbf{85.9} \\\cline{2-6}
         & DCL & 84.32 & 84.43 & 84.4 & 83.86 \\\cline{2-6}
         & SimCLR & 81.12 & 81.23 & 81.4 & 81.3\\\hline
         \multirow{3}{*}{CIFAR100} & MIOv3 & \textbf{55.79} & \textbf{56.97} & \textbf{57.51} & \textbf{56.72}\\ \cline{2-6}
         & DCL & 54.23 & 54.24 & 56.2 & 55.8 \\\cline{2-6}
         & SimCLR & 51.66 & 52.99 & 53.6 & 53.69\\\hline
    \end{tabular}
    \label{tab:ablbs}
\end{table}



\subsection{Effect of Decrease in Number of Parameters}
\label{subsec:numparamabl}

In this ablation study, we mainly discuss the effect of the number of parameters on performance. 
Neural networks are in general over-parameterized. In this section, we intend to conduct an experimental study to determine the efficiency of parameter utilization in SSL. With a decrease in the number of parameters, the performance will surely drop. The performance of any particular framework with decreased parameters implies how much of the total number of parameters is being utilized by the framework for representation learning.

\begin{table}[!ht]
    \centering
    \caption{200-NN accuracy of MIOv3, SimCLR, SimCLR+DCL, BYOL frameworks using 2 different models with decreasing number of parameters on CIFAR-10 and CIFAR-100 datasets obtained after 500 and 200 epochs of pre-training, respectively with a batch size of 128.}
    \begin{tabular}{|c|c|c|}
    \hline
         & ResNet-18 & ResNet-9\\ \hline \hline
        \# Basic Blocks & [2,2,2,2] & [1,1,1,1] \\ 
        Base Channels & 64 & 64 \\
        \# Params & $\sim$ 11M & $\sim$ 5M \\ \hline
        \multirow{2}{*}{Model} & \multicolumn{2}{c|}{Top-1 Accuracy (\%)}\\ \cline{2-3}
         & \multicolumn{2}{c|}{CIFAR-10}\\\hline
        MIOv3 & \textbf{89.00} & \textbf{84.75}\\ 
        SimCLR & 84.97 & 80.18\\
        SimCLR+DCL & 86.4 & 82.82\\
        BYOL & 90.13 & 84.56 \\ \hline \hline
        & \multicolumn{2}{c|}{CIFAR-100}\\\cline{2-3}
        MIOv3 & \textbf{58.18} & \textbf{53.98}\\ 
        SimCLR & 52.99 & 48.19\\
        SimCLR+DCL & 54.24 & 51.21\\
        BYOL & 54.02 & 50.98 \\ \hline
    \end{tabular}
    \label{tab:decparamstudy}
\end{table}

In Table \ref{tab:decparamstudy}, we have presented the 200-NN accuracy of MIOv3, SimCLR, SimCLR+DCL, and BYOL on the CIFAR10 and CIFAR100 datasets. The configuration of the base encoder is also mentioned in the table, along with the number of parameters. Intuitively, a decrease in the number of parameters will eliminate some useful parameters. However, the ability of the respective frameworks to utilize the previously redundant parameters can be observed from the performance as given in \Cref{tab:decparamstudy}.

We see that the performance of all frameworks decreases with the decrease in the number of parameters. However, it is worth noting that, under the effect of decreasing the number of parameters, our proposed MIOv3 framework outperforms all self-supervised learning frameworks on the CIFAR dataset.

\subsection{Overview of Transfer Learning Performance}
\label{subsec:transferlearning}

It is imperative to show the quality of representations learnt by the self-supervised models on other datasets. For this purpose, we chose four medical image datasets, MURA \cite{rajpurkar2017mura}, Chaoyang \cite{zhu2022chaoyang}, ISIC2016 Lesion Classification \cite{gutman2017isic2016}, and MHIST \cite{wei2021mhist} datasets.

We encounter both multi-class and binary classification tasks in this section. The MURA, ISIC2016 Lesion classification and MHIST consist of the binary classification task. The Chaoyang, Flowers, CIFAR10, and CIFAR100 consist of the multi-class classification tasks.

We fine-tuned the models pre-trained on the ImageNet1K \cite{imagenet} dataset for 50 epochs on these datasets, using an SGD optimizer. 
We used class weights to mitigate the effect of imbalance in all datasets, except CIFAR10 and CIFAR100. For all the experiments, a batch size of 128 was used.

For the multiclass and binary classification tasks, we used a learning rate of 0.1 and 1.0, respectively, and a multi-step decay scheduler with a decay by a factor of 0.1 at the 30th and 40th epochs. For the MURA, ISIC2016, and MHIST datasets,  we used positive class weights of 0.7097, 4.24, and 2.45, respectively. For the Chaoyang dataset, the class weights used were $1.264, 1.667, 1.0$, and $2.114$ for the 4 classes.

From the results presented in Table \ref{tab:transferlearning}, we can see that the proposed method outperforms SimCLR \cite{simclr} on all seven datasets. The proposed method also outperforms the contemporary state-of-the-art self-supervised contrastive learning algorithm (DCL \cite{dcl}) on 5 out of 7 datasets. It can also be seen that the performance of our proposed framework is close to the supervised baseline.



\begin{table}[!ht]
    \centering
    \caption{Performance comparison of the proposed method (MIOv3) with contemporary self-supervised contrastive state-of-the-art methods on transfer learning tasks. The results of the supervised learning baseline are also provided here for reference.}
    \begin{tabular}{|c|p{2.5cm}|p{2cm}|>{\columncolor{gray!25}}c||p{3.5cm}||}
    \hline
        Datasets & SimCLR \citep{simclr} & DCL \citep{dcl} & MIOv3 & Supervised \\ \hline \hline 
        MURA & 81.81 & 81.70 & \textbf{82.49} & 82.10 \citep{Nauta_2023_CVPR, nauta2023interpreting}\\ \hline 
        Chaoyang & 83.22 & 83.12 & \textbf{84.34}  & 83.50 \citep{galdran2023mhml}\\\hline 
        ISIC2016 & 84.70 & \textbf{85.48} & 85.22  & 85.50 \citep{isic2016lb}\\\hline 
        MHIST & 83.62 & \textbf{85.26} & 84.03  & 86.90 \citep{springenberg2023hist}\\\hline 
        CIFAR10 & 96.93 & 97.09 & \textbf{97.11} & 97.50 \citep{byol}\\\hline 
        CIFAR100 & 82.99&83.03& \textbf{83.77} & 86.40 \citep{byol}\\ \hline 
        Flowers & 93.82 & 94.11 & \textbf{94.70} & 97.60 \citep{byol}\\\hline \hline 
    \end{tabular}
    \label{tab:transferlearning}
\end{table}

\section{Conclusion}
\label{sec:conc}

In this work, we proposed a novel binary contrastive loss function, MIOv3 loss, that optimizes the mutual information between samples in positive and negative pairs. Initially, we started from the base version MIOv1 and modified it to obtain MIOv3 with superior performance. Through mathematical calculation, we provide a lower bound of the base loss function MIOv3, which is the difference between the mutual information of the samples in the negative and positive pairs. From the eigenspectrum analysis we also observed how the optimization process proceeds on the parameter space in self-supervised learning. We also show both mathematically and empirically that under a longer duration of training, SSL frameworks converge to strict saddle points in the loss landscape. Through experimental evidence, we show that the proposed MIOv3 framework outperforms the contemporary self-supervised learning frameworks. We also study the effect of temperature hyper-parameter, training duration, batch size, and decrease in model parameters on the downstream performance and notice that our proposed framework outperforms the contemporary frameworks in all scenarios.




\newpage

\appendix
\section{List of Symbols}
\label{app:theorem}
\begin{table}[!ht]
\centering
\caption{List of Symbols}
\scriptsize
\begin{tabular}{c|c|p{11cm}}
\toprule
Sl. No. & Symbol & What it means? \\
\midrule
1&$\mathbb{R}^n$&$n$-dimensional Real Space\\
2&$\chi$&Input Pair Space\\
3&$X$&Sampled Input Pair from $\chi$\\
4&$\mathcal{X}_{+}$&Set of all positive pairs\\
5&$\mathcal{X}_{-}$&Set of all negative pairs\\
6&$f$&Encoder\\
7&$\theta$&Encoder Parameters\\
8&$h$&Feature vector obtained after passing a sample $x$ through the encoder $f_{\theta}$\\
9&$g$&Projector\\
10&$\psi$&Projector Parameters\\
11&$z$&Feature vector obtained after passing a sample $x$ through the encoder $f_{\theta}$ and projector $g_{\psi}$\\
12&$\mathcal{Z}$&Space of projected latent feature vectors\\
13&$C_{i,j}$&Cosine similarity between feature vectors of samples in a pair $(x_i, x_j)$\\
14&$s(z_i,z_j)$&Scoring function, calculate a pre-defined metric score between the latent vectors $z_i$ and $z_j$\\
15&$p_+$&Distribution of positive pairs\\
16&$p_-$&Distribution of negative pairs\\
17&$P_+^{i,j}$&Probability of obtaining the positive pair $(x_i,x_j)$, and subsequently $(z_i,z_j)$\\
18&$P_-^{i,j}$&Probability of obtaining the negative pair $(x_i,x_j)$, and subsequently $(z_i,z_j)$\\
19&$\mathcal{L}$&Loss function\\
20&$\mathbb{P}$&Parameter space. Includes {$\theta, \psi$}\\
21&T&Total number of training steps or iterations\\
22&$\mathcal{P}$&A point in the Parameter space\\
22&$\mathcal{P}_T$&Final Parameter state after T training iterations, $\mathcal{P}_T \in \mathbb{P}$\\
23&$\mathcal{P}_t$&Parameter state after t training iterations, $\mathcal{P}_t \in \mathbb{P}$\\
24&$\mathbb{G}$&Gradient space\\
25&$\mathcal{G}$&A point in the Gradient space\\
26&$\mathcal{G}_t$&Gradient state after t training steps, $\mathcal{G}_t \in \mathbb{G}$\\
27&$\mathcal{I}$&Mutual Information\\
28&$\eta$&Step Size / Learning Rate\\
29&$\mathcal{H}$&Hessian Matrix\\
30&$\mathcal{E}$&Joint Embedding Space\\
31&$L$&Lipschitz Constant\\
32&$N$&Batch Size\\
\bottomrule
\end{tabular}
\label{tab:symbtab}
\end{table}

\normalsize
\section{Overview of Sample Arrangement for Implementation and Mathematical Derivations}
\label{sec:suppsec1}

\begin{figure}[!ht]
    \centering
    \includegraphics[width = 0.8 \linewidth]{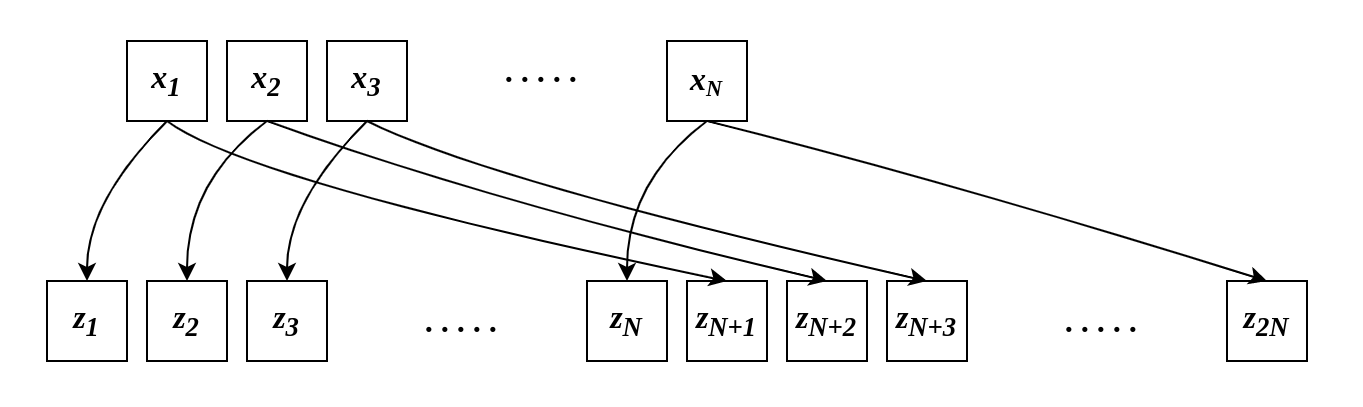}
    \caption{This figure shows how the feature vectors are obtained from the samples ($x_1, x_2, \hdots, x_N$) in a batch.}
    \label{fig:pa}
\end{figure}

\begin{figure}[ht!]
\centering
    \includegraphics[width = 0.7 \linewidth]{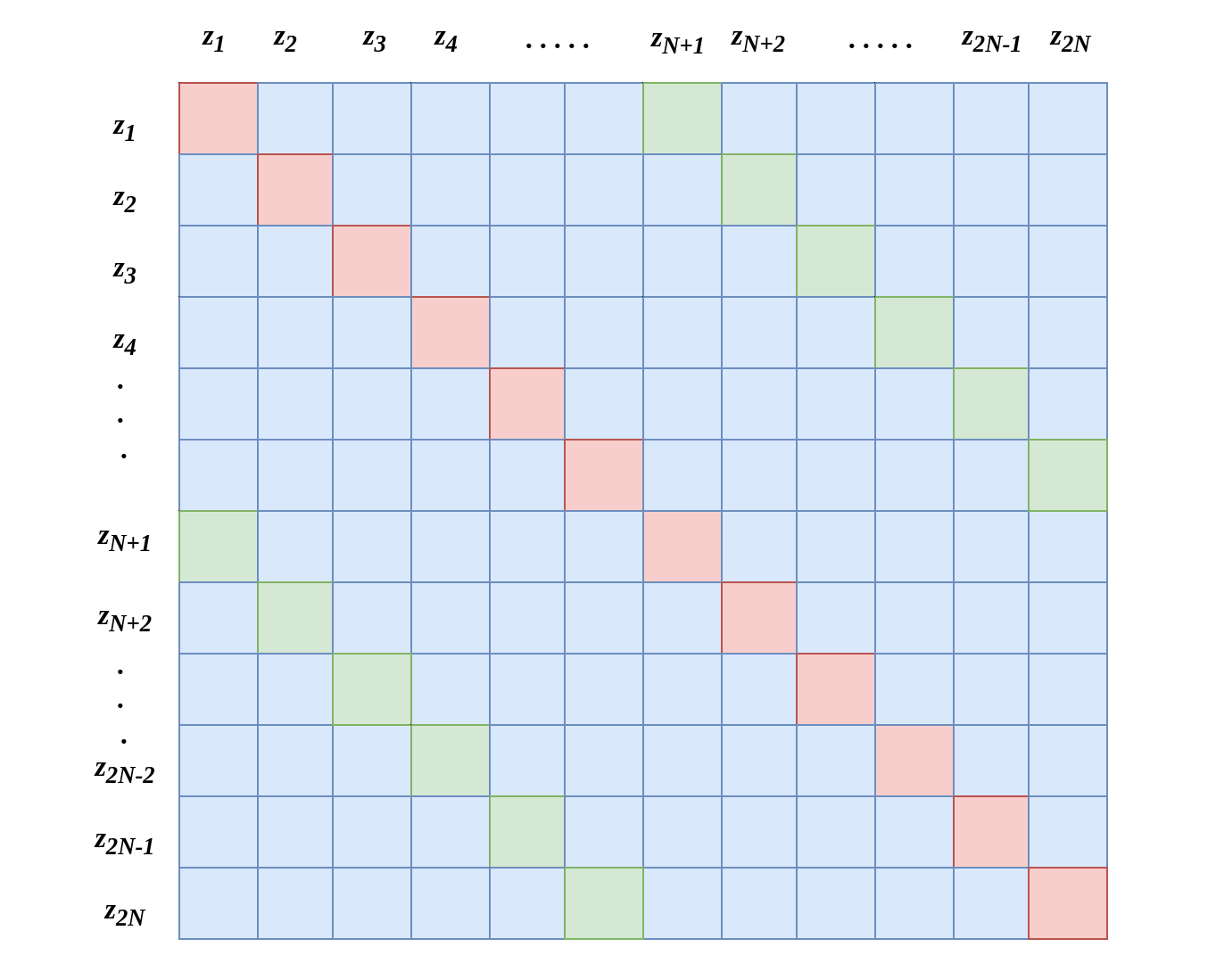}
    \caption{This figure shows how the pairings are obtained. The red cells indicate self-pairs, green cells indicate positive pairs, i.e., pairings between feature vectors of two augmented versions of the same sample, and blue cells indicate negative pairings, i.e. pairings between feature vectors of different samples.}
    \label{fig:pb}
\end{figure}

In Fig. \ref{fig:pa}, we show how the feature vectors are obtained from the samples in a batch and how they are arranged for the final step of calculating the loss. Fig. \ref{fig:pb} shows how the different pairs are obtained from the feature vectors.



\section{Analysis of Gradients and Hessian of Proposed Loss Function $\mathcal{L}_{v3}$ w.r.to the parameters $\psi$}
\label{sec:analloss}
\subsection{Calculate the gradient of $\mathcal{L}_{v3}$ with respect to the parameters $\psi$}
\label{subsec:gradlvipsi}

Let us rewrite the expression for $\mathcal{L}_{v3}$ again,

\begin{equation}
    \label{eqn:expandedlv3}
     \centering
    \begin{split}
        \mathcal{L}_{v3} = & - \frac{1}{N} \sum_{\substack{n=1}}^{N} \frac{C_{n,n+N}}{\tau} + \frac{1}{T_N} \sum_{\substack{n=1}}^{2N} \sum_{\substack{m=1\\m \neq n,n+N}}^{2N} e^{\frac{C_{n,m}}{\tau}} \\
    \end{split}
\end{equation}

\begin{equation}
\label{eqn:expandedlv1}
\begin{split}
	\mathcal{L}_{v1} =& -\frac{1}{N}\sum_{n = 1}^N{\frac{C_{n,n+N}}{\tau}} +\frac{1}{N}\sum_{n = 1}^N{ln \left(1+e^{\frac{C_{n,n+N}}{\tau}}\right)} + \frac{1}{T_N}\sum_{n = 1}^{2N} \sum_{\substack{m=1\\m\neq n,n+N}}^{2N}{ln \left({1+e^{\frac{C_{n,m}}{\tau}}}\right)}
 \end{split}
\end{equation}

where $N$ is the batch size, $\tau$ is the temperature hyperparameter and $C_{ij}$ denotes the cosine similarity between the feature vectors $z_i$ and $z_j$. The feature vectors $z_i$ and $z_j$ are obtained by passing the input $x_i, x_j \in \chi$ through the encoder $f_{\theta}$ with parameters $\theta$ and the projector $g_{\psi}$ with parameters $\psi$. Thus,

\begin{equation}
	z_n = g_{\psi}\left( f_{\theta}\left(x_n\right)\right)
\end{equation}

Without the loss of generality, we can assume that the projector consists of a single perceptron layer or we can consider a two-layered perceptron model with weights $\psi^{(1)}_{Di \times H}$ and $\psi^{(2)}_{H \times Do}$ for the two layers as a single layer with weights $\psi = \psi^{(1)}_{Di \times H}\cdot\psi^{(2)}_{H \times Do}$. 

\begin{equation}
\begin{split}
\psi & = \psi^{(1)}_{Di \times H}\cdot\psi^{(2)}_{H \times Do}\\
& = \begin{pmatrix} \psi^{(1)}_{1, 1}, \psi^{(1)}_{1, 2}, \ldots ,\psi^{(1)}_{1, H}\\ \psi^{(1)}_{2, 1}, \psi^{(1)}_{2, 2}, \ldots ,\psi^{(1)}_{2, H}\\ \vdots\\ \psi^{(1)}_{D_i, 1}, \psi^{(1)}_{D_i, 1}, \ldots ,\psi^{(1)}_{D_i, H} \end{pmatrix} \begin{pmatrix} \psi^{(2)}_{1, 1}, \psi^{(2)}_{1, 2}, \ldots ,\psi^{(2)}_{1, D_o}\\ \psi^{(2)}_{2, 1}, \psi^{(2)}_{2, 2}, \ldots ,\psi^{(2)}_{2, D_o}\\ \vdots\\ \psi^{(2)}_{H, 1}, \psi^{(2)}_{H, 1}, \ldots ,\psi^{(2)}_{H, D_o} \end{pmatrix} = \begin{pmatrix} \psi_{1, 1}, \psi_{1, 2}, \ldots ,\psi_{1, D_o}\\ \psi_{2, 1}, \psi_{2, 2}, \ldots ,\psi_{2, D_o}\\ \vdots\\ \psi_{D_i, 1}, \psi_{D_i, 1}, \ldots ,\psi_{D_i, D_o} \end{pmatrix}_{D_i \times D_o}\\
\end{split}
\end{equation}

To get the final feature vector $z_n$, we multiply the output of the encoder $h_{\theta n}$ with the transpose of the weight matrix $\psi$. The shape of the feature vector $z_n$ is $D_o \times 1$. Continuing, we get,

\begin{equation}
\label{eqn:znasmat}
\begin{split}
	z_n =& (\psi_{D_i \times D_o})^T \cdot f_{\theta}\left(x_n\right)_{D_i \times 1}= (\psi_{D_i \times D_o})^T \cdot (h_{\theta n})_{D_i \times 1} =  \psi_{D_o \times D_i}^T \cdot (h_{\theta n})_{D_i \times 1} \\=& \begin{pmatrix} (\psi_{\forall, 1})^T\\ (\psi_{\forall, 2})^T\\ \cdot\\ (\psi_{\forall, k})^T \\ \cdot \\ (\psi_{\forall, D_o})^T\end{pmatrix} \cdot \begin{pmatrix} (h_{\theta})_{1,1}\\(h_{\theta})_{2,1}\\\cdot\\(h_{\theta})_{k,1}\\\cdot\\(h_{\theta})_{D_i,1} \end{pmatrix}
\end{split}
\end{equation}

where $(\psi_{\forall, k})^T$ denotes the transposed version of the $k$-th column of the weight matrix $\psi$, or the $k$-th row of the matrix $\psi^T$. The $k$-th element of the feature vector $z_n$ is obtained by 
\begin{equation}
\begin{split}
	z_n^{(k)} &= (\psi_{\forall, k})^T \cdot h_{\theta n} = h^T_{\theta n} \cdot \psi_{\forall, k}
 \end{split}
\end{equation}

Taking the derivative of $\mathcal{L}_{v3}$ with respect to a column of $\psi$, or a row of $\psi^T$, we get 

\begin{equation}
	\label{eqn:delL1}
	\begin{split}
		\del{\mathcal{L}_{v3}}{(\psi_{\forall, k})^T} = &-\frac{1}{N}\sum_{n = 1}^N\del{\frac{C_{n,n+N}}{\tau}}{(\psi_{\forall, k})^T} + \frac{1}{T_N}\sum_{n = 1}^{2N} \sum_{\substack{m=1\\m\neq n,n+N}}^{2N}\del{e^{\frac{C_{n,m}}{\tau}}}{(\psi_{\forall, k})^T}\\
		= &-\frac{1}{N\tau}\sum_{n = 1}^N\del{C_{n,n+N}}{(\psi_{\forall, k})^T} + \frac{1}{T_N}\sum_{n = 1}^{2N} \sum_{\substack{m=1\\m\neq n,n+N}}^{2N}\del{e^{\frac{C_{n,m}}{\tau}}}{(\psi_{\forall, k})^T}\\
		= &-\frac{1}{N\tau}\sum_{n = 1}^N\del{\sum_{i = 1}^{Do}z^{(i)}_n\cdot z^{(i)}_{n+N} }{(\psi_{\forall, k})^T} + \frac{1}{T_N}\sum_{n = 1}^{2N} \sum_{\substack{m=1\\m\neq n,n+N}}^{2N}e^{\frac{C_{n,m}}{\tau}}\del{\left(\frac{C_{n,m}}{\tau}\right)}{(\psi_{\forall, k})^T}\\
		= &-\frac{1}{N\tau}\sum_{n = 1}^N\del{\sum_{i = 1}^{Do}z^{(i)}_n\cdot z^{(i)}_{n+N} }{(\psi_{\forall, k})^T} + \frac{1}{T_N \tau}\sum_{n = 1}^{2N} \sum_{\substack{m=1\\m\neq n,n+N}}^{2N}e^{\frac{C_{n,m}}{\tau}}\del{C_{n,m}}{(\psi_{\forall, k})^T}\\
		= &-\frac{1}{N\tau}\sum_{n = 1}^N\del{\sum_{i = 1}^{Do}z^{(i)}_n\cdot z^{(i)}_{n+N} }{(\psi_{\forall, k})^T} + \frac{1}{T_N \tau}\sum_{n = 1}^{2N} \sum_{\substack{m=1\\m\neq n,n+N}}^{2N}e^{\frac{C_{n,m}}{\tau}}\del{\sum_{i = 1}^{Do}z^{(i)}_n\cdot z^{(i)}_m}{(\psi_{\forall, k})^T}\\
	\end{split}
\end{equation}

We have a common expression $\del{\sum_{i = 1}^{Do}z^{(i)}_n\cdot z^{(i)}_m}{(\psi_{\forall, k})^T}$ (a column vector) in all the three terms in the expression for $\del{\mathcal{L}_{v3}}{(\psi_{\forall, k})^T}$. So, we will evaluate it first and then continue with our derivation,

\begin{equation}
	\label{eqn:delcossim}
	\begin{split}
		\del{\sum_{i = 1}^{Do}z^{(i)}_n\cdot z^{(i)}_m}{(\psi_{\forall, k})^T} =& \sum_{i = 1}^{Do} \left(z^{(i)}_n \cdot \del{z^{(i)}_m}{(\psi_{\forall, k})^T} + \del{z^{(i)}_n}{(\psi_{\forall, k})^T} \cdot z^{(i)}_m \right)\\
		= &\sum_{i = 1}^{Do} \left(z^{(i)}_n \cdot \del{\left((\psi_{\forall, i})^T\cdot h_{\theta m}\right)}{(\psi_{\forall, k})^T} + \del{\left((\psi_{\forall, i})^T\cdot h_{\theta n}\right)}{(\psi_{\forall, k})^T} \cdot z^{(i)}_m \right)\\
		= &\sum_{i = 1}^{Do} \left(z^{(i)}_n \cdot \del{ (\psi_{\forall, i})^T}{(\psi_{\forall, k})^T} \cdot h_{\theta m} + \del{(\psi_{\forall, i})^T}{(\psi_{\forall, k})^T} \cdot h_{\theta n} \cdot z^{(i)}_m \right)\\
		= & \left(z^{(k)}_n \cdot \del{(\psi_{\forall, k})^T }{(\psi_{\forall, k})^T} \cdot h_{\theta m} + \del{(\psi_{\forall, k})^T}{(\psi_{\forall, k})^T} \cdot h_{\theta n} \cdot z^{(k)}_m \right)\\
	=& z^{(k)}_n \cdot h_{\theta m} + h_{\theta n} \cdot z^{(k)}_m\\
	\end{split}
\end{equation}

We denote the expression $\sum_{i = 1}^{Do}z^{(i)}_n\cdot z^{(i)}_m$ by $C_{n,m}$ and its derivative with respect to $(\psi_{\forall, k})^T$, i.e. $\del{C_{nm}}{(\psi_{\forall, k})^T} = z^{(k)}_n \cdot h_{\theta m} + h_{\theta n} \cdot z^{(k)}_m$ by $A^{(k)}_{n,m}$. Dimension of $A^{(k)}_{n,m}$ and subsequently of $\del{\mathcal{L}_{v3}}{(\psi_{\forall, k})^T}$ is $D_i \times 1$. That is $A^{(k)}_{n,m}$ and subsequently $\del{\mathcal{L}_{v3}}{(\psi_{\forall, k})^T}$ is a \textbf{\textit{column vector}}.

Putting Eqn. \ref{eqn:delcossim} in Eqn. \ref{eqn:delL1}, we get, 
\begin{equation}
	\label{eqn:delL1_2}
	\begin{split}
		\del{\mathcal{L}_{v3}}{(\psi_{\forall, k})^T} 
		= &-\frac{1}{N\tau}\sum_{n = 1}^N\del{\sum_{i = 1}^{Do}z^{(i)}_n\cdot z^{(i)}_{n'} }{(\psi_{\forall, k})^T} + \frac{1}{T_N \tau}\sum_{n = 1}^{2N} \sum_{\substack{m=1\\m\neq n,n'}}^{2N}e^{\frac{C_{n,m}}{\tau}}\del{\sum_{i = 1}^{Do}z^{(i)}_n\cdot z^{(i)}_m}{(\psi_{\forall, k})^T}\\
		= &-\frac{1}{N\tau}\sum_{n = 1}^N A^{(k)}_{n,n'} + \frac{1}{T_N \tau}\sum_{n = 1}^{2N} \sum_{\substack{m=1\\m\neq n,n'}}^{2N}e^{\frac{C_{n,m}}{\tau}} A^{(k)}_{n,m} \\
	\end{split}
\end{equation}

where $n' = n+N$.

Therefore, using $n' = n+N$,
\begin{equation}
    \label{eqn:delL1_3}
    \begin{split}
    \del{\mathcal{L}_{v3}}{(\psi_{\forall, k})^T} = &-\frac{1}{N\tau}\sum_{n = 1}^N A^{(k)}_{n,n+N} + \frac{1}{T_N \tau}\sum_{n = 1}^{2N} \sum_{\substack{m=1\\m\neq n,n+N}}^{2N}e^{\frac{C_{n,m}}{\tau}} A^{(k)}_{n,m}\\
    \end{split}
\end{equation}


\subsection{Calculate the Hessian of $\mathcal{L}_{v3}$ with respect to the parameters $\psi$}
\label{subsec:calchesslv3}

We have already calculated the first derivative of $\mathcal{L}_{v3}$ with respect to the parameters $\psi$ or $(\psi_{\forall, k})^T$. We proceed to calculate the Hessian of the loss function $\mathcal{L}_{v3}$ with respect to $\psi$ in a similar manner to the first derivative. 

Taking derivative of $\del{\mathcal{L}_{v3}}{(\psi_{\forall, k})^T}$ with respect to $\psi_{\forall, l}$, $l$-th column of $\psi$, we get, 

\begin{equation}
\label{eqn:del2L1}
    \begin{split}
        \DDel{2}{\mathcal{L}_{v3}}{\psi_{\forall, l}}{(\psi_{\forall, k})^T} 
        &= \left[\del{}{\psi_{1, l}}\del{\mathcal{L}_{v3}}{(\psi_{\forall, k})^T},\ldots,\del{}{\psi_{D_i,l}}\del{\mathcal{L}_{v3}}{(\psi_{\forall, k})^T} \right]^T\\
        &= -\frac{1}{N\tau}\sum_{n = 1}^N \del{A^{(k)}_{n,n'}}{\psi_{\forall, l}} + \frac{1}{T_N \tau}\sum_{n = 1}^{2N} \sum_{\substack{m=1\\m\neq n,n'}}^{2N} \del{}{\psi_{\forall, l}} \left(e^{\frac{C_{n,m}}{\tau}} A^{(k)}_{n,m}\right)\\
    \end{split}
\end{equation}

where $n' = n+N$. Now, let us first calculate $\del{\sum_{i = 1}^{Do}z^{(i)}_n\cdot z^{(i)}_m}{\psi_{\forall, l}}$ (a row vector).

\begin{equation}
	\label{eqn:delcossim2}
	\begin{split}
		\del{\sum_{i = 1}^{Do}z^{(i)}_n\cdot z^{(i)}_m}{\psi_{\forall, l}} &= \sum_{i = 1}^{Do} \left(z^{(i)}_n \cdot \del{z^{(i)}_m}{\psi_{\forall, l}} + z^{(i)}_m \cdot \del{z^{(i)}_n}{\psi_{\forall, l}} \right)\\
		=& \sum_{i = 1}^{Do} \left(z^{(i)}_n \cdot \del{\left((\psi_{\forall, i})^T\cdot h_{\theta m}\right)}{\psi_{\forall, l}} + z^{(i)}_m \cdot \del{\left((\psi_{\forall, i})^T\cdot h_{\theta n}\right)}{\psi_{\forall, l}}\right)\\
		=& \sum_{i = 1}^{Do} \left(z^{(i)}_n \cdot \del{\left(h_{\theta m}^T \cdot \psi_{\forall, i}\right)}{\psi_{\forall, l}} +  z^{(i)}_m \cdot \del{\left(h_{\theta n}^T \cdot \psi_{\forall, i}\right)}{\psi_{\forall, l}}\right)\\
		=& \sum_{i = 1}^{Do} \left(z^{(i)}_n \cdot h_{\theta m}^T \cdot \del{ \psi_{\forall, i}}{\psi_{\forall, l}} + z^{(i)}_m \cdot h_{\theta n}^T \cdot \del{\psi_{\forall, i}}{\psi_{\forall, l}} \right)
		= z^{(l)}_n \cdot h_{\theta m}^T +  z^{(l)}_m \cdot h_{\theta n}^T\\
	\end{split}
\end{equation}

Let us denote $\del{\sum_{i = 1}^{Do}z^{(i)}_n\cdot z^{(i)}_m}{\psi_{\forall, l}} = z^{(l)}_n \cdot h_{\theta m}^T +  z^{(l)}_m \cdot h_{\theta n}^T$ by $A^{(l)T}_{n,m}$, whose dimension is $1 \times D_i$.\\

Now, let us separately evaluate, $\del{}{\psi_{\forall, l}} \left(e^{\frac{C_{n,m}}{\tau}}A^{(k)}_{n,m}\right)$ first to make our life easier.
\begingroup
\allowdisplaybreaks
\begin{align*}
    \label{eqn:del2L1_2}
        &\del{}{\psi_{\forall, l}} \left(e^{\frac{C_{n,m}}{\tau}} A^{(k)}_{n,m}\right) \\
        =& e^{\frac{C_{n,m}}{\tau}} \del{A^{(k)}_{n,m}}{\psi_{\forall, l}} + A^{(k)}_{n,m} \del{}{\psi_{\forall, l}} e^{\frac{C_{n,m}}{\tau}}
        = e^{\frac{C_{n,m}}{\tau}} \del{A^{(k)}_{n,m}}{\psi_{\forall, l}} + \frac{1}{\tau}A^{(k)}_{n,m} e^{\frac{C_{n,m}}{\tau}}\del{ C_{n,m}}{\psi_{\forall, l}}\\
        &= e^{\frac{C_{n,m}}{\tau}} \del{A^{(k)}_{n,m}}{\psi_{\forall, l}} + \frac{1}{\tau} e^{\frac{C_{n,m}}{\tau}} \cdot A^{(k)}_{n,m} \cdot A^{(l)T}_{n,m} \stepcounter{equation}\tag{\theequation}\\
\end{align*}
\endgroup

The only thing left to calculate is $\del{A^{(k)}_{n,m}}{\psi_{\forall, l}}$. The final dimension of this quantity will be $D_i \times D_i$. Let us denote this quantity by $B^{(l)(k)}_{n,m}$.

\begin{equation}
    \label{eqn:delAknm}
    \begin{split}
        \del{A^{(k)}_{n,m}}{\psi_{\forall, l}} &= \del{\left(z^{(k)}_n \cdot h_{\theta m} +  z^{(k)}_m \cdot h_{\theta n} \right)}{\psi_{\forall, l}}\\
        &=  h_{\theta m} \cdot \del{z^{(k)}_n}{\psi_{\forall, l}}+  h_{\theta n} \cdot \del{z^{(k)}_m}{\psi_{\forall, l}}\\
        &= h_{\theta m} \cdot \del{\left((\psi_{\forall, k})^T \cdot h_{\theta n} \right)}{\psi_{\forall, l}} + h_{\theta n} \cdot \del{\left( (\psi_{\forall, k})^T \cdot h_{\theta m}\right)}{\psi_{\forall, l}} \\
        &= h_{\theta m} \cdot \del{\left(h_{\theta n}^T \cdot (\psi_{\forall, k}) \right)}{\psi_{\forall, l}} + h_{\theta n} \cdot \del{h_{\theta m}^T \cdot \left( (\psi_{\forall, k})^T \right)}{\psi_{\forall, l}} \\
        &= h_{\theta m} \cdot h_{\theta n}^T + h_{\theta n} \cdot h_{\theta m}^T \big|_{k = l} \text{  or  } 0 \big|_{k \neq l}\\
    \end{split}
\end{equation}

Therefore,
$B^{(l)(k)}_{n,m} = 0$ and $B^{(l)(l)}_{n,m} = h_{\theta m} \cdot h_{\theta n}^T + h_{\theta n} \cdot h_{\theta m}^T$.\\

Putting Eqn. \ref{eqn:del2L1_2} and \ref{eqn:delAknm} in Eqn. \ref{eqn:del2L1}, we get,

\begin{equation}
\label{eqn:del2L1_3}
    \begin{split}
        &\DDel{2}{\mathcal{L}_{v3}}{\psi_{\forall, l}}{(\psi_{\forall, k})^T}\\ =& -\frac{1}{N\tau}\sum_{n = 1}^N \del{A^{(k)}_{n,n+N}}{\psi_{\forall, l}} + \frac{1}{T_N \tau}\sum_{n = 1}^{2N} \sum_{\substack{m=1\\m\neq n,n+N}}^{2N} \del{}{\psi_{\forall, l}} \left(e^{\frac{C_{n,m}}{\tau}} A^{(k)}_{n,m}\right)\\
        =& \frac{1}{T_N \tau^2}\sum_{n = 1}^{2N} \sum_{\substack{m=1\\m\neq n,n+N}}^{2N}  e^{\frac{C_{n,m}}{\tau}} \cdot A^{(k)}_{n,m} \cdot A^{(l)T}_{n,m}\\
    \end{split}
\end{equation}

and, 

\begin{equation}
\label{eqn:del2L1_4}
    \begin{split}
        &\DDel{2}{\mathcal{L}_{v3}}{\psi_{\forall, k}}{(\psi_{\forall, k})^T} \\
        =& -\frac{1}{N\tau}\sum_{n = 1}^N \del{A^{(k)}_{n,n+N}}{\psi_{\forall k}} + \frac{1}{T_N \tau}\sum_{n = 1}^{2N} \sum_{\substack{m=1\\m\neq n,n+N}}^{2N} \del{}{\psi_{\forall k}} \left(e^{\frac{C_{n,m}}{\tau}}A^{(k)}_{n,m}\right)\\
        =& -\frac{1}{N\tau}\sum_{n = 1}^N B^{(k)(k)}_{n,n+N}+ \frac{1}{T_N \tau}\sum_{n = 1}^{2N} \sum_{\substack{m=1\\m\neq n,n+N}}^{2N} \left(e^{\frac{C_{n,m}}{\tau}} B^{(k)(k)}_{n,m} + \frac{1}{\tau} e^{\frac{C_{n,m}}{\tau}} \cdot A^{(k)}_{n,m} \cdot A^{(k)T}_{n,m} \right)\\
    \end{split}
\end{equation}

We can write the two equations \ref{eqn:del2L1_3} and \ref{eqn:del2L1_4}, in a single equation, in a general form, as follows,

\begin{equation}
\label{eqn:hess_lv3}
    \begin{split}
        \DDel{2}{\mathcal{L}_{v3}}{\psi_{\forall, l}}{(\psi_{\forall, k})^T} = &-\frac{1}{N\tau}\sum_{n = 1}^N B^{(l)(k)}_{n,n+N} + \frac{1}{T_N \tau}\sum_{n = 1}^{2N} \sum_{\substack{m=1\\m\neq n,n+N}}^{2N} \left(e^{\frac{C_{n,m}}{\tau}} B^{(l)(k)}_{n,m} + \frac{1}{\tau} e^{\frac{C_{n,m}}{\tau}} \cdot A^{(k)}_{n,m} \cdot A^{(l)T}_{n,m} \right)\\
    \end{split}
\end{equation}

where 
\begin{equation*}
    B^{(l)(k)}_{n,m} = 
    \begin{cases}
    0, \;\text{if} \; l=k\\
    h_{\theta m} \cdot h_{\theta n}^T + h_{\theta n} \cdot h_{\theta m}^T, \;\text{if}\; l\neq k
    \end{cases}
\end{equation*}

We took each row in the weight matrix as a single variable for ease of calculation. This results in the second derivative being a matrix. The terms $\DDel{2}{\mathcal{L}_{v3}}{\psi_{\forall, k}}{(\psi_{\forall, k})^T}$ and $\DDel{2}{\mathcal{L}_{v3}}{\psi_{\forall, l}}{(\psi_{\forall, k})^T}$ are matrices themselves. Each element in the resultant matrix corresponds to each second derivative element in $\DDel{2}{\mathcal{L}_{v3}}{\psi_{\forall, k}}{(\psi_{\forall, k})^T}$ or $\DDel{2}{\mathcal{L}_{v3}}{\psi_{\forall, l}}{(\psi_{\forall, k})^T}$, each with dimensions $D_i \times D_i$.

\noindent
\textbf{Expansion of $\Del{2}{\mathcal{L}_{v3}}{(\psi_{\forall, k})^T}$ and $\DDel{2}{\mathcal{L}_{v3}}{\psi_{\forall, l}}{(\psi_{\forall, k})^T}$}

\begin{equation}
    \label{eqn:dldkexpansion}
    \begin{split}
        \DDel{2}{\mathcal{L}_{v3}}{\psi_{\forall, l}}{(\psi_{\forall, k})^T} &= \begin{pmatrix} 
        \DDel{2}{\mathcal{L}_{v3}}{\psi_{1 l}}{\psi_{1 k}} & \ldots &\DDel{2}{\mathcal{L}_{v3}}{\psi_{1 l}}{\psi_{Di k}}\\ 
        \vdots & \ldots & \vdots\\ 
        \DDel{2}{\mathcal{L}_{v3}}{\psi_{i l}}{\psi_{1 k}} & \ldots & \DDel{2}{\mathcal{L}_{v3}}{\psi_{i l}}{\psi_{Di k}}\\ 
        \vdots & \ldots & \vdots\\ 
        \DDel{2}{\mathcal{L}_{v3}}{\psi_{Di l}}{\psi_{1 k}} & \ldots & \DDel{2}{\mathcal{L}_{v3}}{\psi_{Di l}}{\psi_{Di k}}\\ 
        \end{pmatrix}
    \end{split}
\end{equation}

\begin{equation}
    \label{eqn:dkdkexpansion}
    \begin{split}
        \DDel{2}{\mathcal{L}_{v3}}{\psi_{\forall k}}{(\psi_{\forall, k})^T} &= \begin{pmatrix} 
        \Del{2}{\mathcal{L}_{v3}}{\psi_{1 k}} & \ldots & \DDel{2}{\mathcal{L}_{v3}}{\psi_{1 k}}{\psi_{Di k}}\\ 
        \vdots & \ldots & \vdots\\ 
        \DDel{2}{\mathcal{L}_{v3}}{\psi_{j k}}{\psi_{1 k}} & \ldots & \DDel{2}{\mathcal{L}_{v3}}{\psi_{j k}}{\psi_{Di k}}\\ 
        \vdots & \ldots & \vdots\\ 
        \DDel{2}{\mathcal{L}_{v3}}{\psi_{Di k}}{\psi_{1 k}} & \ldots & \Del{2}{\mathcal{L}_{v3}}{\psi_{Di k}}\\ 
        \end{pmatrix}
    \end{split}
\end{equation}

Essentially, the Hessian matrix $\mathcal{H}$ should be a $\mathcal{N}_{\mathcal{P}} \times \mathcal{N}_{\mathcal{P}}$ matrix, where $\mathcal{N}_{\mathcal{P}}$ is the number of parameters in the model whose parameters are being optimized. The expression of the Hessian matrix $\mathcal{H}$ is as follows.

\begin{equation}
    \label{eqn:hessianexpansion}
    \begin{split}
        \mathcal{H}(\psi) &= \begin{pmatrix} 
        \Del{2}{\mathcal{L}_{v3}}{\psi_{1 1}} & \DDel{2}{\mathcal{L}_{v3}}{\psi_{1 1}}{\psi_{1 2}} & \ldots & \DDel{2}{\mathcal{L}_{v3}}{\psi_{1 1}}{\psi_{D_i D_o}}\\ 
        \DDel{2}{\mathcal{L}_{v3}}{\psi_{1 2}}{\psi_{1 1}} & \Del{2}{\mathcal{L}_{v3}}{\psi_{2 2}} & \ldots &  \DDel{2}{\mathcal{L}_{v3}}{\psi_{1 2}}{\psi_{D_i D_o}}\\ 
        \vdots & \vdots & \ldots & \vdots\\ 
        \DDel{2}{\mathcal{L}_{v3}}{\psi_{j k}}{\psi_{1 1}} & \DDel{2}{\mathcal{L}_{v3}}{\psi_{j k}}{\psi_{1 2}} & \ldots & \DDel{2}{\mathcal{L}_{v3}}{\psi_{j k}}{\psi_{D_i D_o}}\\ 
        \vdots & \vdots & \ldots & \vdots\\ 
        \DDel{2}{\mathcal{L}_{v3}}{\psi_{D_i D_o}}{\psi_{1 1}} & \DDel{2}{\mathcal{L}_{v3}}{\psi_{D_i D_o}}{\psi_{1 2}} & \ldots & \Del{2}{\mathcal{L}_{v3}}{\psi_{D_i D_o}}\\ 
        \end{pmatrix}
    \end{split}
\end{equation}

\subsection{Prove that the function $\mathcal{L}_{v3} \circ g_{\psi} \circ f_{\theta}$, which maps the input space $\chi$ to the loss landscape $\mathbb{L}$ has $L$-Lipschitz continuous gradient}
\label{subsubsec:prooflipschitzgrad}

To prove that the loss function $\mathcal{L}_{v3} \circ g_{\psi} \circ f_{\theta}$ has $L$-Lipschitz continuous gradient, we need to show that the spectral norm of the Hessian matrix $\mathcal{H}$ is upper bounded by $L$. We can also prove that,

\begin{equation}
\label{eqn:abc}
\begin{split}
&\lVert \nabla \mathcal{L}_{v3} \circ g_{\psi_{t+1}} \circ f_{\theta_{t+1}} (x) - \nabla \mathcal{L}_{v3} \circ g_{\psi_{t}} \circ f_{\theta_{t}} (x) \rVert \\ 
&= \lVert \nabla \mathcal{L}_{v3} \circ (g \circ f)_{\mathcal{P}_{t+1}} (x) - \nabla \mathcal{L}_{v3} \circ (g \circ f)_{\mathcal{P}_{t}} (x) \rVert \\
&\leq L \lVert \mathcal{P}_{t+1} - \mathcal{P}_t \rVert  \\
\end{split}
\end{equation}

We take a single element of the Hessian matrix $\mathcal{H}$, $\DDel{2}{\mathcal{L}_{v3}}{\psi_{\forall, l}}{(\psi_{\forall, k})^T}$ and analyse it analytically. Thus, 
\begin{equation}
\label{eqn:hess_bound1}
    \begin{split}
        \DDel{2}{\mathcal{L}_{v3}}{\psi_{\forall, l}}{(\psi_{\forall, k})^T} = &-\frac{1}{N\tau}\sum_{n = 1}^N B^{(l)(k)}_{n,n+N} + \frac{1}{T_N \tau}\sum_{n = 1}^{2N} \sum_{\substack{m=1\\m\neq n,n+N}}^{2N} \left(e^{\frac{C_{n,m}}{\tau}} B^{(l)(k)}_{n,m} + \frac{1}{\tau} e^{\frac{C_{n,m}}{\tau}} \cdot A^{(k)}_{n,m} \cdot A^{(l)T}_{n,m} \right)\\
    \end{split}
\end{equation}

Since the exponential terms exist in the Hessian terms, we can say that the function $(\mathcal{L}_{v3} \circ g_{\psi} \circ f_{\theta}) (x)$ belongs to the class of $C^{\infty}$ functions, provided $\sum_{w \in \mathcal{P}_t} w < \infty$. It remains to be proven, that the norm of the Hessian matrix is bounded by the Lipschitz constant $L$ or the above Eqn. \ref{eqn:abc} holds true.

Without loss of generality, we can say that the space of gradients and parameters belongs to an $\mathbb{D}$-dimensional real vector space. It can be proved empirically that with every different initialization $\mathcal{P}_0$, the endpoint $\mathcal{P}_T$ differs, as performs the model. Since the parameter space is a real space, we can say that the sequence $\{\mathcal{P}^{(1)}_0,\mathcal{P}^{(1)}_1,\hdots,\mathcal{P}^{(1)}_T\}$ obtained with seed $s_1$ is disjoint from the sequence $\{\mathcal{P}^{(2)}_0,\mathcal{P}^{(2)}_1,\hdots,\mathcal{P}^{(2)}_T\}$ obtained with seed $s_2$, where $s_1 \neq s_2$. Thus, we can say that the vector space of parameters $\mathcal{P}$ is a Hausdorff Topological Vector Space with the canonical metric $d = \lVert \cdot \rVert$ of a normed space $(X, \lVert \cdot \rVert)$.

Since any two sequences of parameters on $\mathbb{R}^\mathbb{D}$ is disjoint, the gradient space associated with the sequences will also be disjoint on $\mathbb{R}^\mathbb{D}$. Hence, the gradient space defined on $\mathbb{R}^\mathbb{D}$ is also a Hausdorff Topological Vector Space with the canonical metric $d = \lVert \cdot \rVert$ of a normed space $(X, \lVert \cdot \rVert)$.

Since the parameter space $\mathcal{P}$ and the gradient space $\mathcal{G}$ are both Hausdorff spaces, the sequence converges to a point in the same respective space. In other words, the sequence $(\mathcal{P})_{t=1}^\infty$ and $(\mathcal{G})_{t=1}^\infty$ converges to some $\mathcal{P}_{t=\infty} \in \mathcal{P}$ and $\mathcal{G}_{t=\infty} \in \mathcal{G}$, respectively. The aforementioned statement implies that the parameter space $\mathcal{P}$ and the gradient space $\mathcal{G}$ are Banach spaces.

We can also view the above statement in another way. Since the composite function $(\mathcal{L}_{v3} \circ g_{\psi} \circ f_{\theta}) (x)$ belongs to the class of $C^\infty$ functions, then under the constraint that the inputs to the Linear layers $\psi^{(1)}$ and $\psi^{(2)}$ are finite, i.e., $\sum_d h^{(d)}_{\theta n} < \infty$, and $\sum_{w \in \mathcal{P}} w < \infty$, the gradients values obtained using Eqn. \ref{eqn:delL1_3} is also finite. Thus the change in consecutive values of a sequence in $\mathcal{P}$, i.e, $\lVert \mathcal{P}_{t+1} - \mathcal{P}_t \rVert < r_{\mathcal{P}}$, where $r_{\mathcal{P}} > 0$ and $r_{\mathcal{P}} \in \mathbb{R}$. Therefore, the sequence of parameters $(\mathcal{P}_t)_{t=1}^\infty$ can be called to be Cauchy in $(\mathcal{P}, \lVert \cdot \rVert)$. Thus, we can conclude that the parameter space $\mathcal{P}$ is a Banach space. 

Similarly, since the gradients are finite, we can say that, $\lVert \nabla \mathcal{L}_{v3} \circ (g \circ f)_{\mathcal{P}_{t+1}} (x) - \nabla \mathcal{L}_{v3} \circ (g \circ f)_{\mathcal{P}_{t}} (x) \rVert < r_g$, where $r_g > 0$ and $r_g \in \mathbb{R}$. Therefore, denoting $\nabla \mathcal{L}_{v3} \circ (g \circ f)_{\mathcal{P}_{t+1}} (x)$ by $\mathcal{G}_{t+1}$ and $\nabla \mathcal{L}_{v3} \circ (g \circ f)_{\mathcal{P}_{t}} (x)$ by $\mathcal{G}_t$ we can proceed as,
\begin{equation}
    \label{eqn:218proof}
    \begin{split}
        \lVert \mathcal{G}_{t+1} - \mathcal{G}_t \rVert &< r^{(t)}_g
        \leq L_g \cdot \lVert \nabla \mathcal{L}_{v3} \circ (g \circ f)_{\mathcal{P}_{t}} (x) \rVert
        \leq L^{'}_g \lVert \mathcal{P}_{t+1} - \mathcal{P}_t \rVert\\
    \end{split}
\end{equation}

where $\mathcal{G}$ is a point in the gradient space $\mathbb{G}$ and $\mathcal{G}_t \in \mathbb{G}$ denotes the state of the gradient at timestep $t$.

Thus, the composite function approximated by $\mathcal{L}_{v3} \circ g_{\psi} \circ f_{\theta} : \chi \rightarrow \mathbb{R}$ has a $L^{'}_{g}$-Lipschitz continuous gradient, under the constraint that $\sum_d h^{(d)}_{\theta n} < \infty$, and $\sum_{w \in \mathcal{P}} w < \infty$. In other words, the above discussion indicates that the aforementioned function has locally Lipschitz continuous gradient when the weights are initialized with weights from a \textcolor{black}{normal distribution}. A different proof to arrive at the same conclusion is also provided in Lemma 2.3 in the work by V. Patel, et al. \cite{nogloballip}.


\textbf{Summary}
\begin{itemize}
    \item The Equations \ref{eqn:dldkexpansion} and \ref{eqn:dkdkexpansion} covers the second derivative of every possible parameter present in the weight matrix $\psi$ of the projector $g_{\psi}$.
    \item Combining Equations \ref{eqn:del2L1_3}, \ref{eqn:del2L1_4}, \ref{eqn:dldkexpansion} and \ref{eqn:dkdkexpansion}, we can get the diagonal and off-diagonal elements in the Hessian matrix.
    \item The Hessian matrix is dependent on the weights $\psi$ only through the term $A^{(k)}_{n,m}$ and $A^{(k)T}_{n,m}$.
    \item The existence of the Hessian matrix indicates that the loss $\mathcal{L}_{v3}$ and the mapping function $\mathcal{L}_{v3} \circ g_{\psi} \circ f_{\theta}: \chi \rightarrow \bs{\bbR}$ is twice differentiable, under the constraints $\sum_d h^{(d)}_{\theta n} < \infty$, and $\sum_{w \in \mathcal{P}} w < \infty$.
    \item The parameter space $\mathcal{P}$ being a Banach space also indicates that the end-to-end pipeline mapping function $\mathcal{L}_{v3} \circ g_{\psi} \circ f_{\theta} : \chi \rightarrow \mathbb{R}$ has $L^{'}_{g}$-\textbf{Lipschitz} continuous gradient.
\end{itemize}

\section{Convergence on Non-Convex Functions}

\subsection{Polyak-Lojasiewicz Inequality}

From Hamed et al. \cite{hamed2016linconv}, we can state, for an unconstrained optimization problem,
\begin{equation}
\label{eqn:pl1}
    \text{argmin}_{x \in \mathbb{R}^d} f(x)
\end{equation}

where $f$ is a function with $L$-Lipschitz continuous gradient, we have 

\begin{equation}
\label{eqn:pl2}
    \centering
    f(y) \leq f(x) + \langle \nabla f(x), y-x \rangle + \frac{L}{2} \lVert x-y \rVert ^2 \; \forall \; x,y    
\end{equation}

If $f$ belongs to the class of $C^2$ functions, the eigenvalues of $\nabla^2 f(x)$ are bounded above by $L$, which is called the Lipschitz constant. We also assume that the solution set $\mathcal{X}^* \neq \varnothing$ and $f^*$ is the optimal function value. The Polyak-Lojasiewicz inequality is satisfied if for $\mu > 0$,
\begin{equation}
\label{eqn:pl3}
    \frac{1}{2} \lVert \nabla f(x) \rVert ^2 \geq \mu (f(x) - f^*), \; \forall \; x
\end{equation}

Applying gradient descent with step size $\frac{1}{L}$,
\begin{equation}
\label{eqn:pl4}
    x_{k+1} = x_k - \frac{1}{L} \nabla f(x_k)
\end{equation}

From Eq. \ref{eqn:pl2}, we get,

\begin{equation}
    \label{eqn:pl5}
    \begin{split}
    f(x_{k+1}) &\leq f(x_k) + \langle \nabla f(x_k), x_{k+1}-x_k \rangle + \frac{L}{2} \lVert x_{k+1}-x_k \rVert ^2\\
     &\leq f(x_k) - \frac{1}{2L} \lVert \nabla f(x_k) \rVert ^2\\
     &\leq f(x_k) - \frac{\mu}{L} \left(f(x_k) - f^*\right)\\
    \end{split}
\end{equation}

Subtracting $f^*$ from both sides of Eq. \ref{eqn:pl5}, we get,
\begin{equation}
    \centering
    \label{eqn:pl6}
    \begin{split}
        f(x_{k+1}) - f^* \leq \left(1 - \frac{\mu}{L}\right) \left(f(x_k) - f^*\right)
    \end{split}
\end{equation}

Applying Eqn. \ref{eqn:pl6} recursively, we get, the global linear convergence rate as follows,
\begin{equation}
    \centering
    \label{eqn:pl7}
    \begin{split}
        f(x_{k+1}) - f^* \leq \left(1 - \frac{\mu}{L}\right)^k \left(f(x_0) - f^*\right)
    \end{split}
\end{equation}


\subsection{Convergence of SGD on Non-Convex Functions}

To calculate the rate of convergence of SGD on non-convex functions, we follow the derivation steps followed in \cite{blog}. To see how SGD evolves over time, we take $x = w_t, y = w_{t+1}$ in Eqn. \ref{eqn:pl2}, which yields,

\begin{equation}
\label{eqn:pl101}
    \centering
    \begin{split}
        f(w_{t+1}) \leq f(w_t) + \langle \nabla f(w_t), w_{t+1}-w_t \rangle + \frac{L}{2} \lVert w_t -w_{t+1} \rVert ^2 \\
    \end{split}
\end{equation}

Now, we assume that for the samples indicated by the indices $\xi$, we have an oracle that gives us the gradient $g(w,\xi) \in \mathbb{R}^d$ at the point $w$, where $\xi$ is a random index of a training sample used to calculate the training loss. We also assume that the variance of the stochastic gradient is bounded by as $\mathbb{E}_{\xi}\left[ \lVert \nabla f(w) - g(w,\xi) \rVert_2^2 \right] \leq \sigma^2 < \infty$ for all $w \in dom \nabla f(w)$. 

For SGD, the parameter update proceeds as $w_{t+1} = w_t - \eta_t g(w_t, \xi_t)$, where $\eta_t$ is the learning rate at time step $t$. Now, going back to Eqn. \ref{eqn:pl101}, and putting $w_{t+1} - w_t = -\eta_t g(w_t,\xi_t)$, we get,
\begin{equation}
\label{eqn:pl102}
    \centering
    \begin{split}
        f(w_{t+1}) \leq f(w_t) - \eta_t \langle \nabla f(w_t), g(w_t, \xi_t) \rangle + \eta_t^2\frac{L}{2} \lVert g(w_t, \xi_t) \rVert_2^2
    \end{split}
\end{equation}

Taking expectation with respect to $\xi_t$, keeping $w_t$ constant, we get

\begin{equation}
\label{eqn:pl103}
    \centering
    \begin{split}
        \mathbb{E}_{\xi_t} \left[f(w_{t+1})\right] &\leq \mathbb{E}_{\xi_t} \left[f(w_t)\right] - \eta_t \mathbb{E}_{\xi_t} \left[\langle \nabla f(w_t), g(w_t, \xi_t) \rangle\right] + \eta_t^2\frac{L}{2} \mathbb{E}_{\xi_t} \left[\lVert g(w_t, \xi_t) \rVert_2^2\right]\\
        \implies f(w_{t+1}) &= f(w_t) - \eta_t \lVert \nabla f(w_t) \rVert_2^2 + \eta_t^2\frac{L}{2} \mathbb{E}_{\xi_t} \left[\lVert g(w_t, \xi_t) \rVert_2^2\right]\\
        &= f(w_t) - \eta_t \lVert \nabla f(w_t) \rVert_2^2 + \eta_t^2\frac{L}{2} \mathbb{E}_{\xi_t} \left[\lVert \nabla f(w_t) + g(w_t, \xi_t) - \nabla f(w_t) \rVert_2^2\right]\\
        &=f(w_t) - \eta_t \lVert \nabla f(w_t) \rVert_2^2 + \eta_t^2\frac{L}{2} \left(\mathbb{E}_{\xi_t} \left[\lVert g(w_t, \xi_t) - \nabla f(w_t)\rVert_2^2 \right] + \lVert \nabla f(w_t) \rVert_2^2 \right)\\
        &= f(w_t) - \left(\eta_t-\frac{\eta_t^2 L}{2} \right) \lVert \nabla f(w_t) \rVert_2^2 + \eta_t^2\frac{L}{2} \mathbb{E}_{\xi_t} \left[\lVert g(w_t, \xi_t) - \nabla f(w_t)\rVert_2^2 \right]\\
        &\leq f(w_t) - \left(\eta_t-\frac{\eta_t^2 L}{2} \right) \lVert \nabla f(w_t) \rVert_2^2 + \eta_t^2\frac{L}{2} \sigma^2\\
    \end{split}
\end{equation}

In the last line of the above equation, we have used the fact that the variance of the stochastic gradient is bounded above by $\sigma^2$. Taking the total expectation and reordering terms, we get,

\begin{equation}
\label{eqn:pl104}
    \centering
    \begin{split}
        &\sum_{t=1}^T \left(\eta_t-\frac{\eta_t^2 L}{2} \right) \mathbb{E}_t \left[ \lVert\nabla f(w_t) \rVert_2^2\right]\\ &\leq \sum_{t=1}^T \left( \mathbb{E}_t \left[f(w_t) \right] -\mathbb{E}_t \left[ f(w_{t+1}) \right]\right) + \frac{\sigma^2 L}{2} \sum_{t=1}^T \eta_t^2\\
        &\leq \mathbb{E}_t \left[ f(x_1) \right ] - \mathbb{E}_t \left[ f(w_{T+1}) \right] + \frac{\sigma^2 L}{2} \sum_{t=1}^T \eta_t^2\\
        &\leq f(x_1) - f^* + \frac{\sigma^2 L}{2} \sum_{t=1}^T \eta_t^2\\
    \end{split}
\end{equation}

\subsubsection{Optimization with Constant Step Size}
\label{subsubsec:conststep}

The optimization process continues as long as $\eta_t < \frac{1}{L}$, where $L$ is the Lipschitz constant. 
The left-hand side of the above equation will be maximized for $\eta_t = \frac{1}{L} = \eta$. Hence, putting that value, we have $\eta_t - \frac{\eta_t^2 L}{2} = \eta_t - \frac{\eta_t}{2} = \frac{\eta_t}{2}$. Putting this expression in Eqn. \ref{eqn:pl104}, we get


\begin{equation}
    \centering
    \label{eqn:pl105}
    \begin{split}
        \sum_{t=1}^T \frac{\eta_t}{2}\mathbb{E}_t \left[ \lVert\nabla f(w_t) \rVert_2^2\right] 
        &\leq f(x_1) - f^* + \frac{\sigma^2 L}{2} T \eta_t^2\\
        \implies \frac{1}{T}\sum_{t=1}^T \mathbb{E}_t \left[ \lVert\nabla f(w_t) \rVert_2^2\right] 
        &\leq 
        \frac{2}{\eta_t T} (f(x_1) - f^*) + \sigma^2 L \eta_t \\
        &= \frac{2L}{T} (f(x_1) - f^*) + \sigma^2\\
    \end{split}
\end{equation}



We get an almost convergence result, as the average of the norm of the gradients goes to zero at $\mathcal{O}(\frac{1}{T})$. This means that we can expect the algorithm to make fast progress at the beginning of the optimization and then slowly converge once the number of iterations becomes big enough compared to the variance of the stochastic gradients. In case the noise on the gradients is zero, SGD becomes simply gradient descent and it will converge at a rate of $\mathcal{O}(\frac{1}{T})$.

\subsubsection{Optimization with Time-Varying Step Size}
\label{subsubsec:consttimevar}

Let us consider again Eqn. \ref{eqn:pl104}, but with a time-varying learning rate, 

\begin{equation}
\label{eqn:pl106}
\begin{split}
    \sum_{t=1}^{\infty} \eta_t = \infty \text{ and } \sum_{t=1}^{\infty} \eta_t^2 < \infty\\
\end{split}
\end{equation} 

The above conditions ensure that $\eta_t \to 0$ as $t \to \infty$. \cite{optmetlsml}.

With such a choice, we get,

\begin{equation}
\label{eqn:pl107}
    \centering
    \begin{split}
        &\sum_{t=1}^T \left(\eta_t-\frac{\eta^2 L}{2} \right) \mathbb{E}_t \left[ \lVert\nabla f(w_t) \rVert_2^2\right]
        \leq f(w_1) - f^* + \frac{\sigma^2 L}{2} \sum_{t=1}^T \eta_t^2 < \infty\\
    \end{split}
\end{equation}

Now,
$\sum_{t=1}^{T} \eta_t^2 < \infty \implies \eta_T \rightarrow 0$. So, there exists $T_L$ such that $\eta_t - \frac{\eta_t^2 L}{2} \geq \frac{\eta_t}{2}$ for all $t \geq T_L$. Hence,

\begin{equation}
    \centering
    \label{eqn:pl108}
    \begin{split}
        \sum_{t = T_L}^{\infty} \eta_t \mathbb{E}_t \left[ \lVert\nabla f(w_t) \rVert_2^2\right] < \infty
    \end{split}
\end{equation}

This implies that $\sum_{t=T_L}^{\infty} \eta_t \lVert \nabla f(w_t) \rVert_2^2 < \infty$ with probability 1. From this last inequality and the condition $\sum_{t=1}^{\infty} \eta_t = \infty$, we can derive that $lim \; inf_{t \rightarrow \infty} \; \lVert \nabla f(w_t) \rVert_2 = 0$.

Unfortunately, it seems that we proved something weaker than we wanted to. In words, the \textit{lim inf} result says that there exists a subsequence of $w_t$ that has a gradient converging to zero.

\begin{mlemma}
\label{lemma:l1}
{
Let $(b_t)_{t\geq1},(\eta_t)_{t\geq1}$ be two non-negative sequences and $(a_t)_{t\geq 1}$ a sequence of vectors in a vector space $X$. Let $p \geq 1$ and assume $\sum_{t=1}^{\infty} \eta_t b_t^p < \infty$ and $\sum_{t=1}^{\infty} \eta_t = \infty$. Assume also that there exists $L \geq 0$ such that $|b_{t+\tau} - b_t| \leq L(\sum_{i=t}^{t+\tau-1} \eta_i b_i + \lVert \sum_{i=t}^{t+\tau-1} \eta_i a_i \rVert)$, where $a_t$ is such that $\lVert \sum_{i=1}^{\infty} \eta_t a_t \rVert < \infty$. Then, $b_t$ converges to $0$. [Lemma A.5 in \cite{mairal}, Extension of Proposition 2 in \cite{alber}]
}
\end{mlemma}

Using the above Lemma on $b_t = \lVert \nabla f(w_t) \rVert$, we observe that by the $L$-smoothness of $f$, we have,

\begin{equation}
    \centering
    \label{eqn:pl109}
    \begin{split}
        &\lVert \nabla f(w_{t+\tau}) \rVert - \lVert \nabla f(x_{t}) \rVert \leq \lVert \nabla f(w_{t+\tau}) - \nabla f(w_t) \rVert\\
        & \leq L \lVert w_{t+\tau} - w_t \rVert = L \lVert \sum_{i=t}^{t+\tau-1} \eta_i g(x_i, \xi_i) \rVert\\
        & = L \lVert \sum_{i=t}^{t+\tau-1} \eta_i \left(\nabla f(x_i) + g(x_i, \xi_i) - \nabla f(x_i) \right) \rVert\\
        & \leq L \sum_{i=t}^{t+\tau-1} \eta_i \lVert \nabla f(x_i) \rVert + L \lVert \sum_{i=t}^{t+\tau-1} \eta_i \left( g(x_i, \xi_i) - \nabla f(x_i) \right) \rVert\\
    \end{split}
\end{equation}

The assumptions and the reasoning above imply that, with probability 1, $\sum_{t=1}^{\infty} \eta_t \lVert \nabla f(w_t) \rVert < \infty$. This also suggest to set $a_t = g(x_i, \xi_i) - \nabla f(x_i)$. Also, we have, with probability 1, $\lVert \sum_{t=1}^{\infty} \eta_t a_t  \rVert < \infty$, because $\sum_{t=1}^T \eta_t a_t$ for $T = 1,2,...$ is a martingale, i.e., the conditional expectation of the next value in the sequence is equal to the present value, regardless of all prior values. The variance is also bounded by $\sigma^2 \sum_{t=1}^{\infty} \eta_t^2 < \infty$. Hence, $\sum_{t=1}^T \eta_t a_t$ for $T=1,2,...$ is a martingale in $L^2$, so it converges in $L^2$ with probability 1. Overall, with probability 1, the assumptions of Lemma \ref{lemma:l1} are verified with $p = 2$. 

Finally, we proved that the gradients of SGD do indeed converge to zero with probability 1. This means that with probability 1 for any $\epsilon>0$ there exists $N_\epsilon$ such that $\nabla f(w_t) \leq \epsilon$ for $t\geq N_\epsilon$.\\

\subsubsection{Step Size with Cosine Annealing Decay}

Before proceeding further, we explore the results for step sizes decaying according to a cosine annealing schedule, $\eta_t = \eta_{min} + \frac{1}{2}(\eta_{max} - \eta_{min})\left(1 + \cos\left( \frac{t}{T}\pi\right) \right)$ with $\eta_{min} = 0$. Using cosine annealing step size schedule, 

\begin{equation}
\label{eqn:cos1}
\begin{split}
    \sum_{t=1}^{\infty} \eta_t \to \infty \text{ and } \sum_{t=1}^{\infty} \eta_t^2 < \infty\\
\end{split}
\end{equation} 

Thus, the criteria for convergence still holds for cosine annealing step size schedule. Also, $\sum_{t=1}^{\infty} \eta_t$ grows faster than $\sum_{t=1}^{\infty} \eta^2_t$  However, for finite training periods, $\eta_t = 0$ for $t > T$ . Hence, under this condition, we will always have, $\sum_{t = T}^{\infty} \eta_t \mathbb{E}_t \left[ \lVert\nabla f(w_t) \rVert_2^2\right] = 0 < \infty$ and $\sum_{t = 0}^{T} \eta_t \mathbb{E}_t \left[ \lVert\nabla f(w_t) \rVert_2^2\right] < \infty$. As mentioned before, the aforementioned statement gives rise to a very weak condition for convergence.

In other words, the parameter space $\mathcal{P}$ and the gradient space $\mathcal{G}$, both being a Hausdorff space with complete normed metric $\lVert \cdot \rVert$, the sequence of parameters $(\mathcal{P})_{t=1}^\infty$ converge within a Ball of radius $r \in \mathbb{R}^\mathbb{D}$. Hence, under the assumption of global $L$-Lipschitz continuity, i.e., $\sum_d h_{\theta n} < \infty$ and $\sum_{w \in \mathcal{P}} w < \infty$, we can infer that $\lVert \nabla f(x) \rVert_{t = T} \leq \epsilon$ for $\epsilon > 0$.

\subsubsection{Empirical Observation: }

\begin{figure*}[!ht]
    \centering
    \subfloat[][Eigenvalue spectrum of SimCLR on CIFAR10 dataset, pre-trained for 10 epochs]{\includegraphics[width=0.33\linewidth]{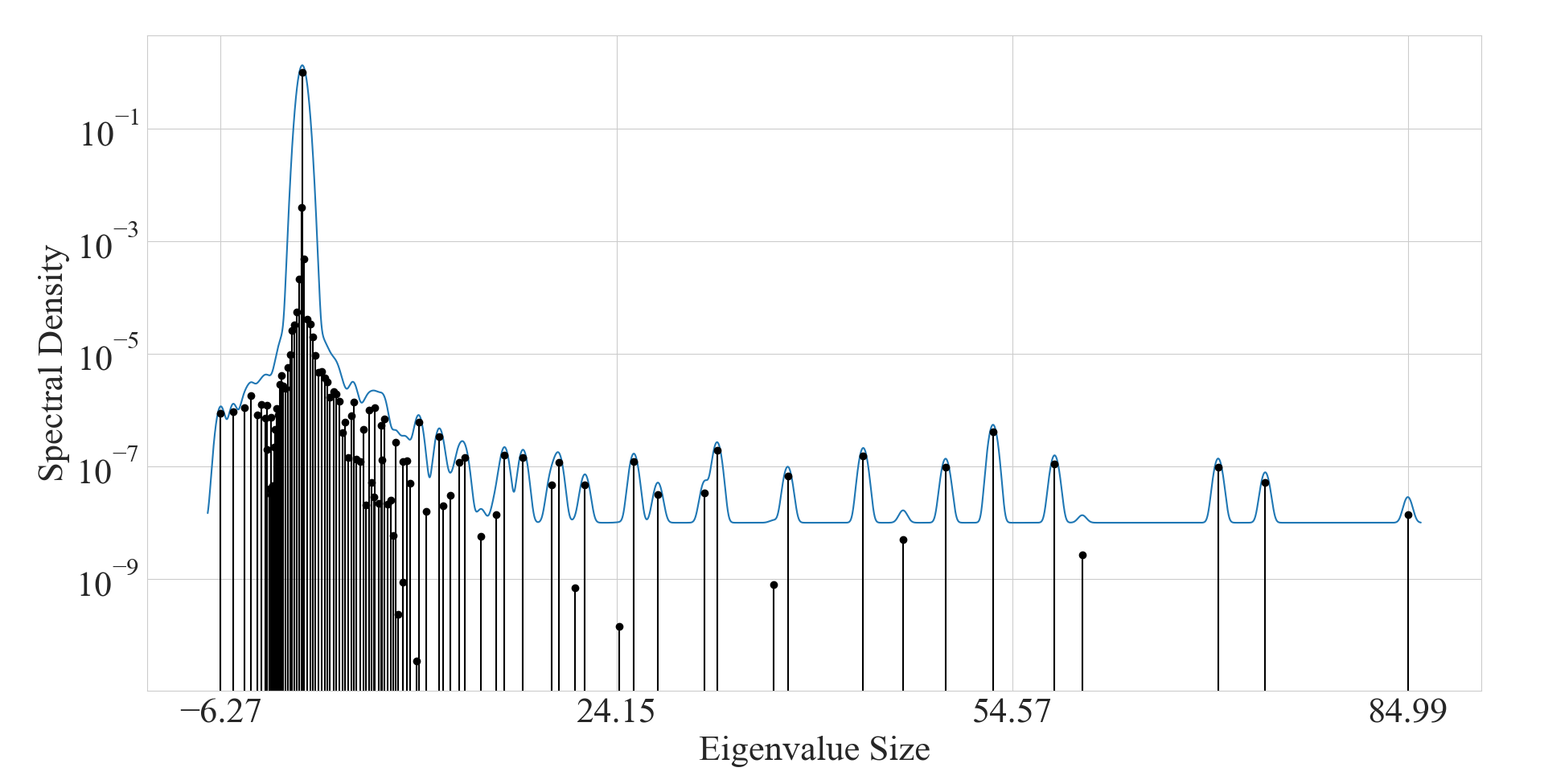}\label{fig:sc10_r18_10eps}}
    \hfill
     \subfloat[][Eigenvalue spectrum of DCL on CIFAR10 dataset, pre-trained for 10 epochs]{\includegraphics[width=0.33\linewidth]{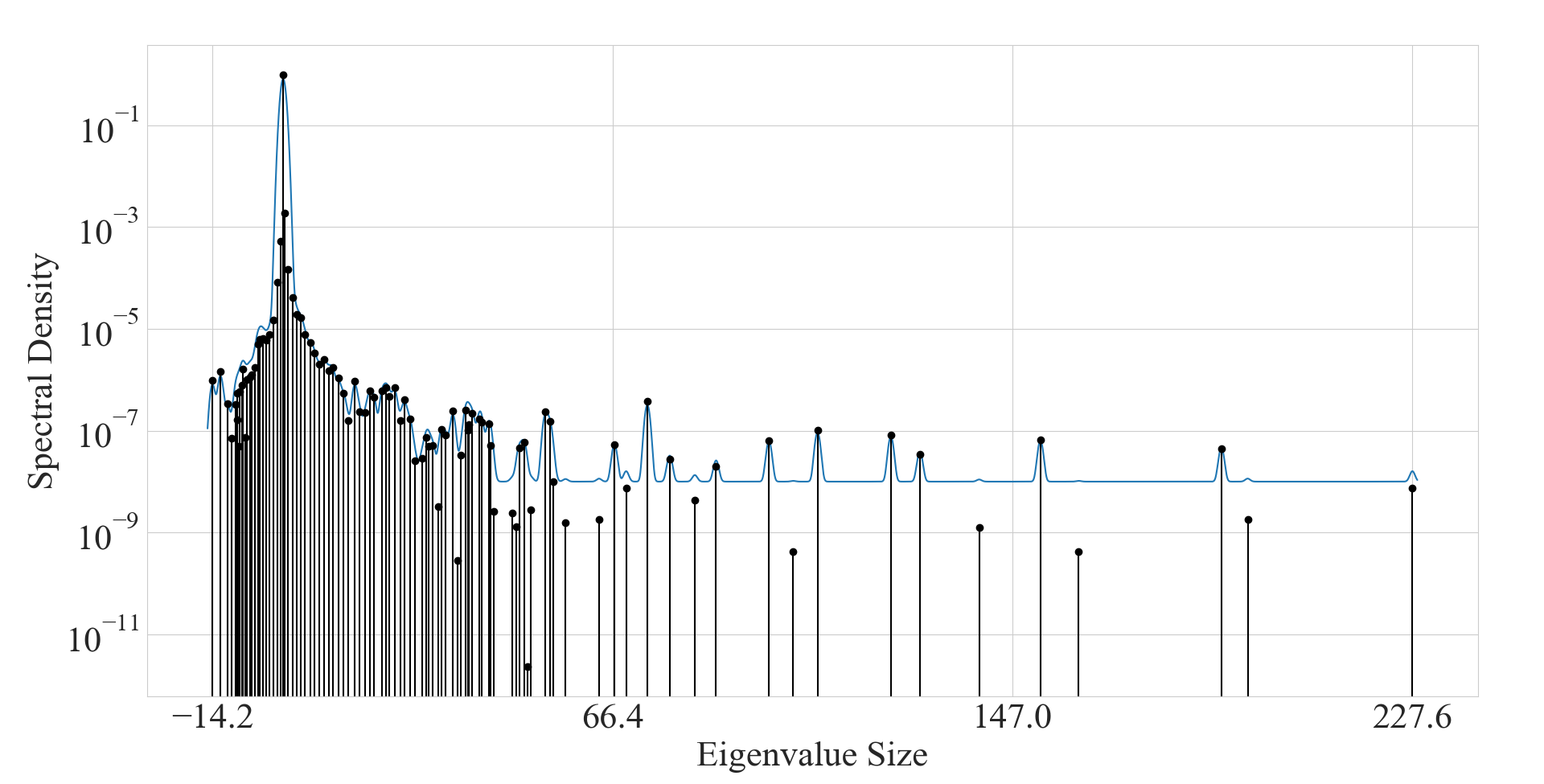}\label{fig:dc10_r18_10eps}}
     \hfill
     \subfloat[][Eigenvalue spectrum of MIOv3 on CIFAR10 dataset, pre-trained for 10 epochs]{\includegraphics[width=0.33\linewidth]{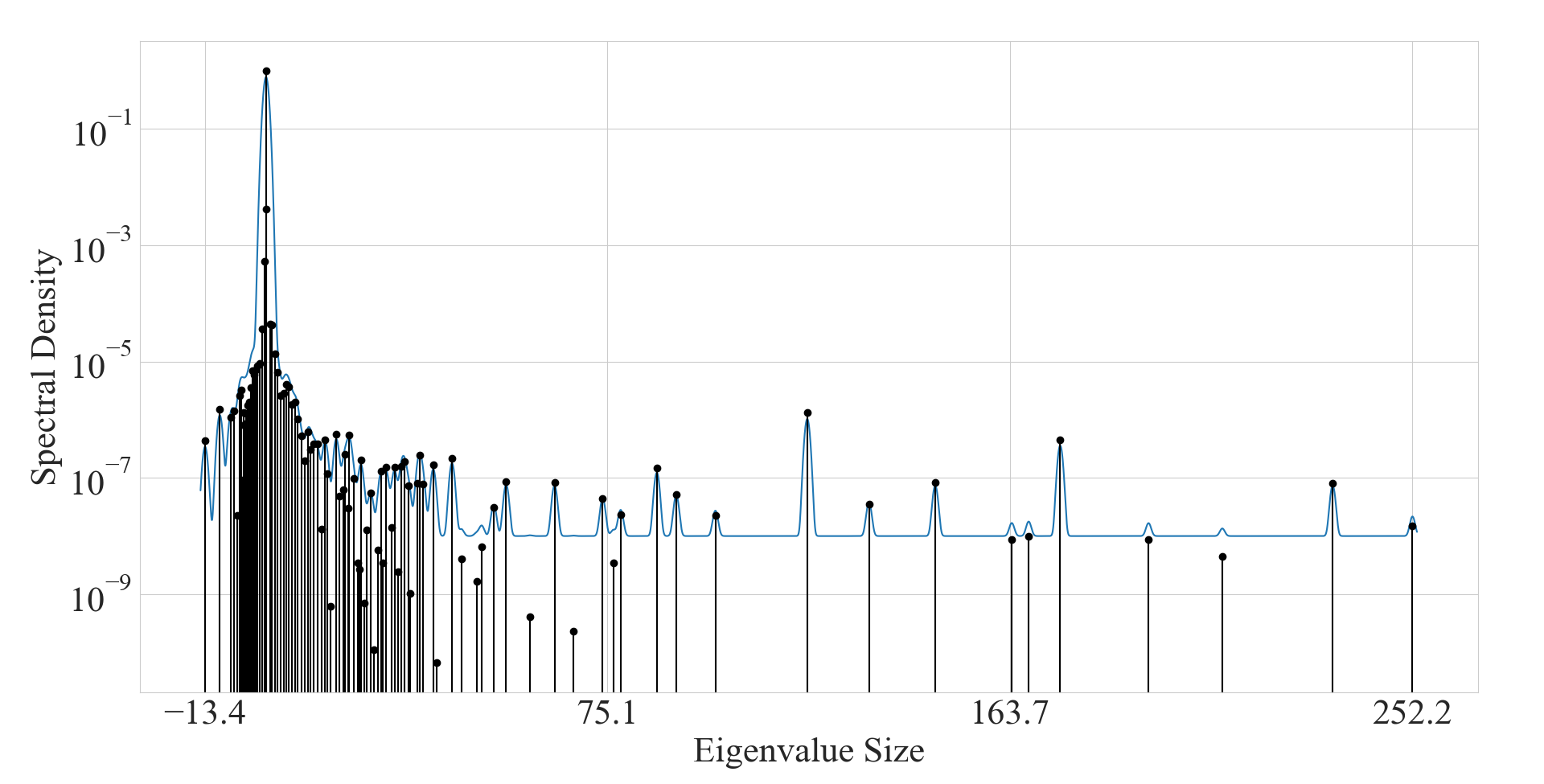}\label{fig:mc10_r18_10eps}}
     \qquad
     \subfloat[][Eigenvalue spectrum of SimCLR on CIFAR100 dataset, pre-trained for 10 epochs]{\includegraphics[width=0.33\linewidth]{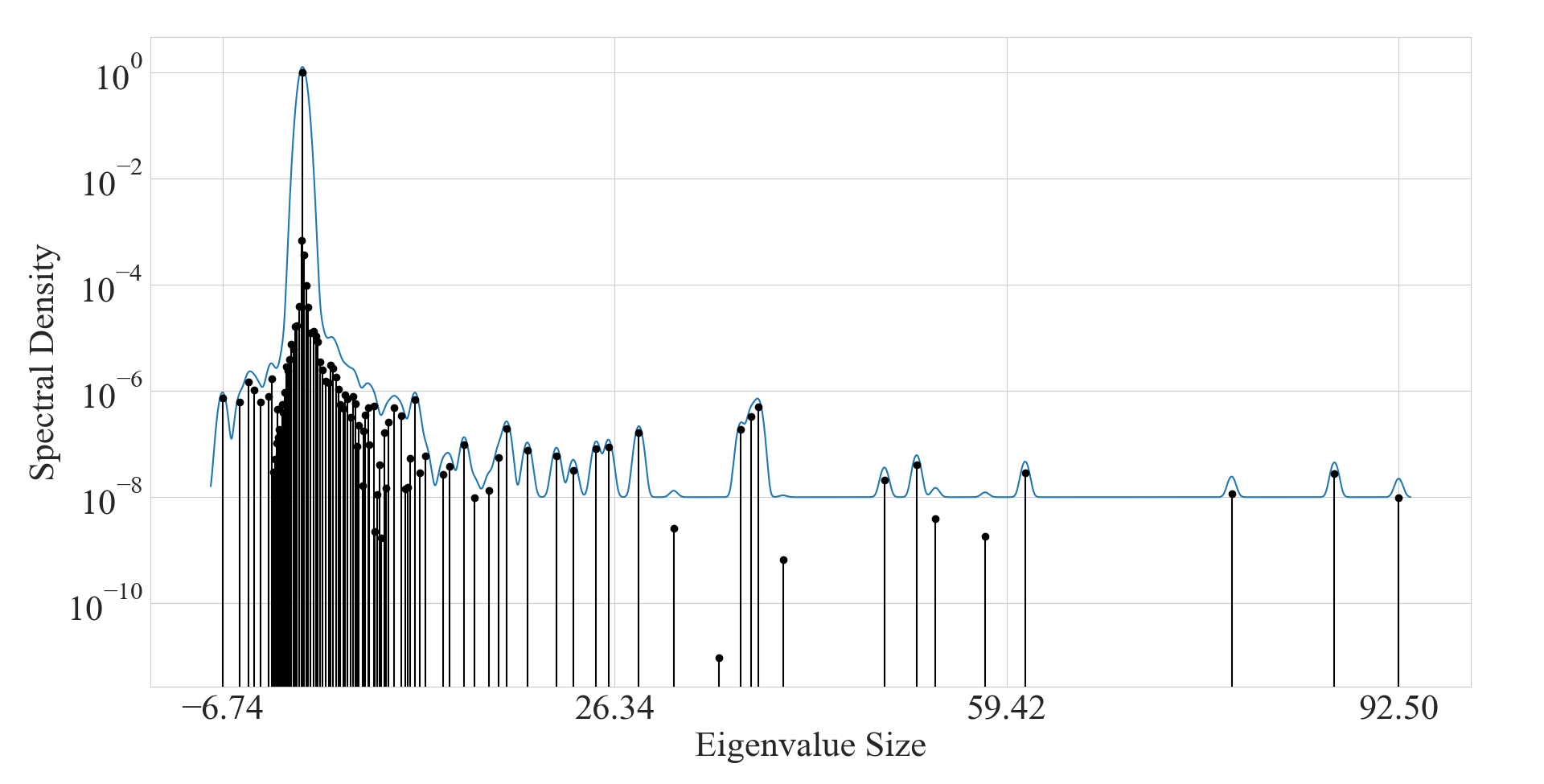}\label{fig:sc100_r18_10eps}}
     \hfill
     \subfloat[][Eigenvalue spectrum of DCL on CIFAR100 dataset, pre-trained for 10 epochs]{\includegraphics[width=0.33\linewidth]{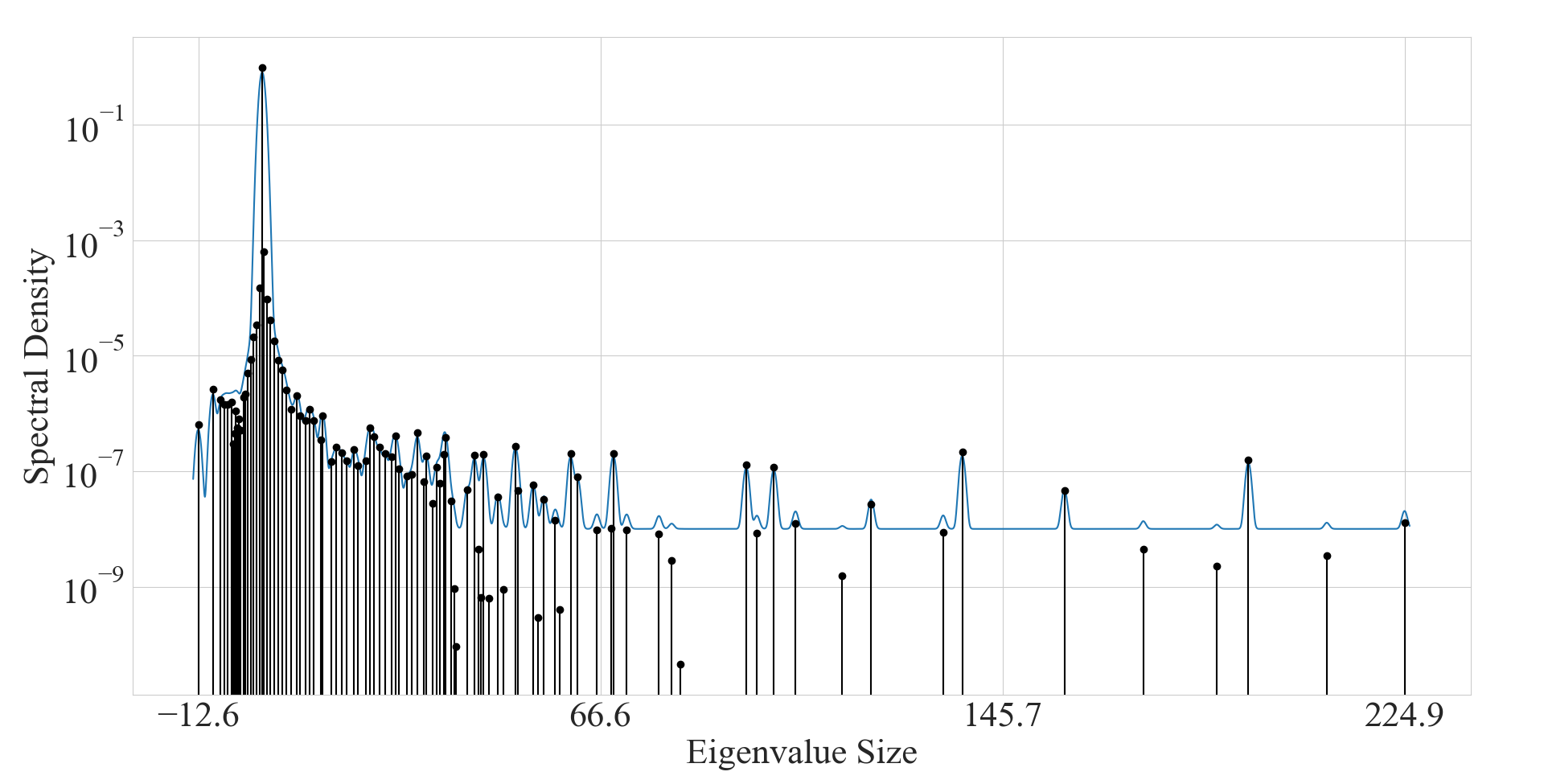}\label{fig:dc100_r18_10eps}}
     \hfill
     \subfloat[][Eigenvalue spectrum of MIOv3 on CIFAR100 dataset, pre-trained for 10 epochs]{\includegraphics[width=0.33\linewidth]{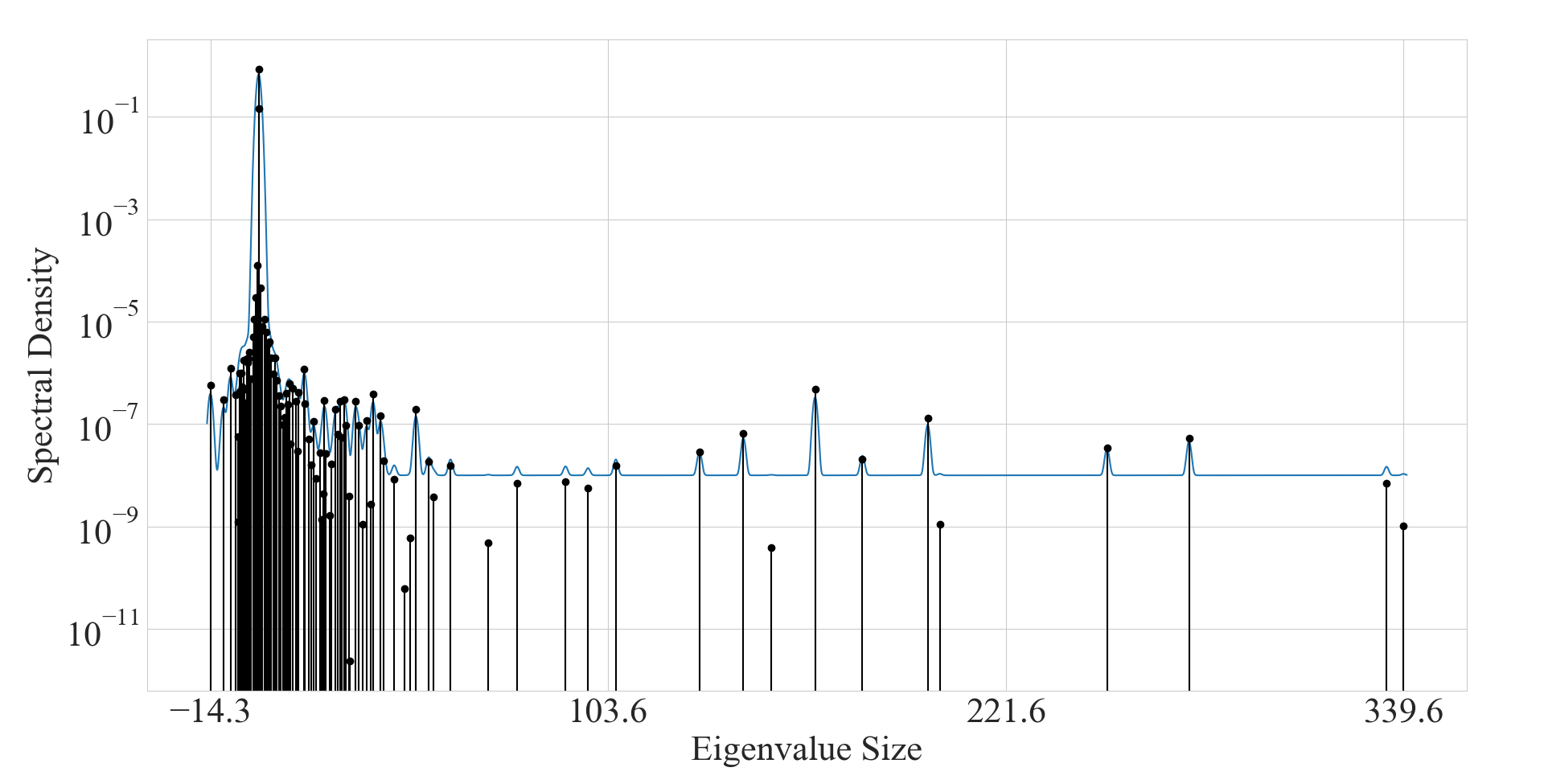}\label{fig:mc100_r18_10eps}}
    \caption{Plot of eigenvalues of parameters of ResNet18, obtained after 10 epochs of pre-training on CIFAR10 and CIFAR100 datasets with different SSL frameworks, namely, SimCLR, DCL and MIOv3.}
    \label{fig:alleigsnew10eps}
\end{figure*}

\begin{figure*}[!ht]
    \centering
    \subfloat[][Eigenvalue spectrum of SimCLR on CIFAR10 dataset, pre-trained for 100 epochs]{\includegraphics[width=0.33\linewidth]{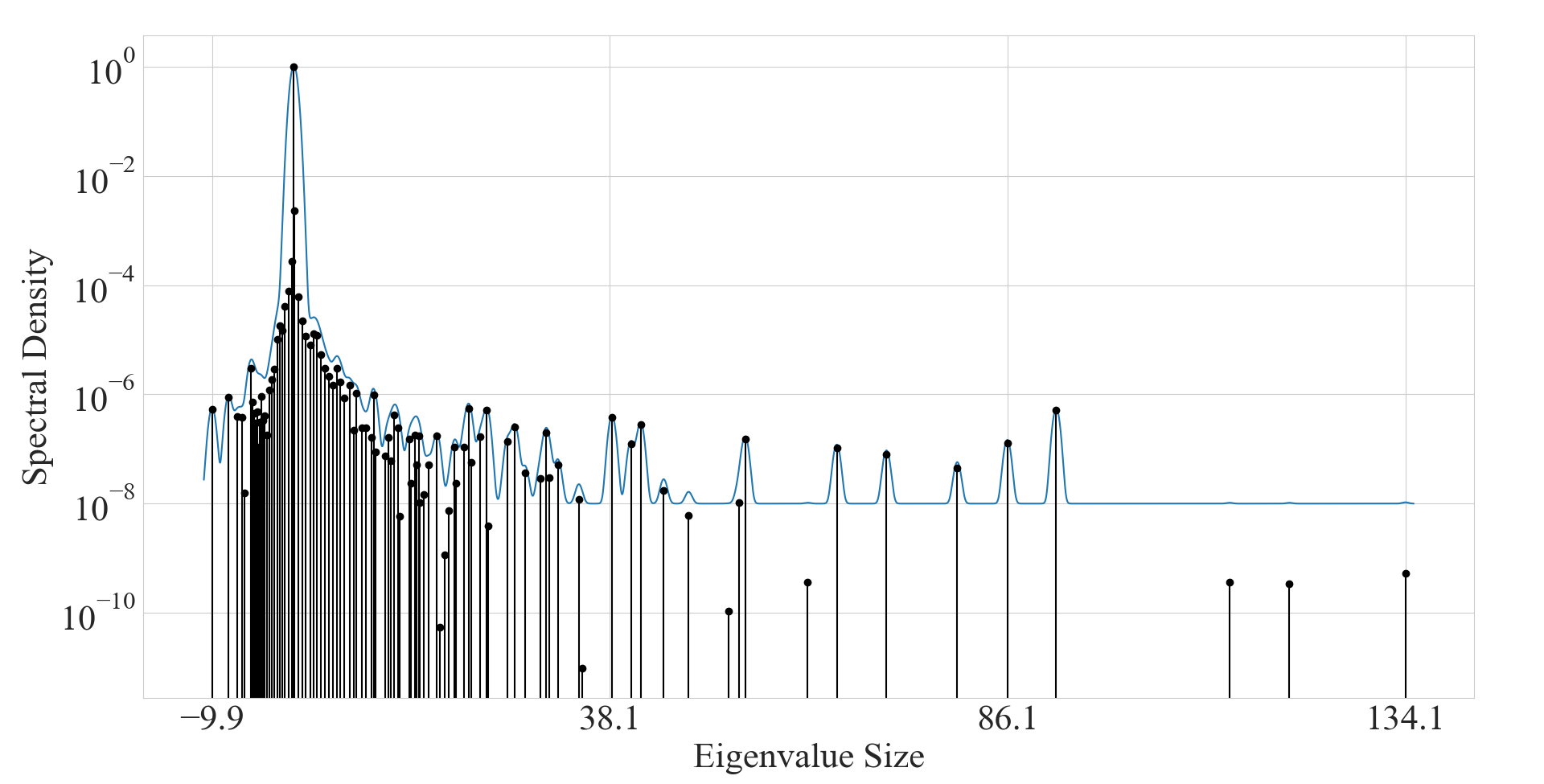}\label{fig:sc10_r18_100eps}}
    \hfill
     \subfloat[][Eigenvalue spectrum of DCL on CIFAR10 dataset, pre-trained for 100 epochs]{\includegraphics[width=0.33\linewidth]{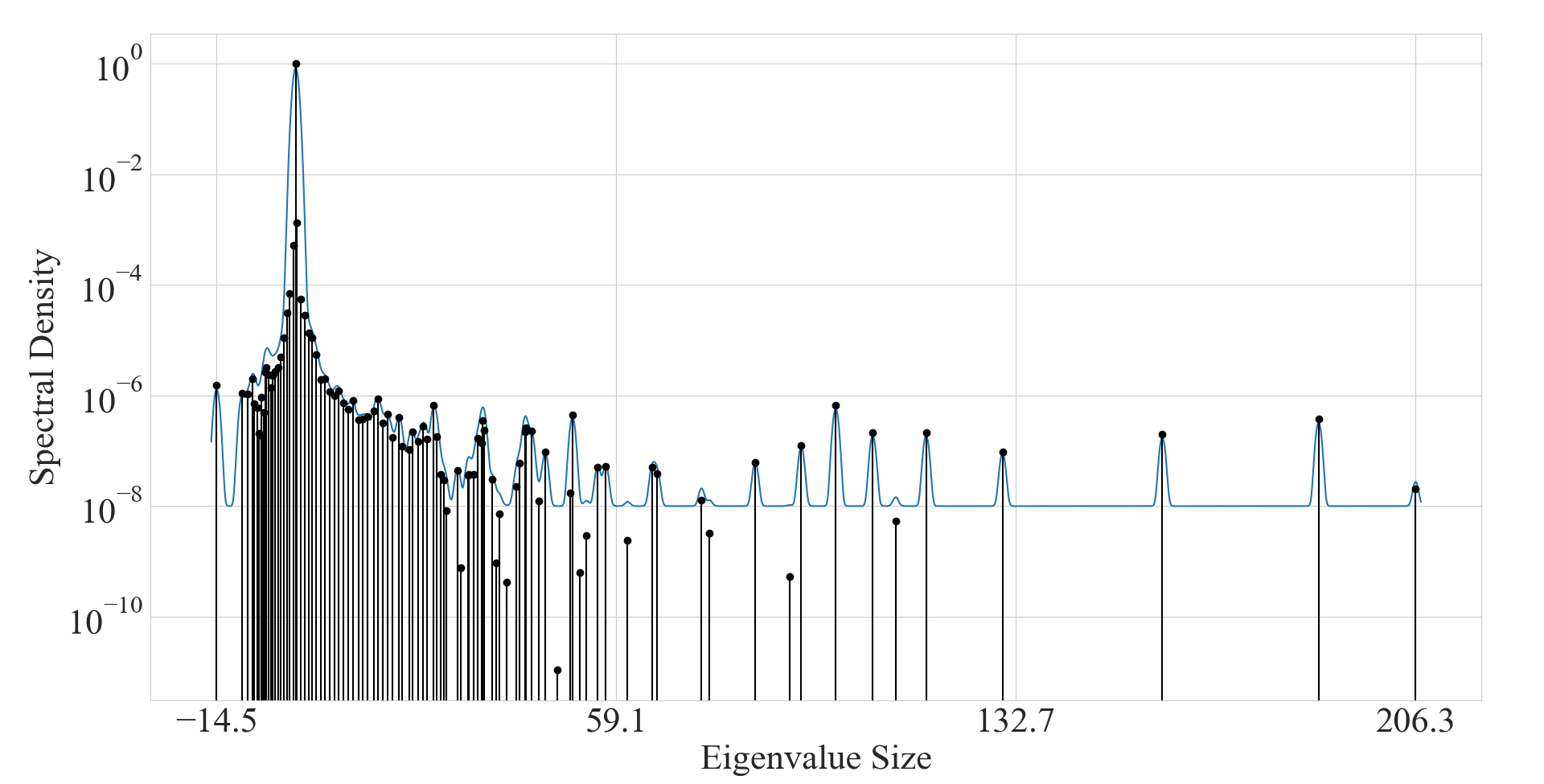}\label{fig:dc10_r18_100eps}}
     \hfill
     \subfloat[][Eigenvalue spectrum of MIOv3 on CIFAR10 dataset, pre-trained for 100 epochs]{\includegraphics[width=0.33\linewidth]{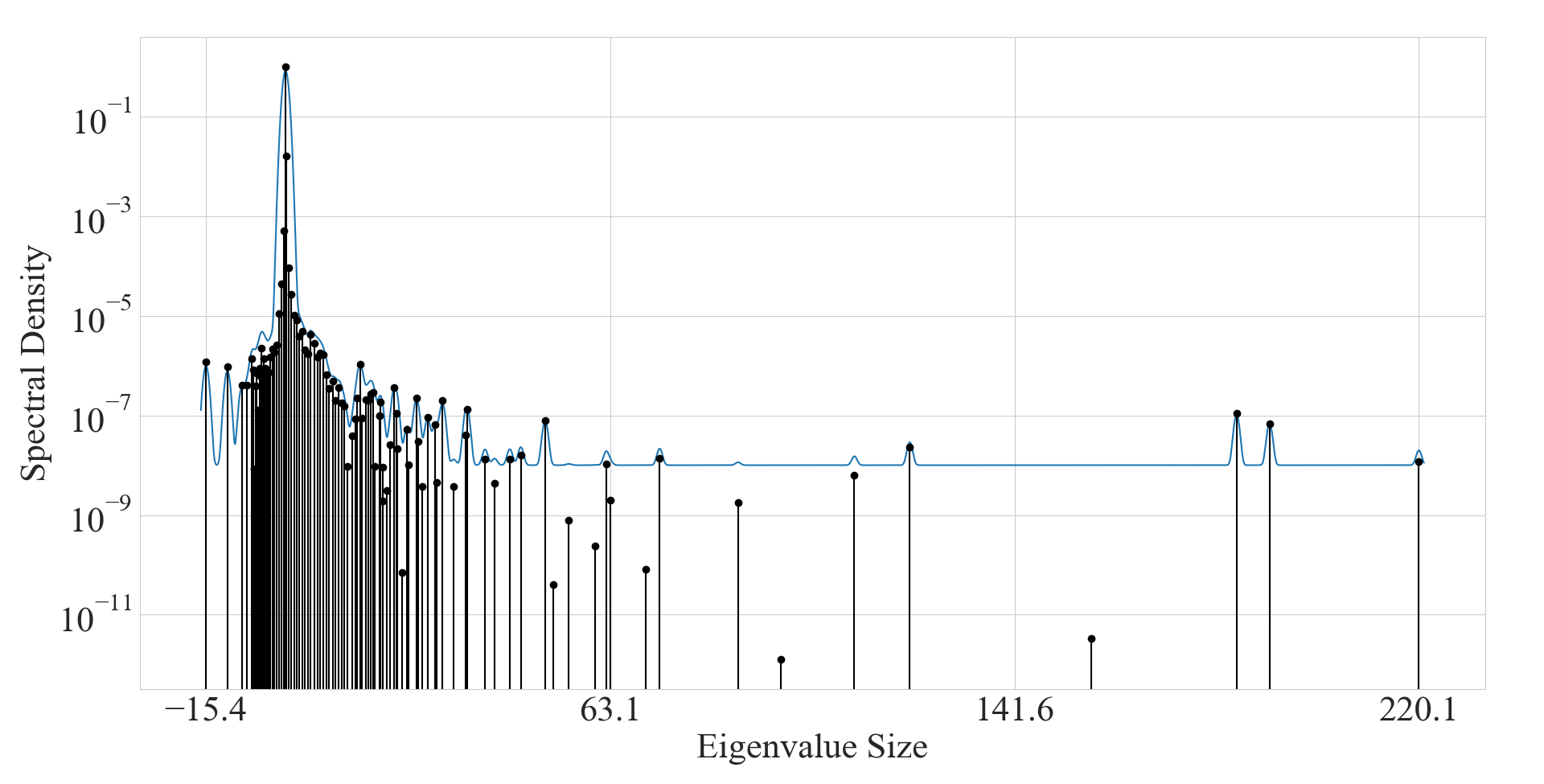}\label{fig:mc10_r18_100eps}}
     \qquad
     \subfloat[][Eigenvalue spectrum of SimCLR on CIFAR100 dataset, pre-trained for 100 epochs]{\includegraphics[width=0.33\linewidth]{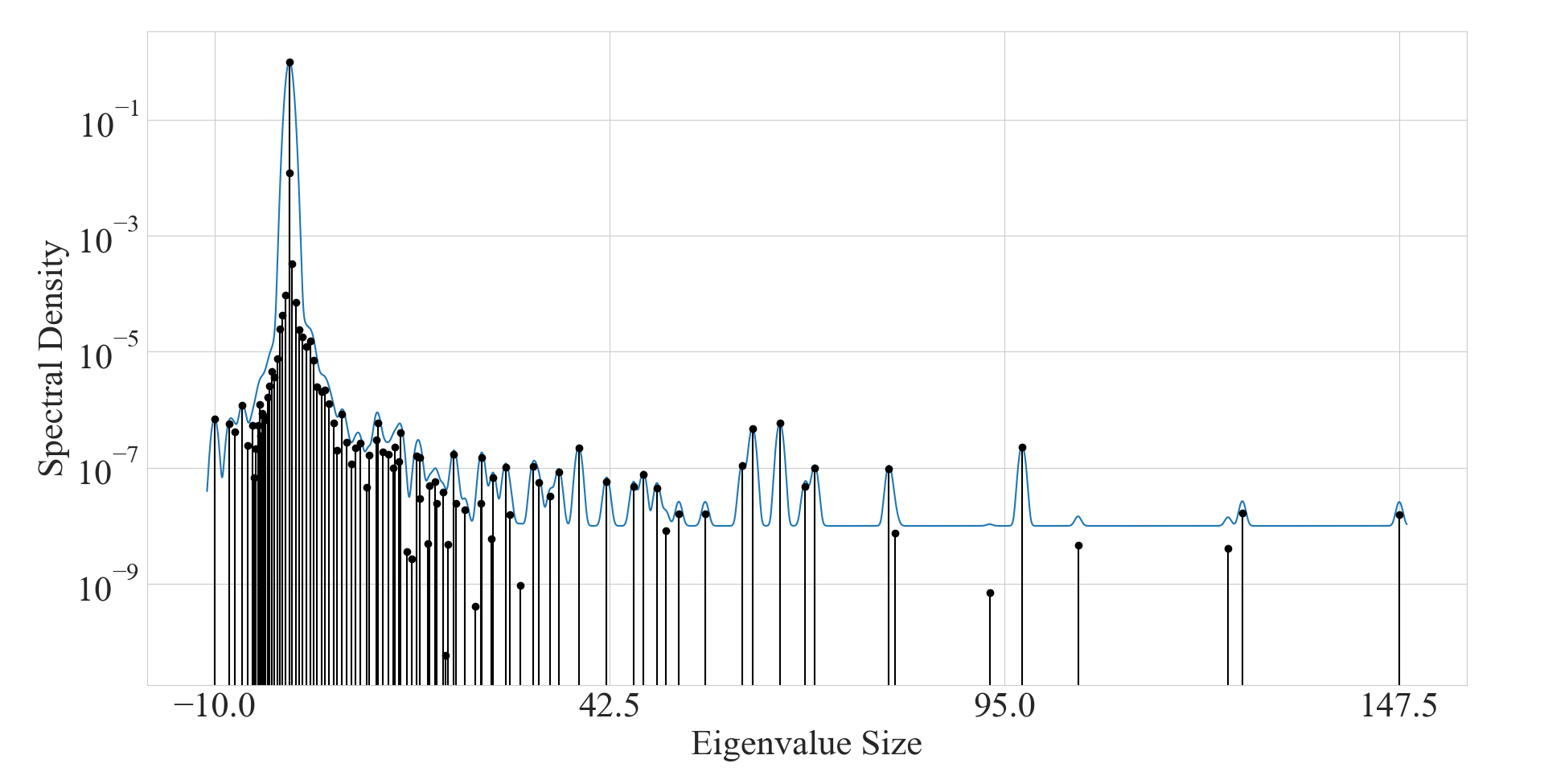}\label{fig:sc100_r18_100eps}}
     \hfill
     \subfloat[][Eigenvalue spectrum of DCL on CIFAR100 dataset, pre-trained for 100 epochs]{\includegraphics[width=0.33\linewidth]{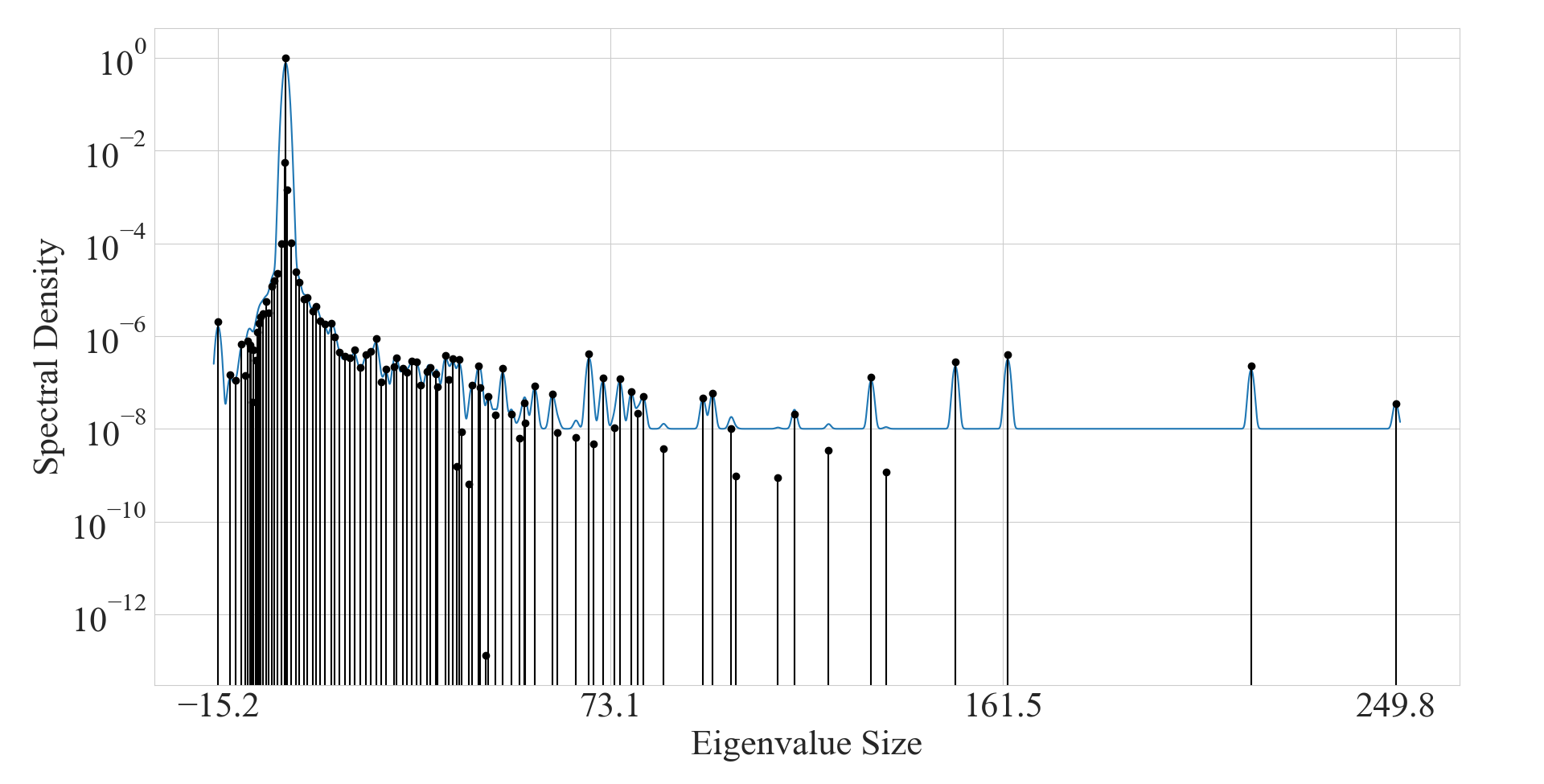}\label{fig:dc100_r18_100eps}}
     \hfill
     \subfloat[][Eigenvalue spectrum of MIOv3 on CIFAR100 dataset, pre-trained for 100 epochs]{\includegraphics[width=0.33\linewidth]{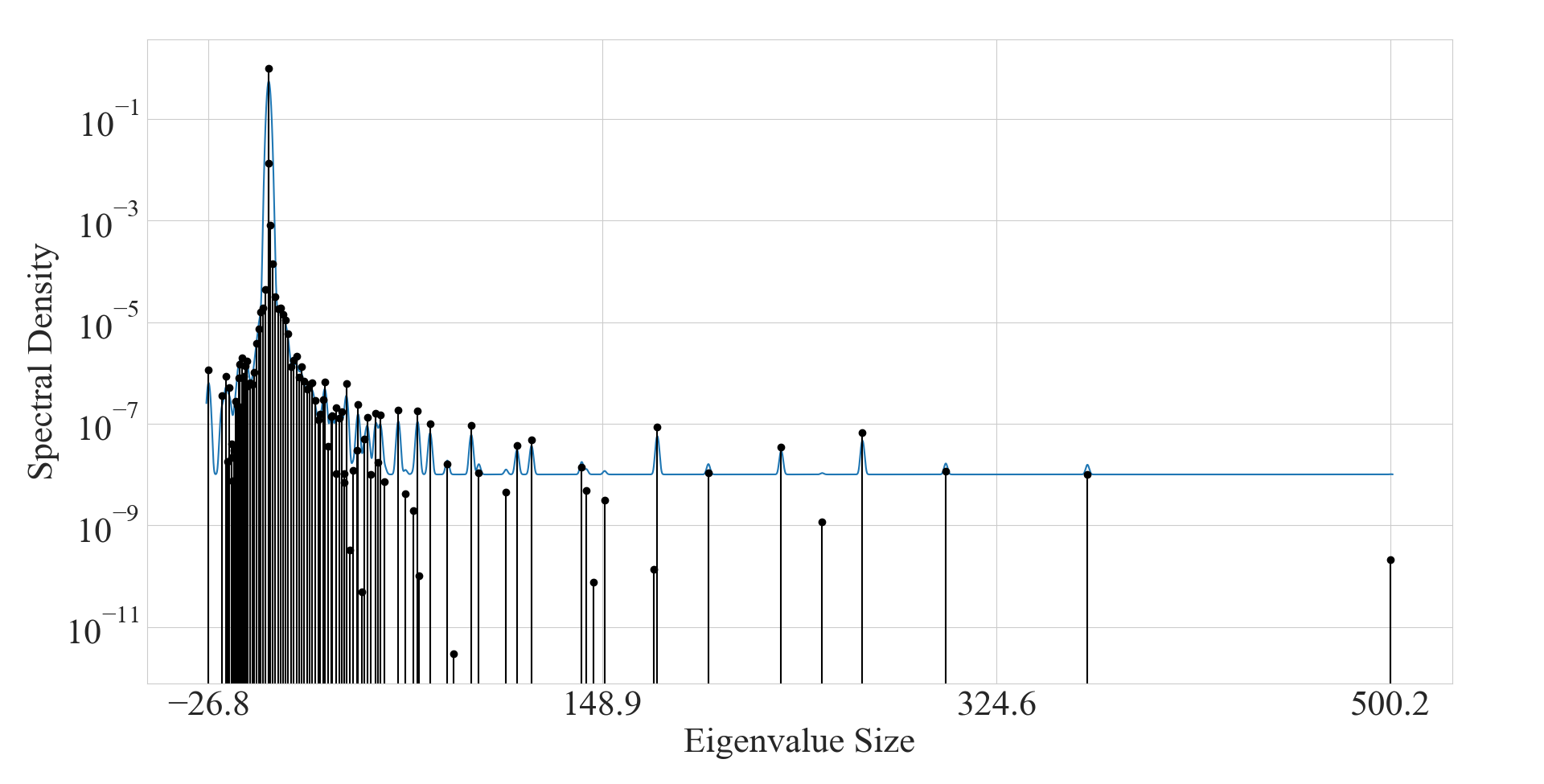}\label{fig:mc100_r18_100eps}}
    \caption{Plot of eigenvalues of parameters of ResNet18, obtained after 100 epochs of pre-training on CIFAR10 and CIFAR100 datasets with different SSL frameworks, namely, SimCLR, DCL and MIOv3.}
    \label{fig:alleigsnew100eps}
\end{figure*}

From the eigenspectrum plots in the manuscript and in Fig. \ref{fig:alleigsnew10eps} and \ref{fig:alleigsnew100eps}, we can observe that the eigen-spectrum consists of both positive and non-positive eigenvalues, at 10th and 100th epochs of pre-training. The eigenspectrum at the 200th epoch is already presented in the main manuscript (Fig. 3). This indicates that convergence does not actually occur after 200 epochs of training in SSL frameworks. Although the norm of the gradient $\lVert \nabla f(x) \rVert_2$ may be non-zero, the decrease in step size causes the parameter update to converge to zero. Hence, a \textit{premature convergence} occurs at a strict saddle point.

\begin{figure*}[!ht]
    \centering
    \subfloat[][Eigenvalue spectrum of SimCLR on CIFAR10 dataset, pre-trained for 10 epochs]{\includegraphics[width=0.33\linewidth]{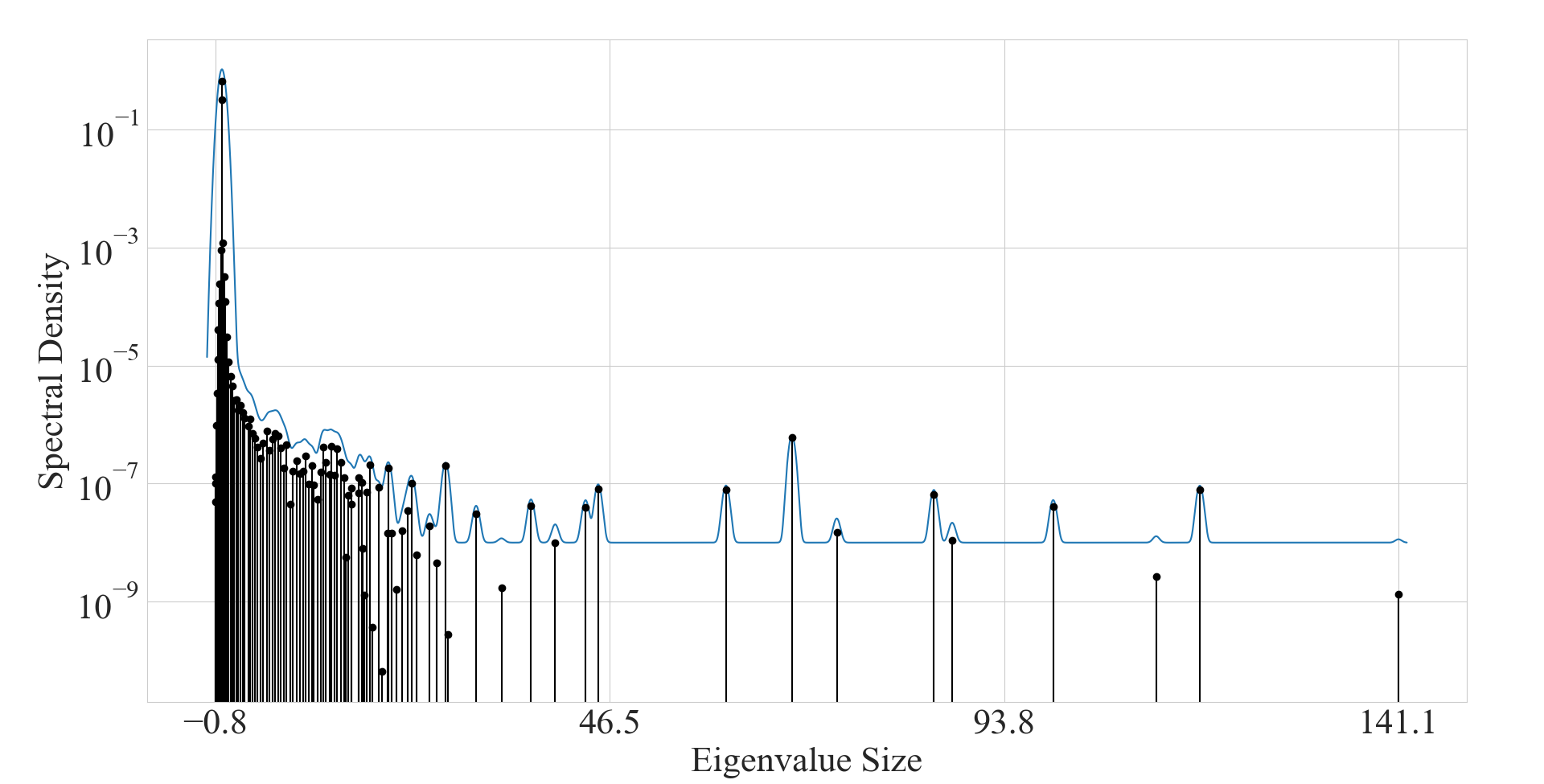}\label{fig:sc10_r18_100epscls10}}
    \hfill
     \subfloat[][Eigenvalue spectrum of DCL on CIFAR10 dataset, pre-trained for 10 epochs]{\includegraphics[width=0.33\linewidth]{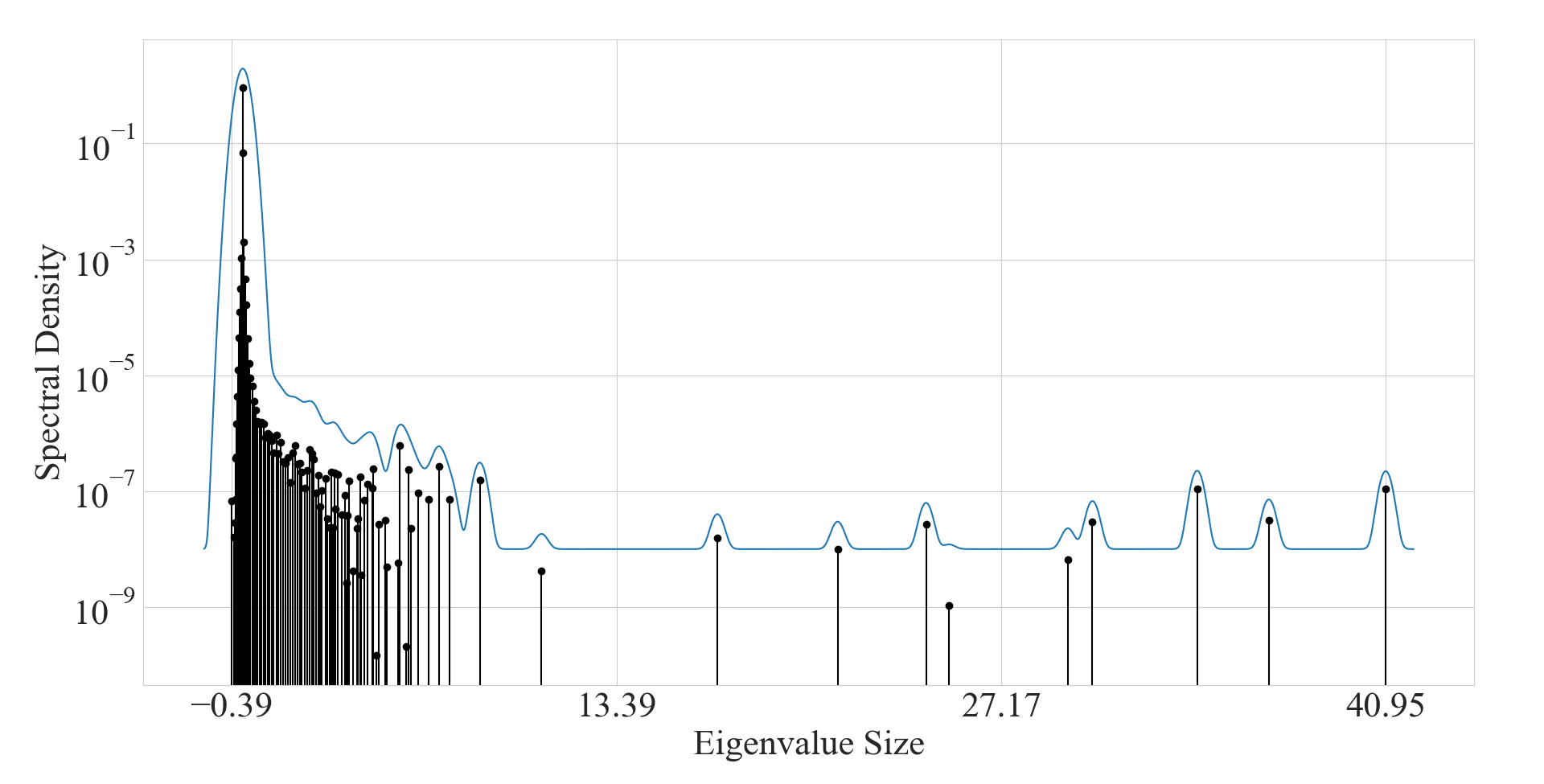}\label{fig:dc10_r18_100epscls10}}
     \hfill
     \subfloat[][Eigenvalue spectrum of MIOv3 on CIFAR10 dataset, pre-trained for 10 epochs]{\includegraphics[width=0.33\linewidth]{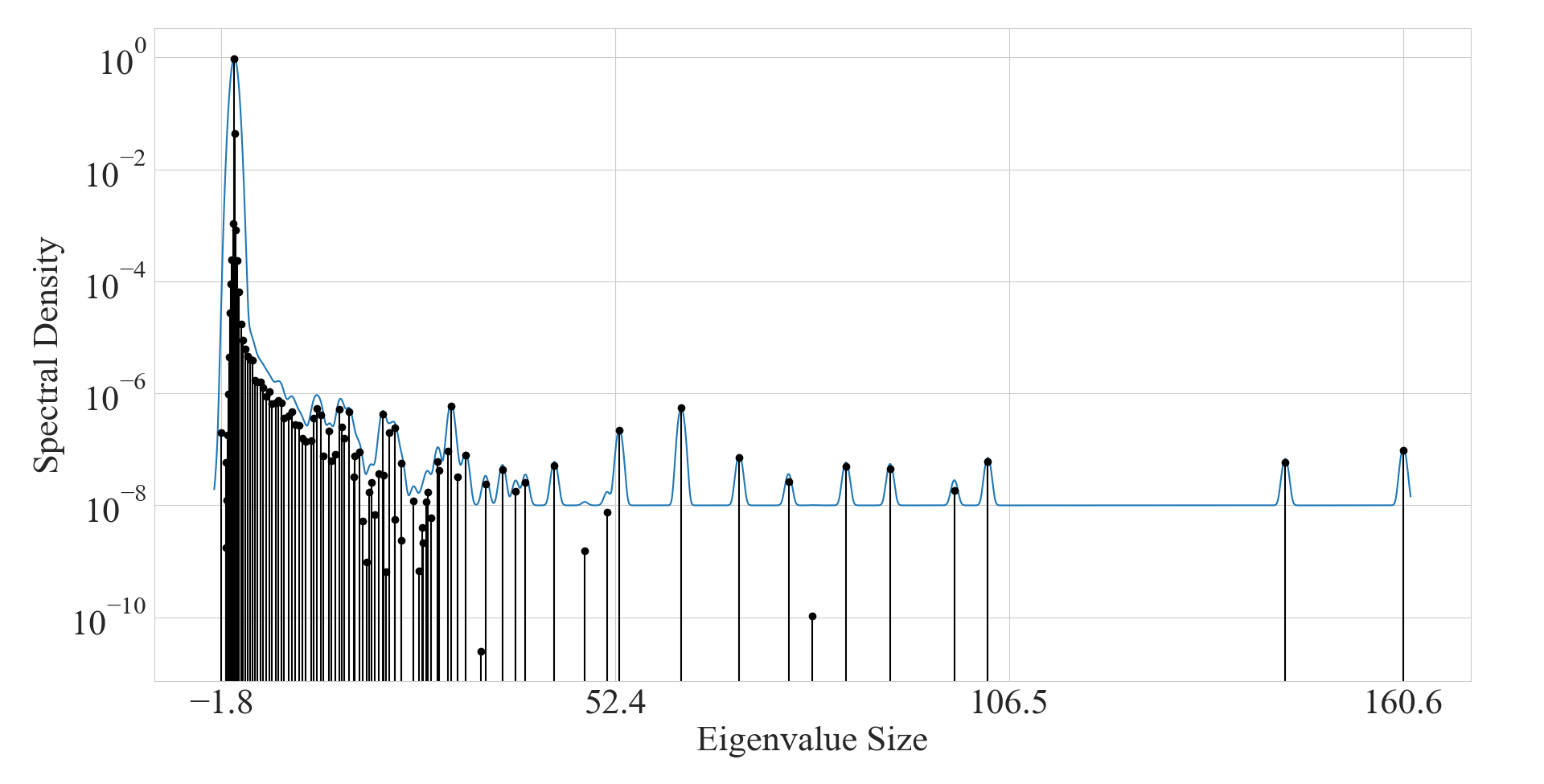}\label{fig:mc10_r18_100epscls10}}
     \qquad
     \subfloat[][Eigenvalue spectrum of SimCLR on CIFAR100 dataset, pre-trained for 10 epochs]{\includegraphics[width=0.33\linewidth]{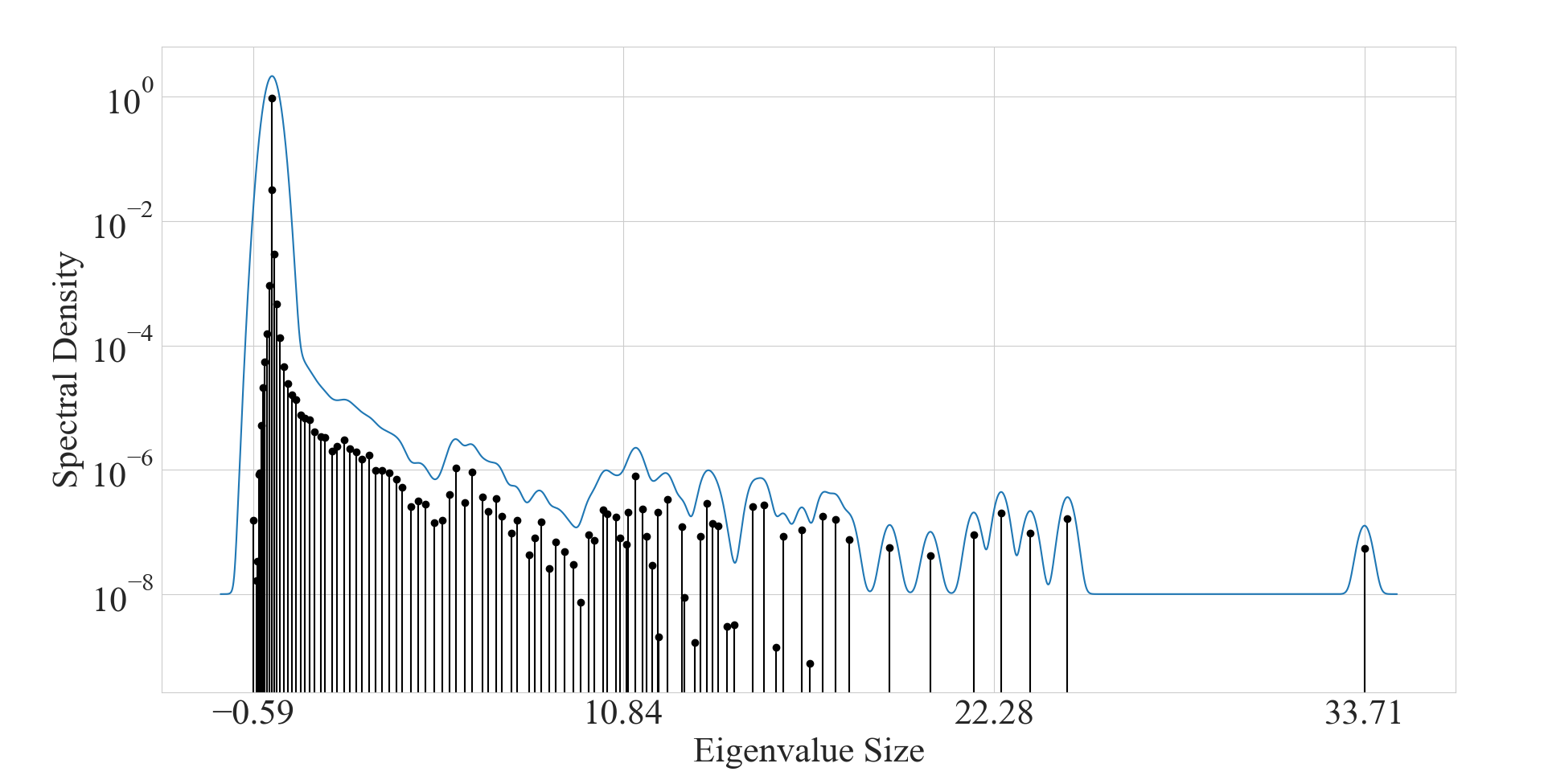}\label{fig:sc100_r18_100epscls10}}
     \hfill
     \subfloat[][Eigenvalue spectrum of DCL on CIFAR100 dataset, pre-trained for 10 epochs]{\includegraphics[width=0.33\linewidth]{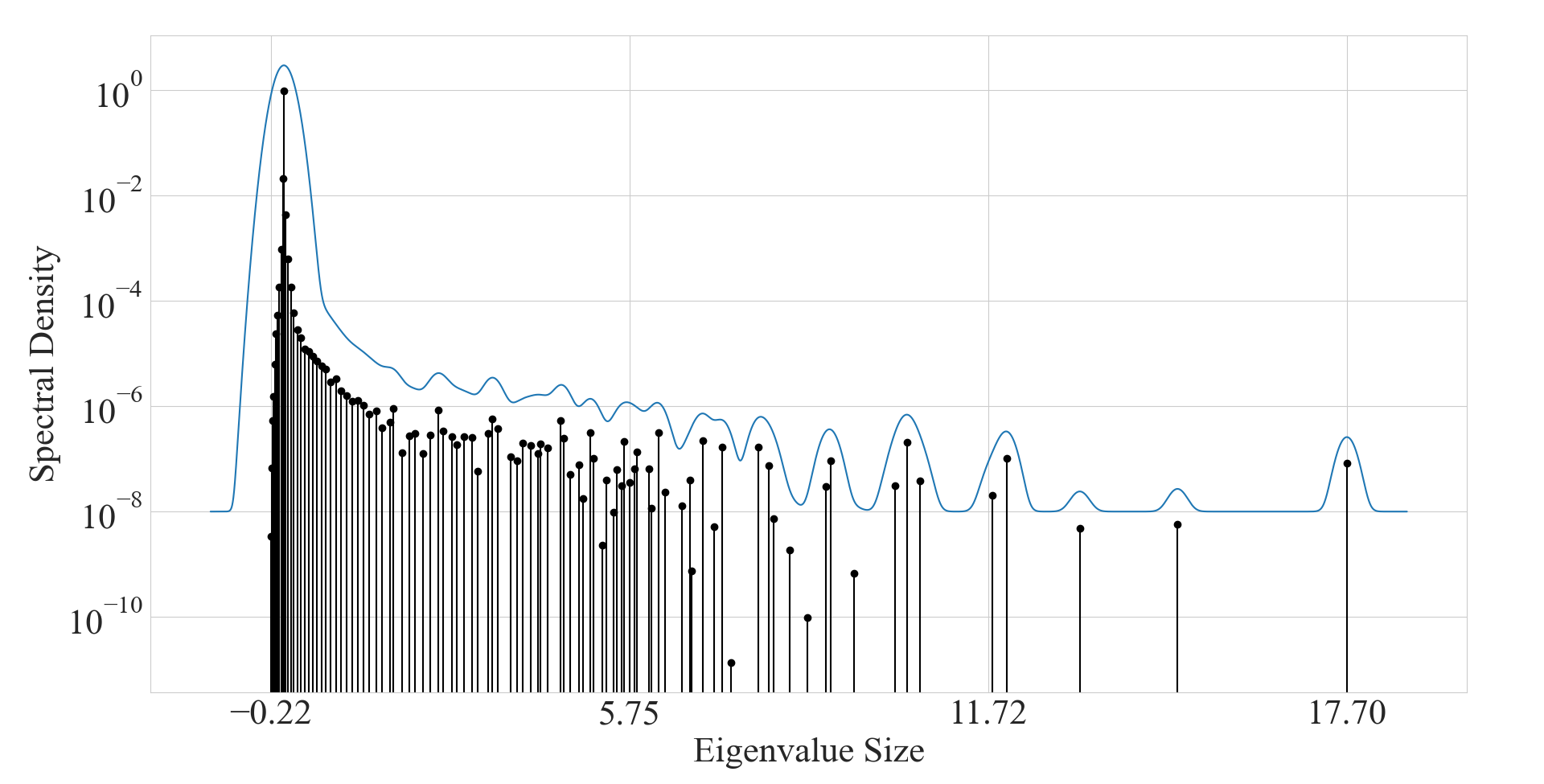}\label{fig:dc100_r18_100epscls10}}
     \hfill
     \subfloat[][Eigenvalue spectrum of MIOv3 on CIFAR100 dataset, pre-trained for 10 epochs]{\includegraphics[width=0.33\linewidth]{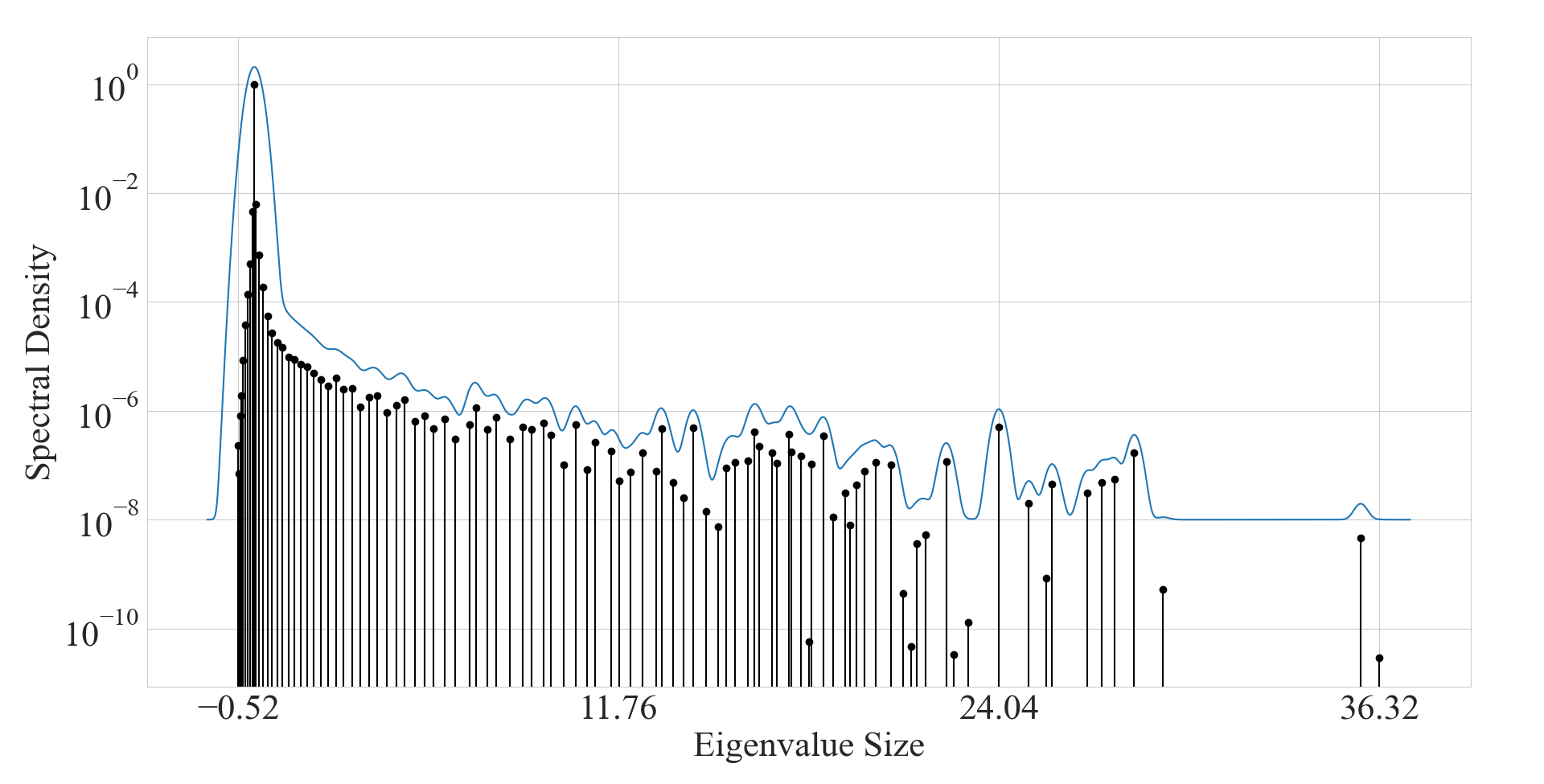}\label{fig:mc100_r18_100epscls10}}
    \caption{Plot of eigenvalues of parameters of ResNet18 encoder, obtained after 10 epochs of pre-training on CIFAR10 and CIFAR100 datasets with different SSL frameworks, namely, SimCLR, DCL and MIOv3, with respect to categorical cross-entropy loss.}
    \label{fig:alleigsnew10epscls}
\end{figure*}

\begin{figure*}[!ht]
    \centering
    \subfloat[][Eigenvalue spectrum of SimCLR on CIFAR10 dataset, pre-trained for 100 epochs]{\includegraphics[width=0.33\linewidth]{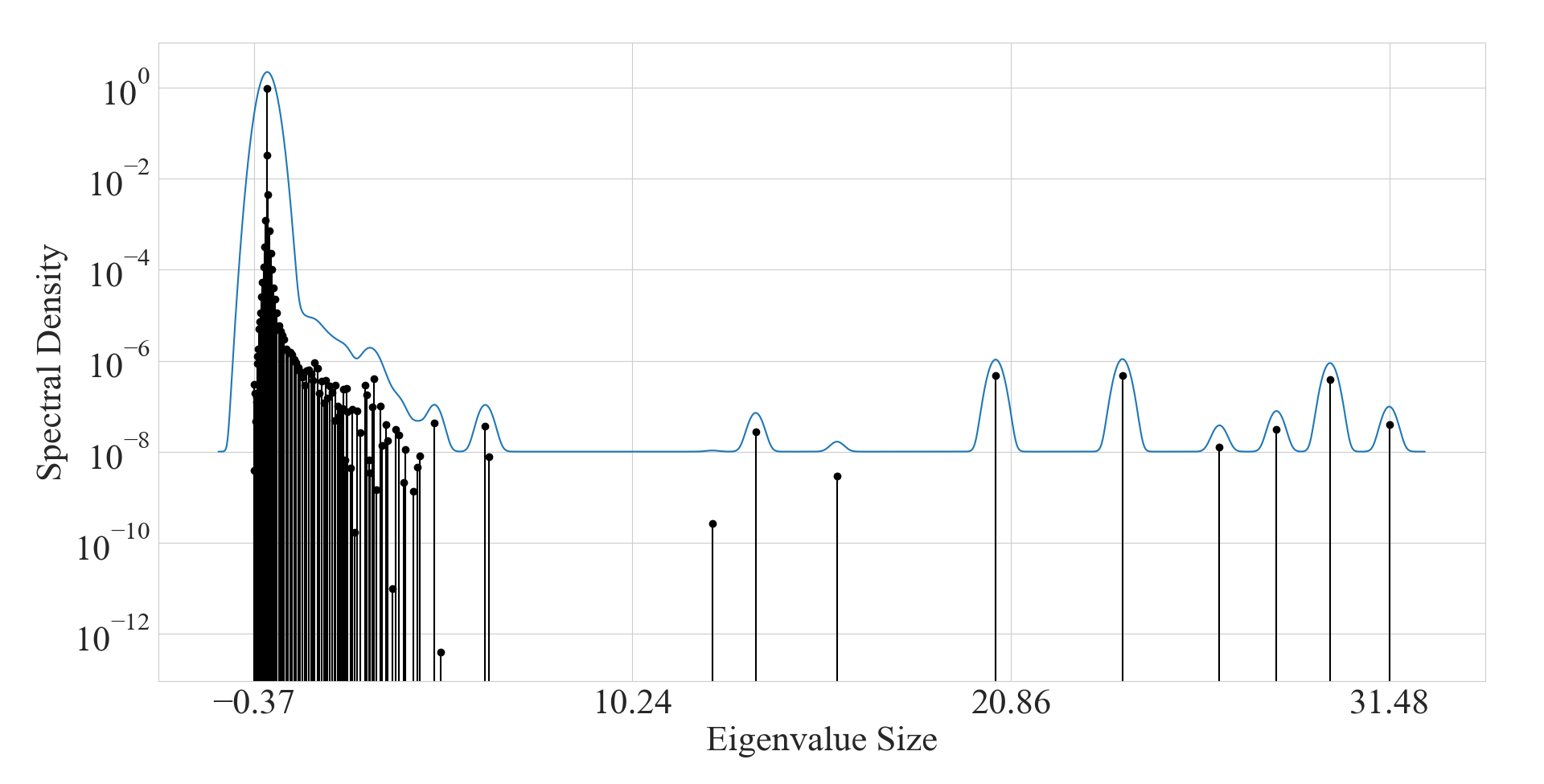}\label{fig:sc10_r18_100epscls100}}
    \hfill
     \subfloat[][Eigenvalue spectrum of DCL on CIFAR10 dataset, pre-trained for 100 epochs]{\includegraphics[width=0.33\linewidth]{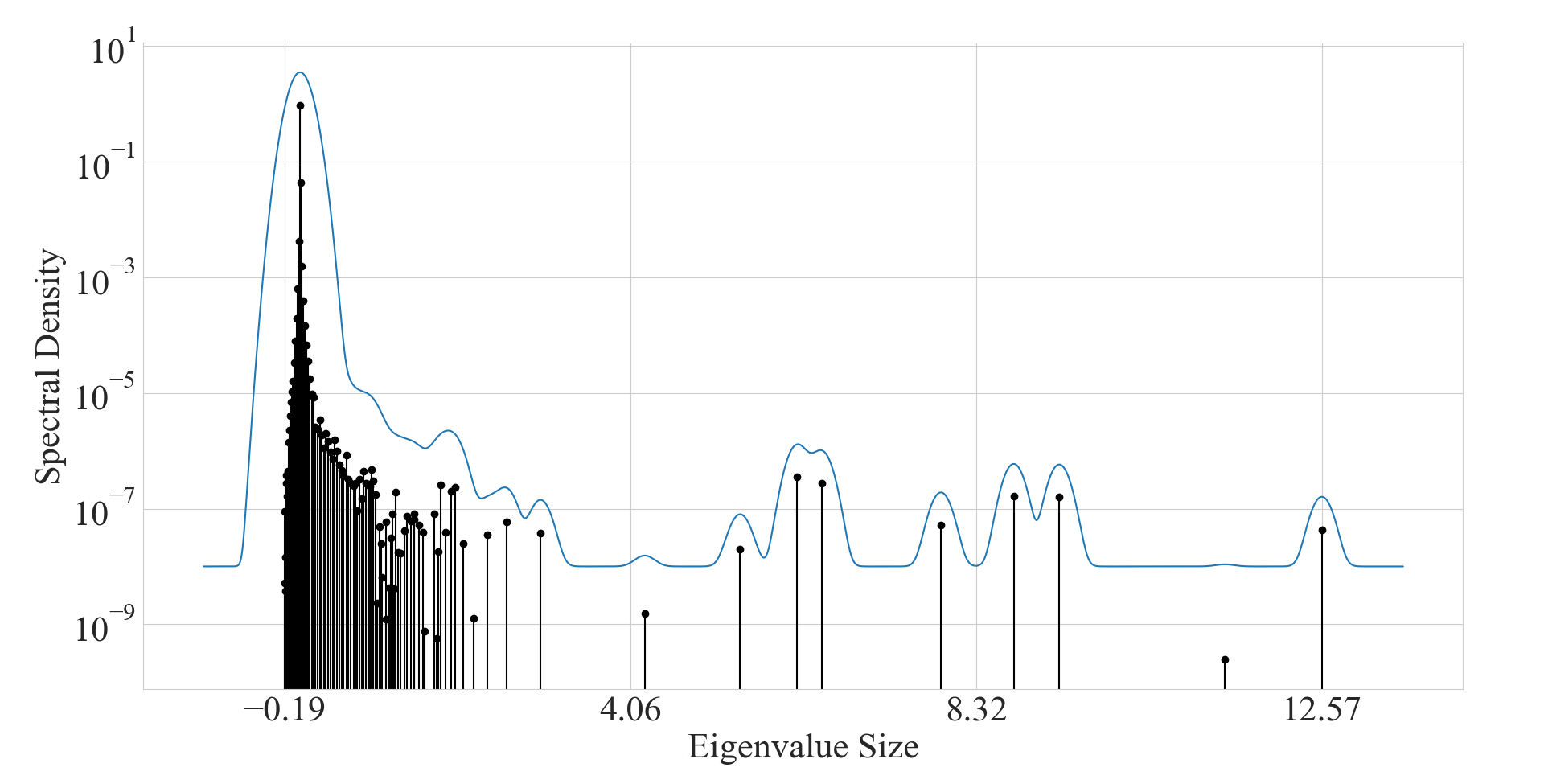}\label{fig:dc10_r18_100epscls100}}
     \hfill
     \subfloat[][Eigenvalue spectrum of MIOv3 on CIFAR10 dataset, pre-trained for 100 epochs]{\includegraphics[width=0.33\linewidth]{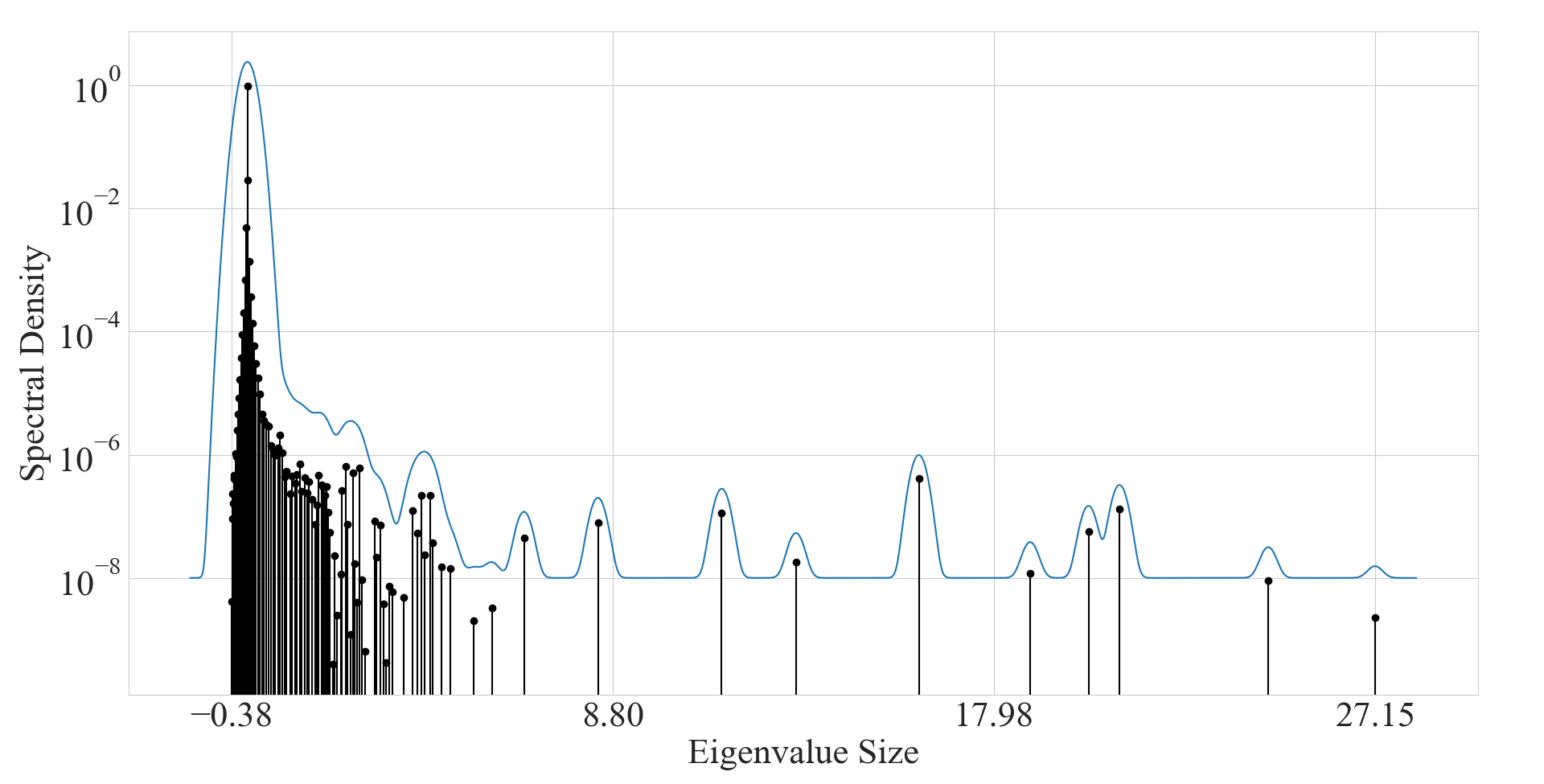}\label{fig:mc10_r18_100epscls100}}
     \qquad
     \subfloat[][Eigenvalue spectrum of SimCLR on CIFAR100 dataset, pre-trained for 100 epochs]{\includegraphics[width=0.33\linewidth]{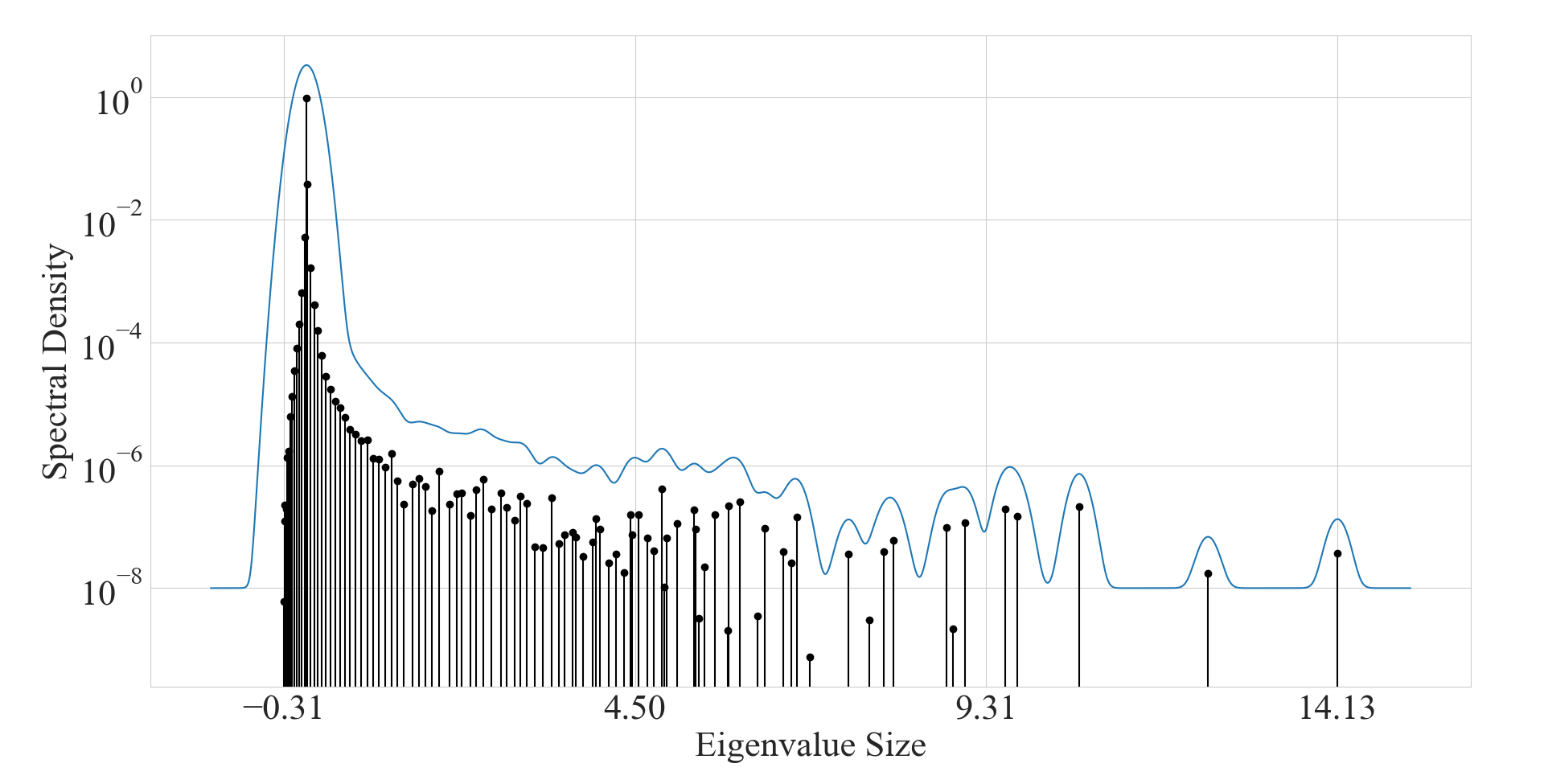}\label{fig:sc100_r18_100epscls100}}
     \hfill
     \subfloat[][Eigenvalue spectrum of DCL on CIFAR100 dataset, pre-trained for 100 epochs]{\includegraphics[width=0.33\linewidth]{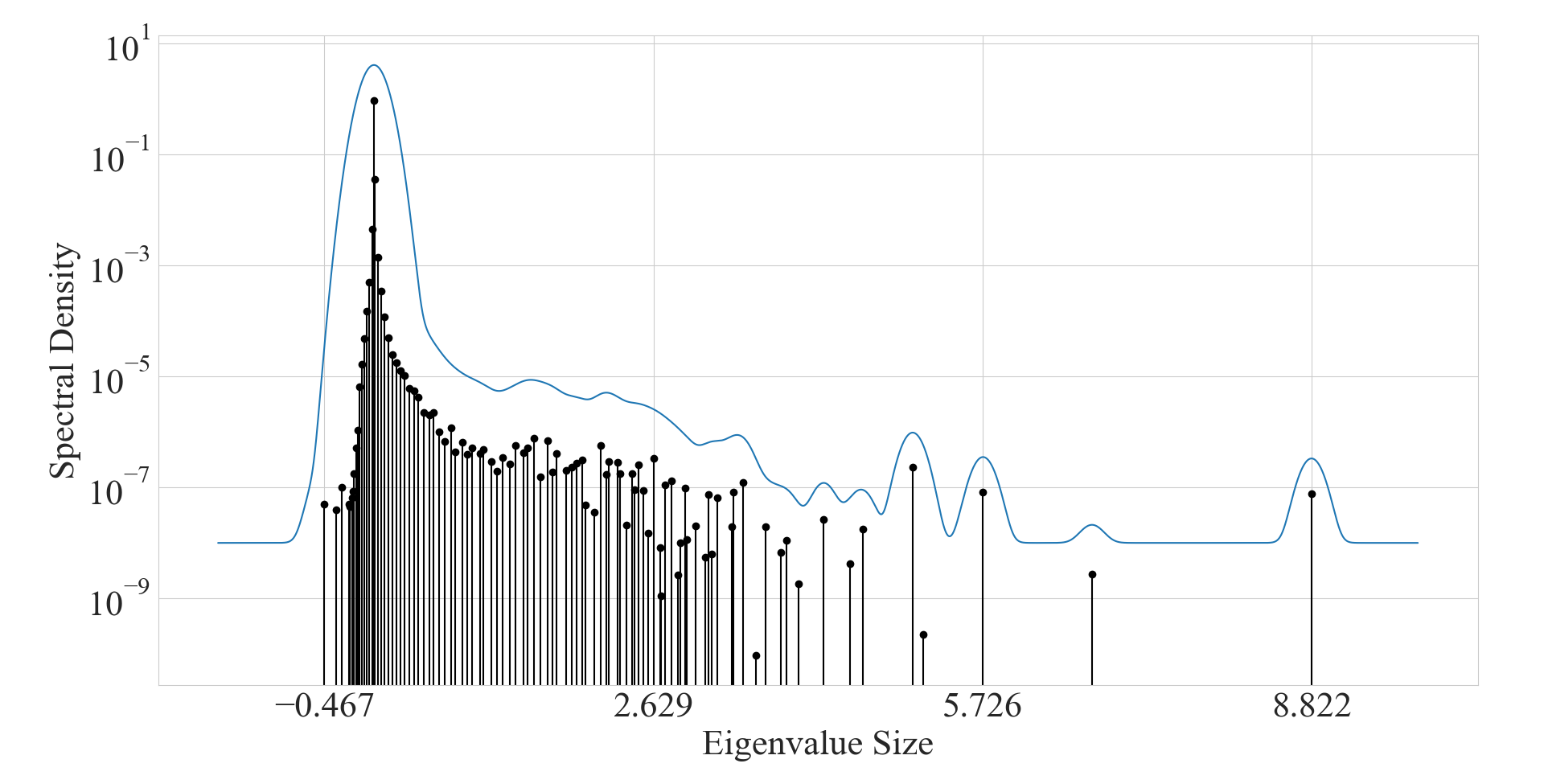}\label{fig:dc100_r18_100epscls100}}
     \hfill
     \subfloat[][Eigenvalue spectrum of MIOv3 on CIFAR100 dataset, pre-trained for 100 epochs]{\includegraphics[width=0.33\linewidth]{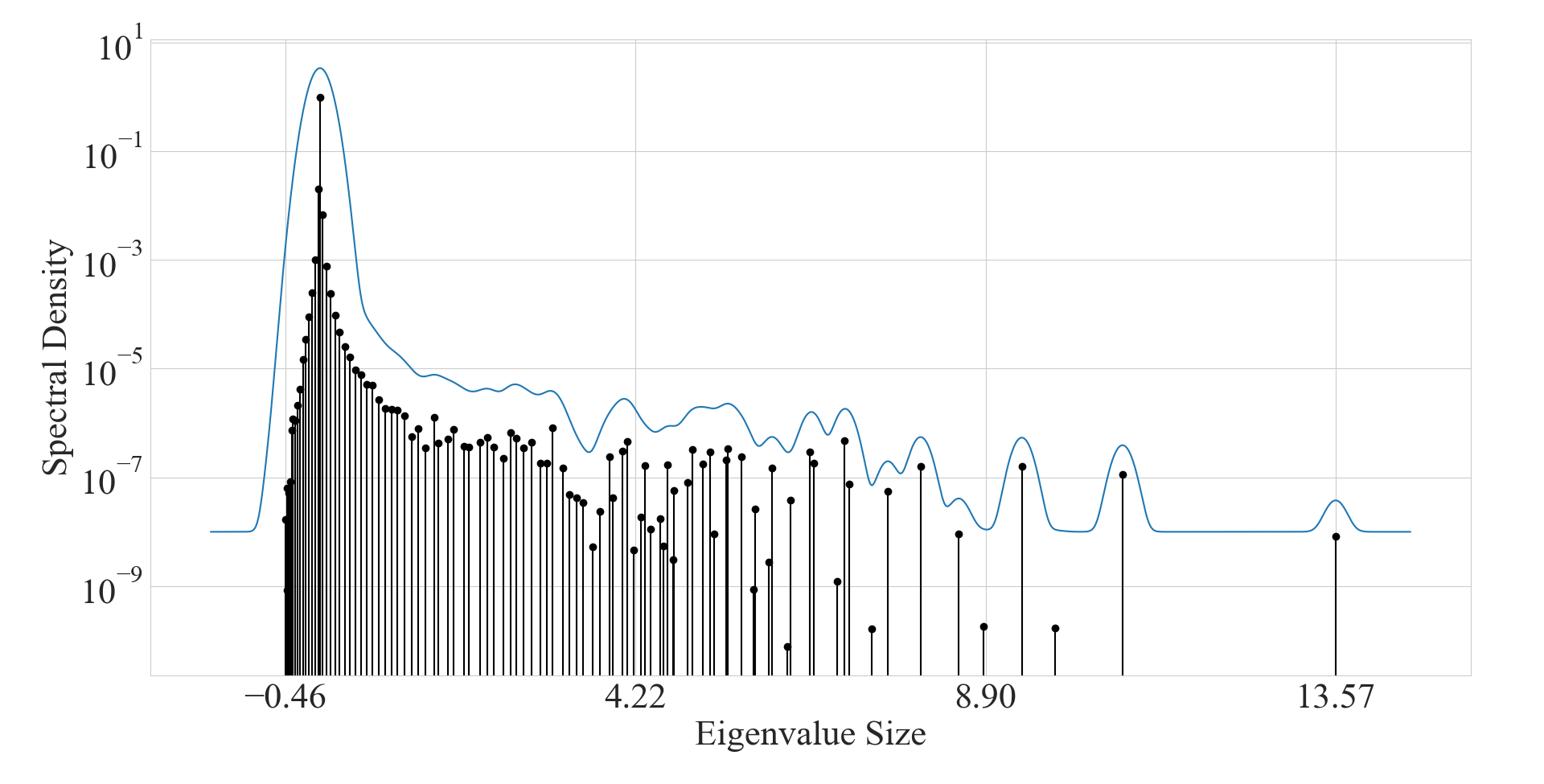}\label{fig:mc100_r18_100epscls100}}
    \caption{Plot of eigenvalues of parameters of ResNet18, obtained after 100 epochs of pre-training on CIFAR10 and CIFAR100 datasets with different SSL frameworks, namely, SimCLR, DCL and MIOv3, with respect to categorical cross-entropy loss.}
    \label{fig:alleigsnew100epscls}
\end{figure*}

\begin{figure*}[!ht]
    \centering
    \subfloat[][Eigenvalue spectrum of SimCLR on CIFAR10 dataset, pre-trained for 200 epochs]{\includegraphics[width=0.33\linewidth]{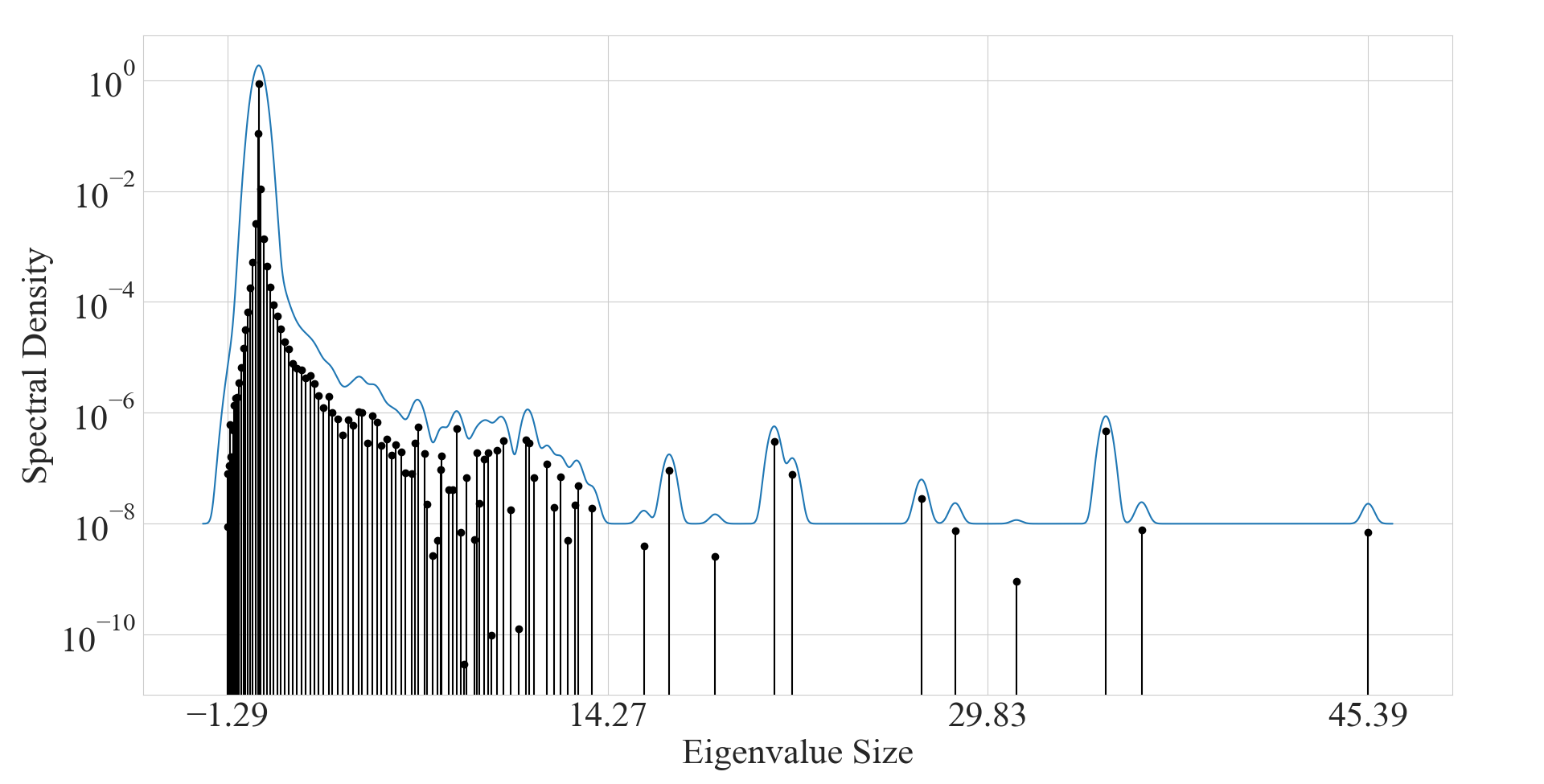}\label{fig:sc10_r18_100epscls200}}
    \hfill
     \subfloat[][Eigenvalue spectrum of DCL on CIFAR10 dataset, pre-trained for 200 epochs]{\includegraphics[width=0.33\linewidth]{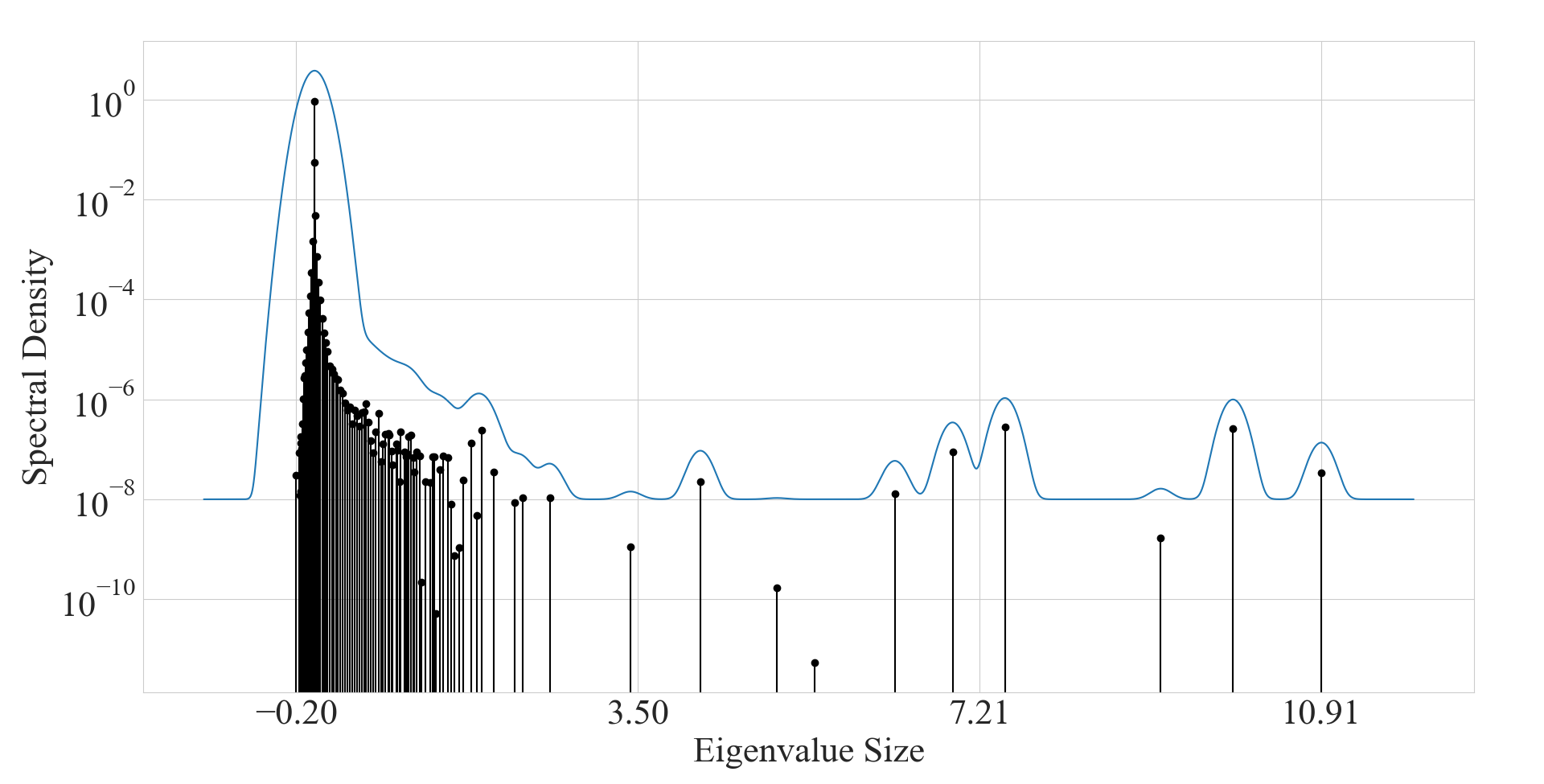}\label{fig:dc10_r18_100epscls200}}
     \hfill
     \subfloat[][Eigenvalue spectrum of MIOv3 on CIFAR10 dataset, pre-trained for 200 epochs]{\includegraphics[width=0.33\linewidth]{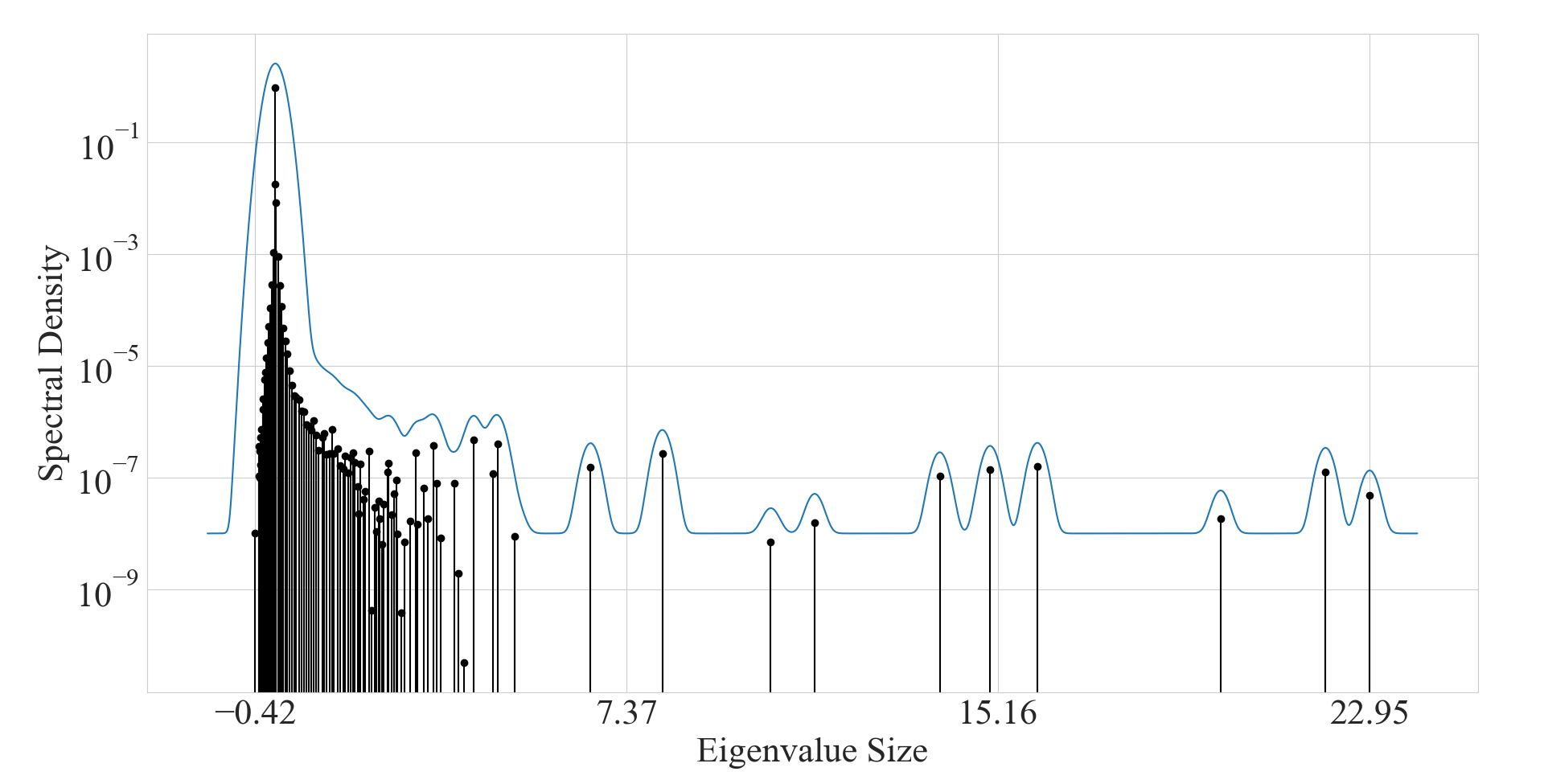}\label{fig:mc10_r18_100epscls200}}
     \qquad
     \subfloat[][Eigenvalue spectrum of SimCLR on CIFAR100 dataset, pre-trained for 200 epochs]{\includegraphics[width=0.33\linewidth]{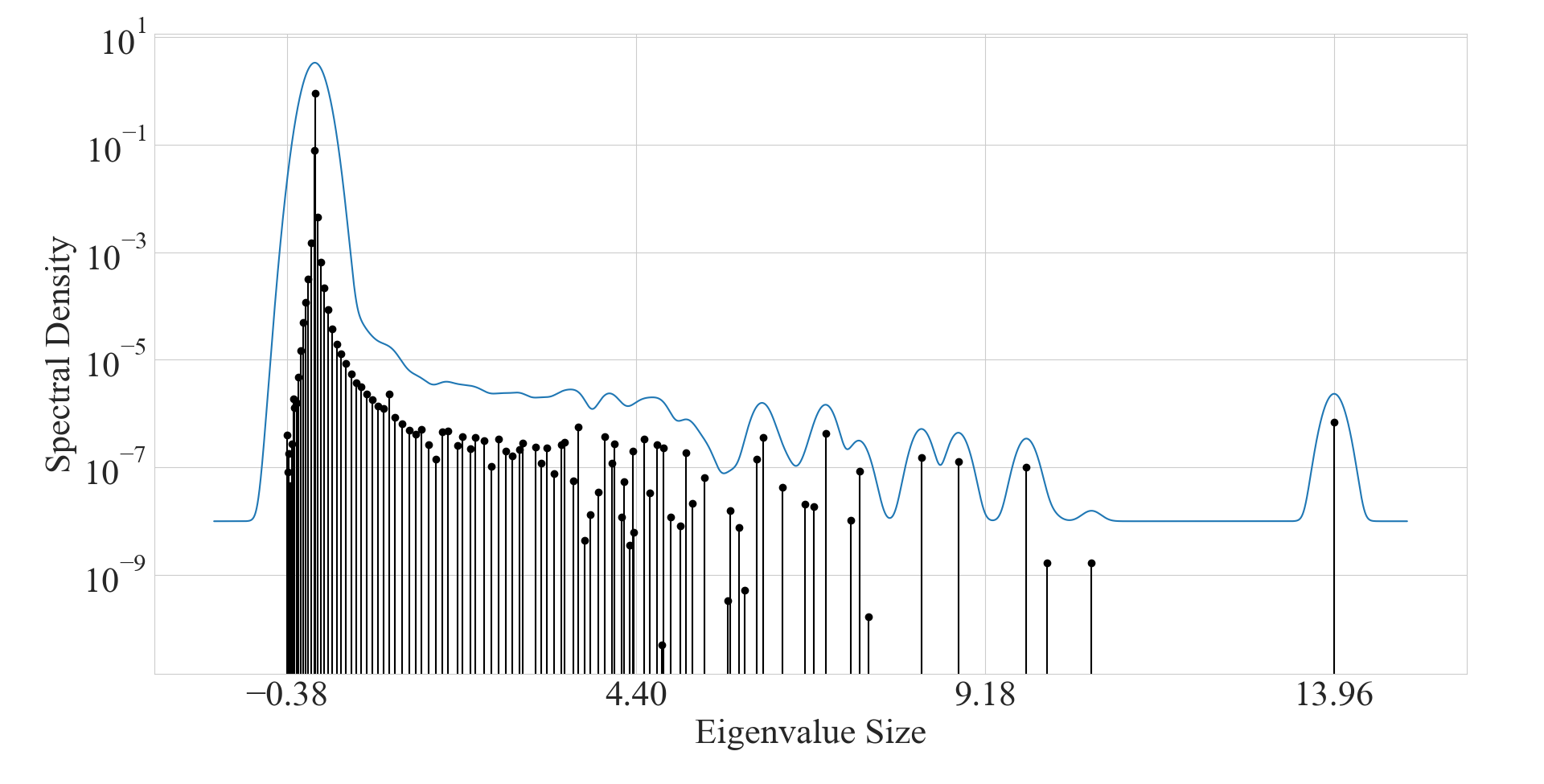}\label{fig:sc100_r18_100epscls200}}
     \hfill
     \subfloat[][Eigenvalue spectrum of DCL on CIFAR100 dataset, pre-trained for 200 epochs]{\includegraphics[width=0.33\linewidth]{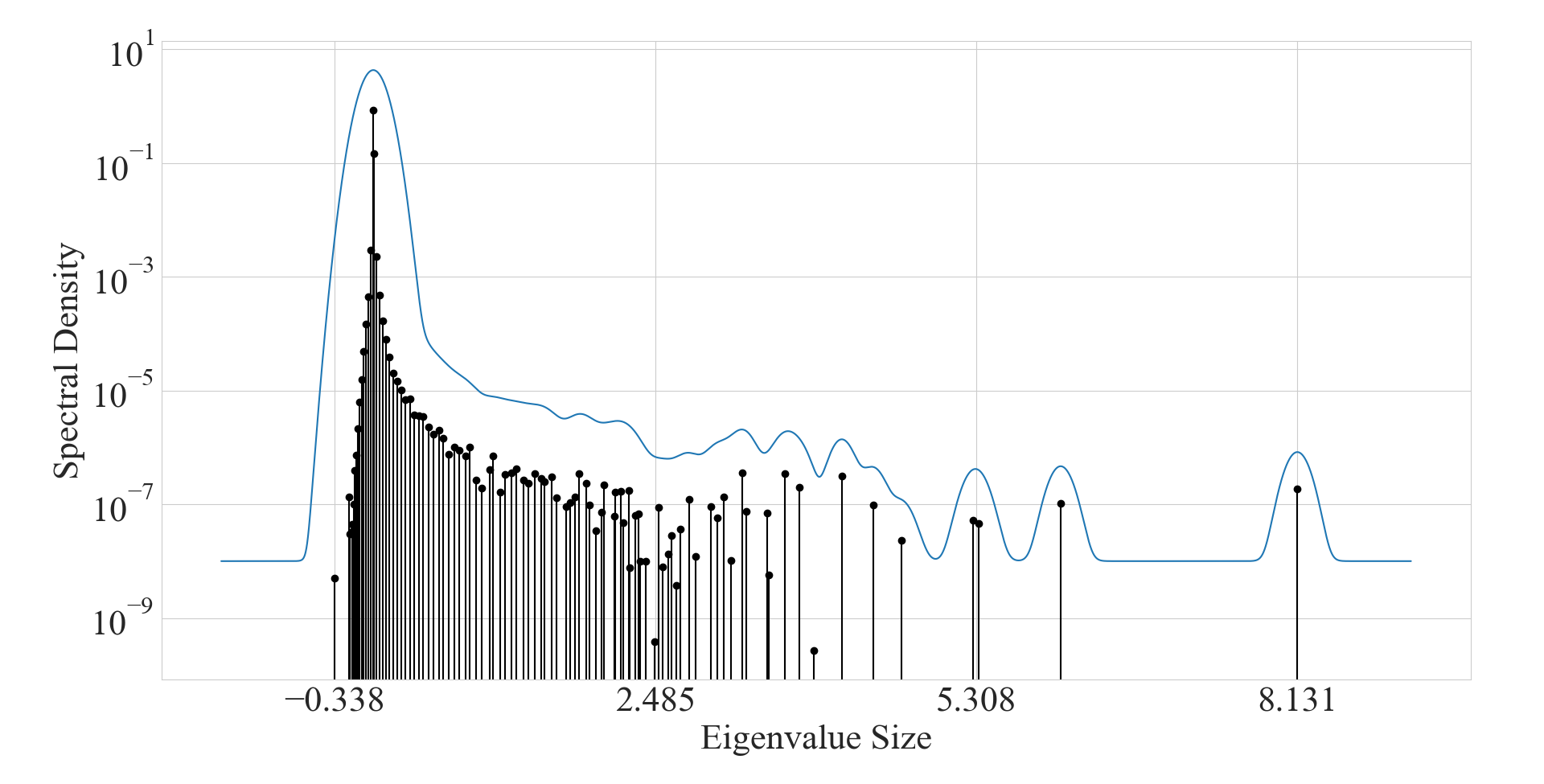}\label{fig:dc100_r18_100epscls200}}
     \hfill
     \subfloat[][Eigenvalue spectrum of MIOv3 on CIFAR100 dataset, pre-trained for 200 epochs]{\includegraphics[width=0.33\linewidth]{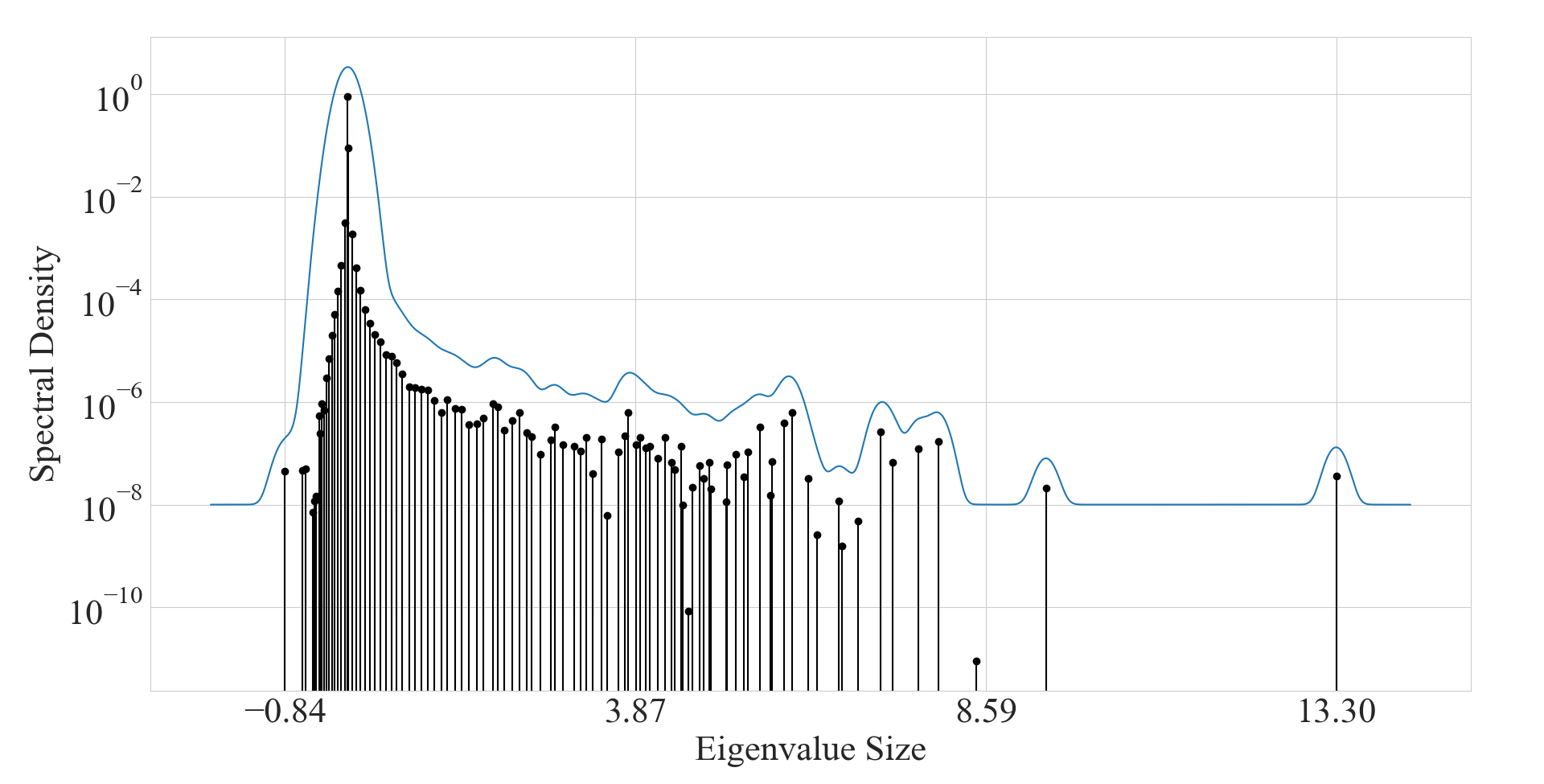}\label{fig:mc100_r18_100epscls200}}
    \caption{Plot of eigenvalues of parameters of ResNet18, obtained after 200 epochs of pre-training on CIFAR10 and CIFAR100 datasets with different SSL frameworks, namely, SimCLR, DCL and MIOv3, with respect to categorical cross-entropy loss.}
    \label{fig:alleigsnew200epscls}
\end{figure*}

We also use the pre-trained weights in a linear probing (classification) task, and after fine-tuning the linear classifier to the best possible performance, we computed the 100 eigenvalues of the encoder parameters only with respect to the categorical cross-entropy loss using the Lanczos iteration method. We found the same behaviour as the contrastive losses, that is, the eigenspectrum contains negative eigenvalues, hinting at premature convergence to a saddle point. The eigenspectrum of the encoders pretrained with SimCLR, DCL and MIOv3 are presented in Fig. \ref{fig:alleigsnew10epscls}, \ref{fig:alleigsnew100epscls}, and \ref{fig:alleigsnew200epscls} for 10th, 100th and 200th epoch, respectively.

\section{Comparison of Binary Contrastive Loss with Other Losses}

\subsection{Comparison by performance in downstream task}
\label{sec:perfdown}

In Self-supervised learning (SSL), each sample is considered to be a class in its own right. The objective of self-supervised learning is to learn representation from the data, such that the feature vectors of semantically different classes in the latent feature space are linearly separable. In other words, the base encoder $f$ is trained to learn separating hyperplanes between $N$ convex sets in the latent feature space, where $N$ is the inherent number of classes. Ideally, samples from each class are mapped to a closed convex set in the latent feature space.

The linear separability of convex sets in the latent feature space dictates the accuracy in linear probing or kNN classification tasks, as generally done to evaluate the quality of representations learned by SSL frameworks. In other words, kNN classification accuracy or linear probing classification determines, how well the class-specific convex sets in the latent feature space can be separated by hyperplanes. That being said, the final parameter state $\mathcal{P}_{fA}^T$ of the base encoder of framework $A$, can be compared to the final parameter state $\mathcal{P}_{fB}^T$ of the base encoder of framework $B$, by investigating the linear separability of the mapped feature vectors in the latent space. 

Hence, in this work, we use 200-NN classification accuracy as a metric to measure the linear separability of the features in the latent space of the base encoder.

\subsection{Comparative Analysis of Loss Functions}
\label{sec:comp_anal_loss_func}

In this section, we identify and segregate the components of different loss functions used in self-supervised learning into two categories: Attracting and Repulsing terms. The attracting term refers to the term which on being optimized, brings the two samples in a positive pair closer. On the other hand, the repulsing term refers to the term, on being optimized pushes the samples in a negative pair away from each other. The comparative analysis is presented in Table \ref{tab:comp_anal_label}. The description of each term used in the table is given in Table \ref{tab:losscomptab}.

\begin{table*}[!ht]
\caption{Description of notations used in the different loss functions given in Table \ref{tab:comp_anal_label}.}
    \label{tab:losscomptab}
    \centering
    \begin{tabular}{c|p{9cm}}
    \hline
         Frameworks & Description of Notations in Loss Functions\\ \hline \hline
         \textbf{SimCLR} & Notations Used : $x_+$, $(x_+,x_k)$, $(x_+,x_i)$, $C_{+,k}$/$C_{+,i}$\\ \hline \hline
         $x_+$ & Anchor Sample \\ \hline
         $(x_+,x_k)$ & Positive Pair\\\hline
         $(x_+,x_i)$ & Negative Pair\\\hline
         $C_{+,k}$/$C_{+,i}$ & Cosine Similarity of sampels in Positive / Negative Pair\\\hline \hline
         \textbf{BYOL} & Notations Used : $(x_i,x_j)$, $\theta$, $\xi$, $z^{\theta}_i$, $q_{\theta}(z^{\theta}_i)$, $z^{\xi}_j$\\ \hline \hline
         $(x_i,x_j)$ & Positive Pair\\\hline
         $\theta$ & Parameters of Online Encoder\\\hline
         $\xi$ & Parameters of Target Encoder\\\hline
         $z^{\theta}_i$ & Output features obtained from the projector of the online encoder for the sample $x_i$\\\hline
         $z^{\xi}_j$ & Output features obtained from the projector of the target encoder for the sample $x_j$\\\hline
         $q_{\theta}$ & Projector of the Online Encoder\\ \hline
         $q_{\theta}(z^{\theta}_i)$ & Output features obtained from the predictor of the online encoder for the sample $x_i$\\\hline\hline
         \textbf{MIOv1}/\textbf{MIOv2}/\textbf{MIOv3} & Notations Used : $(x_i,x_{i+N})$,$(x_k,x_l)$,$C_{i,j}$\\\hline\hline
         $(x_i,x_{i+N})$ & Positive Pair\\\hline
         $(x_k,x_l)$& Negative Pair\\\hline
         $C_{i,j}$& Cosine Similarity between the samples $x_i$ and $x_j$\\\hline \hline
         \multicolumn{2}{c}{\textbf{Other Notations used in the main manuscript}}\\ \hline \hline
        $\mathcal{L}_{v3}\circ g_{\psi} \circ f_{\theta}$ & Non-convex function, consisted of encoder, projector, and the loss function MIOv3 \\ \hline
    \end{tabular}
\end{table*}

\begin{sidewaystable}[htbp]
    \centering
    \caption{Comparison of components of loss functions used in different self-supervised algorithms}
    \begin{tabular}{c|c|c|c}
        \hline
        Method & Loss Function & Attracting Term & Repulsing Term \\ \hline \hline
        \multirow{4}{*}{SimCLR} & \multirow{8}{*}{$- \mathop{\mathbb{E}}_{\substack{(x^+, x_k) \in \mathcal{X}_{+} \\ (x^+, x_i) \in \mathcal{X}_{-}}} \left [ ln \frac{e^{\frac{C_{+,k}}{\tau}}}{e^{\frac{C_{+,k}}{\tau}} + \sum_{\substack{i = 1 \\ i \neq k }}^{N} e^{\frac{C_{+,i}}{\tau}}} \right]$} & \multirow{8}{*}{$- \mathop{\mathbb{E}}_{\substack{(x^+, x_k) \in \mathcal{X}_{+}}}\frac{C_{+,k}}{\tau}$} & \multirow{8}{*}{$\mathop{\mathbb{E}}_{(x^+, x_i) \in \mathcal{X}_{-}} ln \left (\sum_{\substack{i = 1}}^{N} e^{\frac{C_{+,i}}{\tau}} \right )$} \\ 
        &&&\\
        &&&\\
        &&&\\ \cline{1-1}
        \multirow{4}{*}{MoCoV2}&&&  \\ 
        &&&\\
        &&&\\
        &&&\\ \hline
        \multirow{4}{*}{BYOL} & \multirow{4}{*}{$- 2  \mathop{\mathbb{E}}_{\substack{(x_i, x_j) \in \mathcal{X}_{+}}} \left [ \frac{<q_{\theta}(z_i^{\theta}),z_j^{\xi}>}{\lVert q_{\theta}(z_i^{\theta}) \rVert_2 \lVert z_j^{\xi} \rVert_2} \right ]$}  & \multirow{4}{*}{$- 2 \mathop{\mathbb{E}}_{\substack{(x_i, x_j) \in \mathcal{X}_{+}}} \left [ \frac{<q_{\theta}(z_i^{\theta}),z_j^{\xi}>}{\lVert q_{\theta}(z_i^{\theta}) \rVert_2 \lVert z_j^{\xi} \rVert_2} \right ]$} & \multirow{4}{*}{-} \\ 
        &&&\\
        &&&\\
        &&&\\ \hline 
        \multirow{6}{*}{MIOv1} & \multirow{3}{*}{$-\mathop{\mathbb{E}}_{(x_i,x_j) \in \mathcal{X}_{+}}\left[ln \left ( \frac{1}{1+e^{-\frac{C_{i,j}}{\tau}}} \right ) \right]$} & \multirow{6}{*}{$-\mathop{\mathbb{E}}_{(x_i,x_j) \in \mathcal{X}_{+}}\left[ln \left ( \frac{1}{1+e^{-\frac{C_{i,j}}{\tau}}} \right ) \right]$} & \multirow{6}{*}{$-\mathop{\mathbb{E}}_{(x_k,x_l) \in \mathcal{X}_{-}}\left[ln \left(1 - \frac{1}{1+e^{-\frac{C_{k,l}}{\tau}}} \right ) \right]$}\\
        &&&\\
        &&&\\
        &\multirow{3}{*}{$-\mathop{\mathbb{E}}_{(x_k,x_l) \in \mathcal{X}_{-}}\left[ln \left(1 - \frac{1}{1+e^{-\frac{C_{k,l}}{\tau}}} \right ) \right]$}&&\\
        &&&\\
        &&&\\ \hline 
        \multirow{3}{*}{MIOv2} & \multirow{3}{*}{$-\bbE_{(x_i,x_{j})\in \chi^+}\frac{C_{i,j}}{\tau} -\bbE_{(x_k,x_{l})\in \chi^-}{ln \left(1 - \frac{1}{1+e^{-\frac{C_{k,l}}{\tau}}}\right)}$} & \multirow{3}{*}{$-\bbE_{(x_i,x_{j})\in \chi^+}\frac{C_{i,k}}{\tau}$} & \multirow{3}{*}{$-\bbE_{(x_k,x_{l})\in \chi^-}{ln \left(1 - \frac{1}{1+e^{-\frac{C_{k,l}}{\tau}}}\right)}$}\\
        &&&\\
        &&&\\\hline 
        \multirow{3}{*}{MIOv3} & \multirow{3}{*}{$-\bbE_{(x_i,x_{j})\in \mathcal{X}_+}\frac{C_{i,j}}{\tau} -\bbE_{(x_k,x_{l})\in \mathcal{X}_-}e^{\frac{C_{k,l}}{\tau}} $} & \multirow{3}{*}{$-\bbE_{(x_i,x_{j})\in \mathcal{X}_+}\frac{C_{i,j}}{\tau}$} & \multirow{3}{*}{$-\bbE_{(x_k,x_{l})\in \chi^-}e^{\frac{C_{k,l}}{\tau}}$}\\
        &&&\\
        &&&\\\hline 
    \end{tabular}
    \label{tab:comp_anal_label}
\end{sidewaystable}

\section{Contrastive Loss}

\subsection{InfoNCE Loss Function}
\label{subsec:contr}

 In the InfoNCE-based contrastive learning framework, each sample is treated as a separate class. The learning principle generally involves maximizing the similarity between two augmented versions of a sample comprising a positive pair and minimizing the similarity between samples in a negative pair. Feature vectors of samples in a positive pair are mapped close to each other in the feature space, whereas feature vectors of samples are pushed away in the case of a negative pair. This allows the encoder to learn transformation and context invariant representations such that the feature vectors obtained from samples of different classes are easily separable in the feature space.

\indent
The InfoNCE \cite{cpc} loss function is the negative of the expected logarithm of the probability of correctly predicting the positive pair. The InfoNCE \cite{cpc} loss function ($\mathcal{L}_{C}$) is generally used in the form given below

\begin{equation}
\label{eqn:1}
    \mathcal{L}_{C} = - \mathop{\mathbb{E}}_{\substack{(x^+, x_k) \in \mathcal{X}_{+} \\ (x^+, x_i) \in \mathcal{X}_{-}}} \left [ ln \frac{e^{\frac{C_{+,k}}{\tau}}}{e^{\frac{C_{+,k}}{\tau}} + \sum_{\substack{i = 1 \\ i \neq k }}^{N} e^{\frac{C_{+,i}}{\tau}}} \right]
\end{equation}

where $C_{i,j}$ is the cosine similarity between two feature vectors denoted by $z$ obtained by passing $x$ through the encoder and the projector. Furthermore, $\mathcal{X}_{+}$ and $\mathcal{X}_{-}$ are the sets of all positive pairs and negative pairs, respectively, on $\mathbb{R}^n \times \mathbb{R}^n$. Also, $\tau$ is the temperature parameter. $(x^+,x_k)$ and $(x^+, x_i)$ are samples obtained from $\mathcal{X}_{+}$ and $\mathcal{X}_{-}$, respectively, where $x^+$ is the anchor sample.

\newpage

\subsection{How we thought of MIOv1 Loss?}

The primary motivation of our proposed framework is to classify the type of pairs in self-supervised contrastive learning (SSCL) setting. In SSCL, we generally construct two type of pairs, positive and negative. Let us denote, the set of positive and negative pairs as $\mathcal{X}_+$ and $\mathcal{X}_-$, respectively. A pair $(z_i, z_j)$ is assigned a binary class label $k_{ij}$: $k_{ij} = 1$ if $(z_i, z_j) \in \mathcal{X}_+$ and $k_{ij} = 0$ if $(z_i, z_j) \in \mathcal{X}_-$.

The objective of our SSCL framework is to calculate the posterior probabilities of the classes, given the pair of samples. In self-supervised learning, the distribution of the data, as well as the distribution of the samples are also unknown. While this objective is similar to the Noise Constrastive Estimation (NCE) \cite{nce2012}, the formulation of our first loss MIOv1 differs in some aspects. We will discuss the same in the following paragraphs. First, we will discuss the basic notions of NCE, and then discuss the reasons behind the deviation in MIOv1.

\noindent
\textbf{The general notion of Noise Contrastive Estimation}

Let us define the class-conditional probabilities be as follows,

\begin{equation*}
\begin{split}
    p((z_i,z_j)|k=1) & = p_m((z_i,z_j);\theta) \\
    p((z_i,z_j)|k=0) & = p_n((z_i,z_j))
\end{split}
\end{equation*}

When the samples in a batch with size $N$ are paired as shown in Sec. \ref{sec:suppsec1}, then we get $4N^2 - 4N$ negative pairs, and $2N$ positive pairs. We discard the $2N$ number of self-pairs. Therefore, the prior probabilities should be as follows,

\begin{equation*}
\begin{split}
    P(k=1) &= \frac{2N}{4N^2-2N} = \frac{1}{2N-1}\\
    P(k=0) &= \frac{4N^2 - 4N}{4N^2-2N}=\frac{2N-2}{2N-1}
\end{split}
\end{equation*}

Following \cite{nce2012}, we observe that $\nu = 2N-2$. Hence, the posterior probabilities for the classes can be defined as follows,

\begin{equation}
    \begin{split}
        P(k=1|(z_i,z_j);\theta) &= \frac{p_m((z_i,z_j);\theta)}{p_m((z_i,z_j);\theta) + \nu p_n((z_i,z_j))}\\
        P(k=0|(z_i,z_j);\theta) &= \frac{\nu p_n((z_i,z_j))}{p_m((z_i,z_j);\theta) + \nu p_n((z_i,z_j))}\\
    \end{split}
\end{equation}

Denoting $P(k=1|(z_i,z_j);\theta)$ by $h((z_i,z_j);\theta)$, the loss function can be defined as follows

\begin{equation}
\label{eqn:nceL}
    \begin{split}
        \mathcal{L} = &- \frac{1}{T_p + T_n}\sum_{(z_i, z_j) \in \mathcal{X}_+ \bigcup \mathcal{X}_-} \left [ k_{ij} \ln P(k_{ij} = 1|(z_i,z_j);\theta)  + (1-k_{ij}) \ln P(k_{ij} = 0|(z_i,z_j);\theta) \right ]\\
        = &-\frac{1}{T_p + T_n} \left [\sum_{(z_i, z_j) \in \mathcal{X}_+}\ln h((z_i, z_j); \theta) .-\sum_{(z_k, z_l) \in \mathcal{X}_-}\ln (1 - h((z_k, z_l);\theta) \right]\\
    \end{split}
\end{equation}

\noindent
where $h((z_i,z_j);\theta)$ can be written as

\begin{equation}
\label{eqn:nceh}
    h((z_i,z_j);\theta) = \frac{1}{1 + \nu \exp(-\mathcal{G}((z_i,z_j);\theta)}
\end{equation}

\noindent
where $\mathcal{G}((z_i,z_j);\theta) = \ln p_m((z_i,z_j);\theta) - \ln p_n((z_i,z_j))$.\\

\noindent
\textbf{Why does the formulation of MIOv1 differ even when it is doing binary classification of the pairs?} \\

As previously mentioned, the primary objective of MIOv1 is to classify the type of pairs, positive or negative in self-supervised contrastive learning. This provides us with a binary logistic regression problem which requires estimating the maximum likelihood estimator (MLE). As explained in \cite{pihlaja2010}, the objective of MLE can be expressed as a variational problem, by writing the objective functional as follows,

\begin{equation}
\label{eqn:mlevarf}
    \tilde{\mathcal{J}}[f] = \int p_d \log(\exp(f)) - \int p_n \frac{\exp(f)}{p_n}
\end{equation}

Taking variational derivative wih respect to $f$, the only stationary point is given by $p_d = \exp(f)$ or $f = \log p_d$.

Replacing logarithm and identity by $g_1(\cdot)$ and $g_2(\cdot)$, respectively, Eqn. \ref{eqn:mlevarf} can be expressed as,

\begin{equation}
    \label{eqn:mlevarfg1g2}
    \tilde{\mathcal{J}_g}[f] = \int p_d \;\; g_1\left(\frac{\exp(f)}{p_n}\right) - \int p_n \;\; g_2\left(\frac{\exp(f)}{p_n}\right)
\end{equation}

The sample version of Eqn. \ref{eqn:mlevarfg1g2}, can be expressed as,

\begin{equation}
    \label{eqn:mlevarsamplever}
    \mathcal{J}_g(\theta) = \frac{1}{N_d} \sum_{i=1}^{N_d} g_1\left(\frac{p_m(x_i;\theta)}{p_n(x_i)}\right) - \frac{1}{N_n} \sum_{i=1}^{N_n} g_2\left(\frac{p_m(y_i;\theta)}{p_n(y_i)}\right)
\end{equation}

\noindent
where, $(x_1, x_2, x_3, \hdots, x_{N_d})$ and $(y_1, y_2, y_3, \hdots, y_{N_n})$ are the samples from the data and auxiliary (noise) distributions, respectively.

As $N_d \rightarrow \infty$ and $N_n \rightarrow \infty$, Eqn. \ref{eqn:mlevarsamplever} reduces to,

\begin{equation}
    \label{eqn:mlevarcont}
    \mathcal{J}^{\infty}_g(\theta) = \int p_d \; g_1\left(\frac{p_m(x_i;\theta)}{p_n(x_i)}\right) - \int p_n \; g_2\left(\frac{p_m(y_i;\theta)}{p_n(y_i)}\right)
\end{equation}

Using $g_1(q) = \log(\frac{q}{1+q})$ and $g_2(q) = \log(\frac{1}{1+q})$ in Eqn. \ref{eqn:mlevarsamplever}, and rearranging, we get,

\begin{equation}
    \label{eqn:mlevarnce}
    \begin{split}
    \mathcal{J}_{NC}(\theta) = & \int p_d \; \log\left(\frac{1}{1 + \exp\left(- \log\frac{p_n}{p_m(\theta)}\right)}\right) + \int p_n \; \log\left(\frac{1}{1 + \exp\left(-\log \frac{p_m(\theta)}{p_n}\right)}\right)\\
    = & \int p_d \; \log\left(\frac{1}{1 + \exp\left(- \log\frac{p_n}{p_m(\theta)}\right)}\right) + \int p_n \; \log\left(1 - \frac{1}{1 + \exp\left(-\log \frac{p_n}{p_m(\theta)}\right)}\right)\\
    \end{split}
\end{equation}

Hence, this objective function can be related to the log-likelihood in a nonlinear logistic regression model which discriminates the observed sample of $p_d$ from the noise sample of the auxiliary density $p_n$, which is the very objective of MIOv1.

However, in self-supervised learning, we need to note the following two points: (1) we can always sample a batch which has an equal number of positive and negative pair, (2) we use a non-parametric softmax / sigmoid classifier. Furthermore, the value of the $\nu \; ( = 2N - 2)$ is dependent on the batch size. Thus, for a large batch size, $P(k=1|(z_i, z_j);\theta) \rightarrow 0$ and $P(k=0|(z_i, z_j);\theta) \rightarrow 1$. Consequently, from Eqn. \ref{eqn:nceL}, $\mathcal{L} \rightarrow -\infty$. Hence, the above interpretation of noise contrastive estimation is not entirely valid for self-supervised contrastive learning. To make the posterior probabilities independent of the batch size, we assume that the prior probabilities $P(k=1) = P(k=0) = 0.5$, that is, $\nu = 1$. However, we still need to ensure that the contribution of the positive and negative terms in the loss is equal to each other. Otherwise, the effect of imbalance may have adverse effect on the learning process. Hence, we separately take average of the likelihood terms of the positive and negative pairs, following Eqn. \ref{eqn:mlevarcont}. The resulting form of MIOv1 is thus similar to $\mathcal{J}_g(\theta)$ in Eqn. \ref{eqn:mlevarsamplever}. That is,

\begin{equation}
\label{eqn:ncemiov1}
    \begin{split}
        \mathcal{L}_{v1} = &- \frac{1}{T_p}\sum_{(z_i, z_j) \in \mathcal{X}_+}\ln h((z_i, z_j);\theta)\\ &- \frac{1}{T_n}\sum_{(z_k, z_l) \in \mathcal{X}_-}\ln (1 - h((z_k, z_l);\theta)\\
    \end{split}
\end{equation}

\noindent
where $h((z_i,z_j);\theta)$ can be written as

\begin{equation}
\label{eqn:nceh2}
    h((z_i,z_j);\theta) = \frac{1}{1 + \exp(-\mathcal{G}((z_i,z_j);\theta)}
\end{equation}

\noindent
where $\mathcal{G}((z_i,z_j);\theta)$ gives the logit value for the non-parametric sigmoid classifier (logistic regression), that is, $\mathcal{G}((z_i,z_j);\theta) = \frac{1}{\tau} \left( z_i \cdot z_j^T \right)= \frac{C_{ij}}{\tau}$. $C_{ij}$ denotes the cosine similarity between $z_i$ and $z_j$ and is the SSL equivalent to logit values in non-parametric logistic regression. $\tau$ is the temperature hyper-parameter.\\

\textbf{A different perspective:} If we use Eqn. \ref{eqn:nceL} for MIOv1, where we have already assumed $\nu = 1$, we are causing the contribution of the positive and negative terms to the loss to be imbalanced. On the other hand, using differential averaging for the likelihood of the positive and negative terms, we cause the contribution of the respective likelihood to be equal to each other, and giving the virtual notion of a single positive and negative pair being used in the loss function. Therefore, the differential averaging compensates for the assumption $\nu = 1$. This approach of cost-sensitive learning is often used for learning on imbalanced data using neural networks \cite{he2009limbdata}. Although, for large batch sizes, that is, $T_p, T_n \rightarrow \infty$, the scenario of imbalanced sampling no longer holds true, and $\nu \rightarrow 1$.




\vskip 0.2in
\bibliography{main}

\end{document}

%% file: letterfonts.tex
\newcommand{\opname}{\operatorname}

\renewcommand{\sl}{\opname{\mathfrak{sl}}} 

\newcommand{\bbE}{\mathbb{E}}

	\newcommand{\bbR}{\mathbb{R}}

